\documentclass[11pt,letterpaper]{article}

\usepackage[english]{babel}
\usepackage[T1]{fontenc}
\usepackage[utf8]{inputenc}
\usepackage{lmodern}
\usepackage{microtype}

\usepackage[margin=1in]{geometry}
\usepackage{setspace}

\usepackage{amsmath, amsthm, amssymb, amsfonts}

\usepackage{algorithm}

\usepackage{graphicx}
\graphicspath{{Figure/}}
\usepackage{caption}
\usepackage{float}

\usepackage{booktabs}
\usepackage{multirow}
\usepackage{makecell}
\usepackage{array}

\usepackage{natbib}
\usepackage{xurl}

\usepackage{algpseudocode}

\usepackage{xcolor}
\usepackage{enumerate}

\definecolor{linkblue}{RGB}{24,64,112}
\usepackage[colorlinks=true,allcolors=linkblue,breaklinks=true,pdfencoding=auto,psdextra]{hyperref}
\hypersetup{pdftitle={Smooth Flow Matching for Synthesizing Functional Data},pdfauthor={Jianbin Tan and Anru R. Zhang}}
\floatstyle{ruled}
\restylefloat{algorithm}
\renewcommand{\emptyset}{\varnothing}

\theoremstyle{definition}
\newtheorem{Theorem}{\textbf{Theorem}}

\newtheorem{D}{Definition}
\newtheorem{Ex}{Example}

\newtheorem{remark}{Remark}
\newtheorem{Proposition}{Proposition}
\newtheorem{asum}{Assumption}

\newcommand{\argmin}{\mathop{\rm arg\min}}
\newcommand{\argmax}{\mathop{\rm arg\max}}

\newcommand{\supp}{{\rm supp}}

\def\bm{\boldsymbol}

\newtheorem{LemmaS}{Lemma}

\newtheorem{asumS}{Assumption}

\newtheorem{ExS}{Example}

\newtheorem{TheoremS}{Theorem}

\newtheorem{ConS}{Condition}

\newtheorem{PropositionS}{Proposition}

\title{Smooth Flow Matching for  Synthesizing Functional Data}
\date{}
  \author{Jianbin Tan$^1$ ~ 
     ~ and ~
    Anru R. Zhang$^{1,2}$}

\begin{document}

\maketitle

\footnotetext[1]{Department of Biostatistics \& Bioinformatics, Duke University, Durham, NC, USA}

\footnotetext[2]{Department of Computer Science, Duke University, Durham, NC, USA}

\begin{abstract}
\singlespacing
Functional data, i.e., random functions observed over a continuous domain, are increasingly available in areas such as biomedical research, health informatics, and epidemiology. However, effective statistical analysis for functional data is often hindered by challenges such as privacy constraints, sparse and irregular sampling, infinite-dimensionality, and non-Gaussian structures.
To address these challenges, we introduce a novel framework named Smooth Flow Matching (SFM), tailored for generative modeling of functional data that enables statistical analysis without exposing sensitive real data. Under a copula framework, SFM constructs a semiparametric smooth flow to generate infinite-dimensional functional data, free of Gaussianity and low-rank assumptions.
It is computationally efficient, handles irregular observations, and guarantees the smoothness of the generated functions, offering a practical and flexible solution in scenarios where existing deep generative methods are not applicable.
Through extensive simulation studies, we demonstrate the advantages of SFM in terms of both synthetic data quality and computational efficiency. We then apply SFM to generate clinical trajectory data from the MIMIC-IV patient electronic health records (EHR) longitudinal database. Our analysis showcases the ability of SFM to produce high-quality surrogate data for downstream tasks, highlighting its potential to boost the utility of EHR data for clinical applications.
\end{abstract}
\setstretch{1.08}

\noindent{\it Keywords:} Copula process, Continuous normalizing flow, Functional data, Generative model, Synthetic data analysis

\section{Introduction}

Functional data, comprising sequential observations over a continuous domain, have become increasingly prevalent in scientific domains such as biomedical research \citep{shi2024tempted,happ2018multivariate}, health informatics \citep{tan2024functional,tan2025integrated}, epidemiology \citep{carroll2020time, luo2024functional}, and spatio-temporal analysis \citep{tan2024graphical}. 
Many of these fields, particularly those involving individual-, institutional-, or regional-level data, often deal with sparse and irregular observations and require secure data sharing. 
These constraints limit data quality and availability for the effective analysis of functional data.
To address these challenges, generative models offer a principled approach to learning the underlying mechanisms of data, enabling the synthesis of realistic surrogate observations and boosting data utility for statistical analysis \citep{shen2023boosting, tian2024reliable, golda2024privacy, wu2025denoising}. 
Such models support statistical tasks under privacy constraints, showing strong potential for advancing generative modeling in the context of functional data.

Generative modeling for functional data is a more challenging topic than for traditional multivariate tabular data, as the observations are often smooth random curves residing in infinite-dimensional spaces \citep{ramsay1997functional, hsing2015theoretical}. 
To account for these characteristics, a widely adopted framework is the Karhunen--Lo\`eve (KL) expansion, which models a random function by decomposing it into the sum of a smooth mean function, smooth orthonormal eigenfunctions, and uncorrelated random scores \citep{ramsay1997functional, hsing2015theoretical}. 
However, the KL expansion often struggles when functions are sparsely and irregularly observed, which leads to inaccuracy in score estimation due to limited time observations \citep{yao2005functional}.
To address this, it is common to assume that functional data possess a low-rank structure and are realizations of Gaussian processes \citep{yao2005functional, yao2007functional, paul2009consistency}. 
This assumption enables accurate estimation under sparse and irregular designs of functional data.

Despite the success of the above methods, their limitation is the reliance on the Gaussian assumption, which captures only the mean and covariance signals of the data and may not hold in practice---particularly for data exhibiting heavy-tailed distributions. 
Meanwhile, the low-rankness may impose unnecessary restrictions on the underlying structure, conflicting with the infinite-dimensional nature of functional data and limiting the expressive power of generative methods. 
A relaxation of these assumptions is offered by the copula process framework \citep{wilson2010copula}, which models the joint distribution of random functions by decomposing it into two components: the marginal distributions and a copula that captures  dependencies. 
However, existing copula-based methods typically rely on parametric forms for the marginals or assume dense and regular sampling of functional data \citep{staicu2012modeling, zhang2022high}, limiting their applicability in sparse and irregular settings as well as for data with general marginals.

Recently, another line of research has focused on deep generative models for functional data, providing flexible nonparametric approaches for synthesizing data samples from infinite-dimensional spaces. One prominent framework is built on score-based techniques \citep{lim2023score, kerrigan2023diffusion, franzese2023continuous}, which extends score functions for tabular data \citep{vincent2011connection, song2019generative} to the setting of random functions. In these models, the functional scores are learned via deep neural networks \citep{lecun2015deep}; new samples are then generated by starting from random noise and evolving them along a probabilistic flow guided by the scores.  
Another related framework builds on more general flow-based techniques using vector fields \citep{lipman2022flow, liu2022flow, tong2023improving}, generalizing the concepts of vector fields from vector spaces to infinite-dimensional spaces \citep{kerrigan2023functional}. As in score-based models, the underlying vector fields are learned by deep learning, and new samples are then generated by transporting random noise through the learned vector field \citep{chen2018neural}.

\begin{figure}[htbp]
    \centering
    \includegraphics[width=0.75\linewidth]{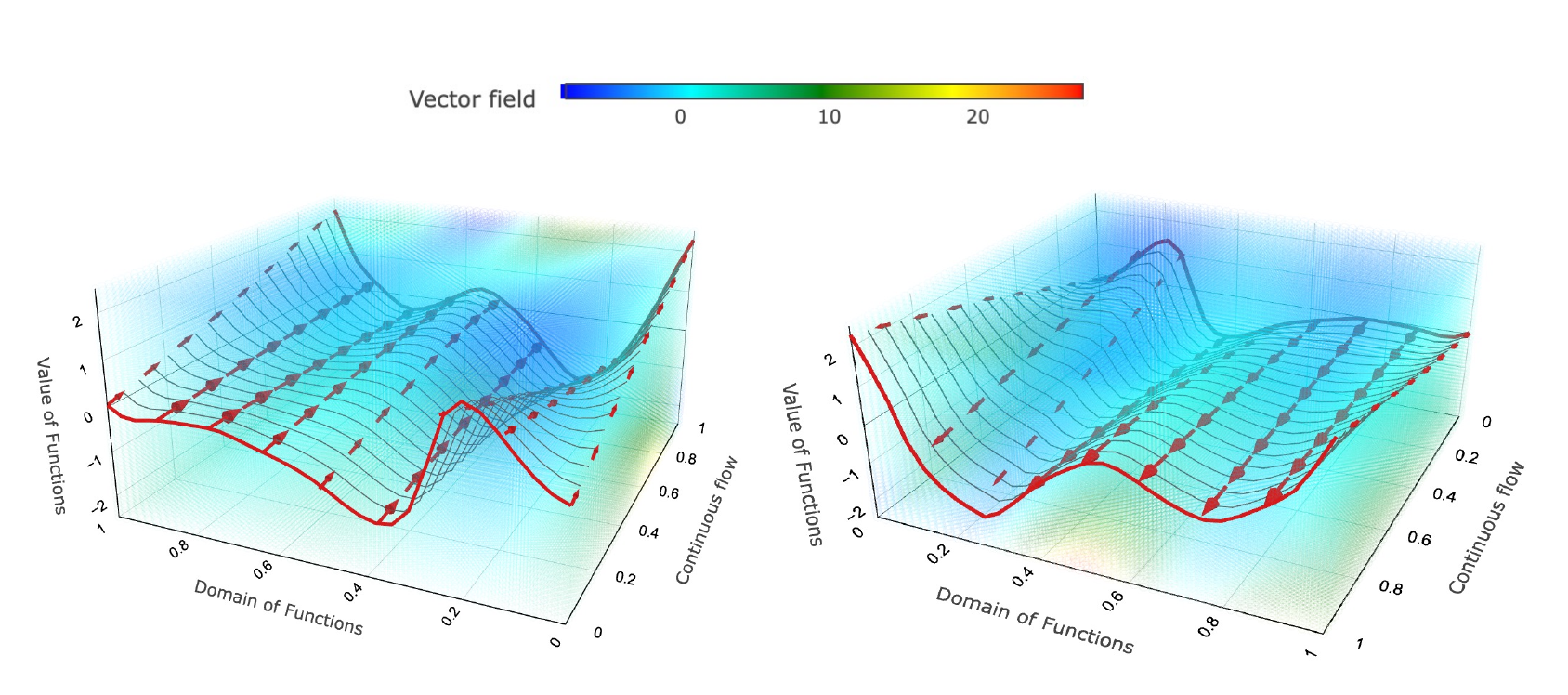}
    \caption{A pictorial illustration of functional data generation via a three-dimensional flow. The left panel shows the base function input into the flow, while the right panel displays the output functions generated by applying the base function through the flow transformation.}
    \label{fig:flow_illu}
\end{figure}

While the above methods offer flexibility, they generally require functional data to be fully (or at least densely) observed over the entire domain. 
This requirement is often impractical in real-world scenarios, where observation times are typically irregular across subjects and frequently sparse \citep{yao2005functional, wang2016functional, tan2024functional}. 
Moreover, the scores and vector fields in these methods are typically formulated as operators between function spaces \citep{lim2023score, kerrigan2023functional}, which are complex objects that demand substantial data samples. 
These operators are modeled using deep neural networks, which further increase the computational burden of model training.
Finally, existing deep generative models provide no guarantee of smoothness in the generated functions, potentially reducing statistical efficiency and leading to unrealistic, non-smooth generations.

To address these issues, we propose a novel generative method named \textit{Smooth Flow Matching} (SFM), designed for generating new functional data based on irregularly and sparsely observed functional samples. 
Our method adopts a copula-based generative framework built upon flow-based techniques, capable of producing non-Gaussian and heavy-tailed functional data without relying on the low-rank condition. 
The SFM framework is constructed via a novel functional-form flow, which enables data generation through a smooth three-dimensional vector field rather than high-dimensional operator-based generators as in \cite{lim2023score, kerrigan2023functional}. 
This construction is significantly simpler---requiring no deep neural networks, allowing computationally efficient generation of functional data while ensuring the smoothness of synthetic samples. 
A schematic illustration of the three-dimensional vector field is provided in Figure~\ref{fig:flow_illu}.

SFM offers several advantages over existing deep generative models and Gaussian-based methods. 
First, SFM leverages flow-based techniques to model the marginal distributions of random functions rather than their full joint distribution, enabling direct flow estimation from irregular and sparse functional data. 
This approach preserves the nonparametric flexibility of the marginals to capture distributional features beyond Gaussianity, while avoiding the high computational cost and dense observation requirements associated with full joint modeling in deep generative methods. 
Second, SFM naturally incorporates smoothness into the flow estimation, allowing it to borrow strength across curves and generate smooth functional data from sparse samples---analogous to mean and covariance estimation in classical functional data approaches \citep{yao2005functional,yao2007functional, hsing2015theoretical}. 
With the estimated smooth flow, we can construct its inverse flow (i.e., data-to-latent flow) to transport the data to latent copula observations, yielding a flexible semiparametric framework for functional data generation. 
A summary of the above comparisons is provided in Table~\ref{generation_comparison}.

\begin{table}[htbp]
\centering
\caption{Comparison of SFM with existing methods for functional data}
\label{generation_comparison}
\renewcommand{\arraystretch}{1.1}
\setlength\tabcolsep{2pt}
\footnotesize
\begin{tabular}{l >{\raggedright\arraybackslash}p{3cm} >{\raggedright\arraybackslash}p{3.5cm} >{\raggedright\arraybackslash}p{2.5cm}}  
\hline
\textbf{Method} & \textbf{Deep Generative Models} & \textbf{Gaussian-based Methods} & \textbf{Smooth Flow Matching (this work)} \\
\hline
\textbf{Data Generator} 
& Operator-form vector field (or score) estimated via deep learning \citep{lim2023score, kerrigan2023functional} 
& Gaussian process modelling via classical functional data approaches \citep{yao2005functional,yao2007functional,hsing2015theoretical}.
& Three-dimensional vector field \\
\textbf{Data Resolution} & Dense & Dense or sparse & Dense or sparse \\
\textbf{Computational Cost} & High & Relatively low & Relatively low \\
\textbf{Guaranteed Smoothness} & No & Yes & Yes \\
\textbf{Joint Distribution} & Nonparametric & Parametric & Semiparametric \\
\hline
\end{tabular}
\end{table}

To demonstrate the utility of SFM, we establish the statistical consistency of the proposed method under practical sparse and irregular designs of functional data. 
Furthermore, we compare SFM with existing deep generative models and Gaussian-based approaches through extensive simulation studies, examining both Gaussian process and heavy-tailed process scenarios. 
Our results show that SFM consistently outperforms existing methods in both cases while substantially reducing computational cost compared to deep generative methods.

Finally, we apply SFM to a clinical trajectory dataset derived from the MIMIC-IV patient electronic health records (EHR) longitudinal database \citep{johnson2020mimic}, where observation times across patients are typically irregular and often sparse. 
Our analysis demonstrates that SFM effectively recovers the dominant longitudinal patterns and key distributional features of the original EHR data. 
We further assess the synthetic performance of SFM through a privacy risk calculation and a prediction task, showing that SFM-generated data exhibit superior predictive power under privacy constraints.
These results highlight SFM's advantage in producing high-quality surrogate EHR data for downstream statistical tasks.

The remainder of this paper is organized as follows. 
In Section~\ref{sec: copula functional data}, we introduce the copula flow framework for functional data. 
After a discussion of notations, we define copula functional data in Section~\ref{sec: copula_fda}, discuss continuous normalizing flows for functional data in Section~\ref{sec:functional-flow}, and examine the smoothness properties of flow generation in Section~\ref{Sec: smooth}. 
We then present the estimation of flows and data generation using SFM in Sections~\ref{sec: flow_matching} and~\ref{sec: SMO}, respectively, focusing on fully observed and sparsely observed functional data, and demonstrate statistical consistency in Section~\ref{sec: con_smo}. 
In Section~\ref{sec:simu}, we conduct extensive simulation studies comparing SFM with existing generative methods. 
Section~\ref{sec:real} applies SFM to clinical trajectory data from electronic health records. 
We conclude with a discussion in Section~\ref{sec:discussions}. 
The code is publicly available at \href{https://github.com/Jianbin-Tan/Smooth-Flow-Matching}{https://github.com/Jianbin-Tan/Smooth-Flow-Matching}.

\section{Flow for Copula Functional Data}\label{sec: copula functional data}

Denote \( \mathcal{L}^2(\mathcal{S}) \) as the space of square-integrable functions from \( \mathcal{S} \) to \( \mathbb{R} \), equipped with the \( \mathcal{L}^2 \) norm \( \|\cdot\|_{\mathcal{L}^2} \) and inner product \( \langle \cdot, \cdot \rangle_{\mathcal{L}^2} \), where \( \mathcal{S} \subset \mathbb{R} \) is a compact support.
For any vector \( \bm{a} = (a_1, \dots, a_n)^\top \), we denote its norm by \( \|\bm{a}\| := \sqrt{\sum_{i=1}^n a_i^2} \). 
For a right-continuous, non-decreasing function \( F \), define its generalized inverse by \( F^{-1}(y) = \inf\{u : F(u) \geq y\} \). 
Let \( \mathbb{I}(\cdot) \) denote the indicator function and \( \delta(\cdot; \cdot) \) be the Dirac measure.
We use ``\( \overset{d}{=} \)'' to denote equality in distribution, and \( \Phi \) to denote the cumulative distribution function (CDF) of the standard normal distribution. 
Let \( \mathcal{N}(a, b) \) denote the normal distribution with mean \( a \) and variance \( b \). 
The support of a continuous random variable \( X \) is denoted by \( \supp(X) \). 
In what follows, we may abbreviate a functional object or stochastic process \( f(\cdot) \) simply as \( f \).

In this section, we develop a copula-based normalizing flow framework for modeling a random function \(X\), with the goal of generating new functional samples that follow the same distribution as \(X\). 
In essence, we utilize copula techniques to decompose the joint distribution of \(X\) into marginals and a copula, and construct a family of maps
\(\{\phi_{u,t}(\cdot): u\in[0,1]\}\) that connects the marginal distribution of \(X\) to a base stochastic process \(Z\).
This construction enables us to capture complex data distributions while ensuring the generation of smooth functions.
In the following, we outline the general principles of copula-based functional data in Section~\ref{sec: copula_fda}, and then introduce a normalizing flow framework for copula modeling and smooth data generation in Sections~\ref{sec:functional-flow}--\ref{Sec: smooth}.

\subsection{Copula Functional Data}\label{sec: copula_fda}

Denote the marginal distribution of \( X(t) \) by \( F_t(x) \), where \( t \in \mathcal{S} \) and \( x \in \mathbb{R} \). 
The copula of \( X(\cdot) \) is defined as the joint cumulative distribution function \( c_{t_1,\ldots,t_m} \), such that for any \( t_1, \ldots, t_m \in \mathcal{S} \) and \( u_1, \ldots, u_m \in [0,1] \),
\begin{eqnarray}\label{copula_def}
    c_{t_1,\ldots,t_m}(u_1, \ldots, u_m) 
    := \mathbb{P}\big(F_{t_1}(X(t_1)) \leq u_1, \ldots, F_{t_m}(X(t_m)) \leq u_m\big).
\end{eqnarray}
Note that \( F_{t_1}(X(t_1)), \ldots, F_{t_m}(X(t_m)) \) are uniformly distributed random variables, whose joint distribution---the copula---characterizes the dependencies among \( X(t_1), \ldots, X(t_m) \). 
The copula framework thus provides a flexible approach for separately modeling the marginal distributions of \( X(t_1), \ldots, X(t_m) \) and their dependence structure, which is particularly useful when direct modeling of the full joint distribution of \( X \) is challenging.

A common approach to modeling the copula is through a latent variable representation known as a copula process \citep{wilson2010copula}.  
We formalize this concept below.

\begin{D}[Copula Process]\label{Def_copula_process}
A stochastic process \( X \) is said to be a copula process with a base process \( Z \) if, for any \( t_1, \ldots, t_m \in \mathcal{S} \) and \( u_1, \ldots, u_m \in [0,1] \),
\begin{eqnarray*}
    c_{t_1,\ldots,t_m}(u_1,\ldots,u_m)
    = H_{\text{base}}\big(
        F_{t_1,\text{base}}^{-1}(u_1),\ldots,
        F_{t_m,\text{base}}^{-1}(u_m);
        t_1,\ldots,t_m
      \big),
\end{eqnarray*}
where \( F_{t,\text{base}} \), \( t \in \mathcal{S} \), are the marginal distributions of \( Z \), and \( H_{\text{base}}(\cdot; t_1,\ldots,t_m) \) denotes the joint distribution of \( Z(t_1), \ldots, Z(t_m) \).
\end{D}

This definition indicates that the copula of \( X \) is determined by a latent base process \( Z \).  
When \( Z(t) \), \( t \in \mathcal{S} \), is a collection of continuous random variables, then
\begin{eqnarray}\label{copula_formula}
&& H_{\text{base}}(h_1,\ldots,h_m; t_1,\ldots,t_m) \nonumber\\
&=& \mathbb{P}\big(
    F_{t_1,\text{base}}^{-1} \!\circ\! F_{t_1}(X(t_1)) \leq h_1, \ldots,
    F_{t_m,\text{base}}^{-1} \!\circ\! F_{t_m}(X(t_m)) \leq h_m
  \big),
\end{eqnarray}
for \( h_1 \in \operatorname{supp}(Z(t_1)), \ldots, h_m \in \operatorname{supp}(Z(t_m)) \),
which implies that the distribution of \( Z \) can be recovered from \( X \) via the transformations \( F_{t,\text{base}}^{-1} \!\circ\! F_t \), \( t \in \mathcal{S} \).

The concept of copula processes has been employed in the modeling of functional data in the literature \citep{staicu2012modeling, zhang2022high}. A commonly used base for copula processes is the Gaussian process (see the example below). More general bases, such as the Student-\(t\) distribution, can also be adopted for copula processes; see Part~B.1 of the Supplementary Materials for details.

\begin{Ex}[Gaussian Copula Process]\it
   Let $\Phi_{\rho}(\cdot; t_1, \ldots, t_m)$ denote the joint distribution function of a mean-zero Gaussian process evaluated at \( t_1, \ldots, t_m \), with a covariance function $\rho$. When $F_{t,\text{base}}$ is taken as $\Phi$, $t\in \mathcal{S}$, and $H_{\text{base}}(\cdot; t_1,\ldots,t_m) =\Phi_{\rho}(\cdot; t_1, \ldots, t_m)$, $t_1,\ldots,t_m\in \mathcal{S}$, we refer to \( X(\cdot) \) as a Gaussian copula process with latent correlation function \( \rho \). 
    It is worth noting that a Gaussian copula process \( X \) satisfies that \( \Phi^{-1} \circ F_t(X(t)) \), for \( t \in \mathcal{S} \), is a Gaussian process with covariance function \( \rho \).  That means
\begin{eqnarray}\label{latent_correlation}
    \rho(t,s)=\mathbb{E}\Big[\big\{\Phi^{-1}\circ F_{t}(X(t))\big\}\big\{\Phi^{-1}\circ F_{s}(X(s))\big\}\Big],\ t,s\in \mathcal{S}.
\end{eqnarray}
\end{Ex}

A copula process can be generated by transporting a latent process, as stated below. 

\begin{Proposition}[Latent Transport for Copula Process Generation]\label{Pro_copula}
Assume that $X(t)$ and $Z(t)$ are continuous random variables for all $t\in \mathcal{S}$.
\( X(\cdot) \) is a copula process with a base $Z(\cdot)$ if and only if there exists a family of continuous and strictly increasing functions \( \{g_t\colon \supp(Z(t)) \to \supp(X(t)) \}_{t \in \mathcal{S}} \) such that the two stochastic processes $\{X(t)\}_{t\in \mathcal{S}}$ and $\{g_t\big(Z(t)\big)\}_{t\in \mathcal{S}}$ are equal in distribution.
\end{Proposition}

Proposition~\ref{Pro_copula} shows that the Gaussian process \( X(t) = \mu(t) + \sigma(t) Z(t) \), the log-Gaussian process \( X(t) = \exp\{\mu(t) + \sigma(t) Z(t)\} \), and skewed processes of the form \( X(t) = \mu(t) + \sigma(t) G^{-1}(\Phi(Z(t))) \) are special cases of copula processes. 
Here, \(Z(\cdot)\) is a Gaussian process with \(Z(t)\sim\mathcal{N}(0,1)\) for \(t\in\mathcal{S}\), \(\mu(t)\) and \(\sigma(t)\) are fixed functions with \(\sigma(t)>0\), and \(G\) is a parametric cumulative distribution function with zero mean, unit variance, and a shape parameter. Gaussian-type processes are commonly used for modelling sparse functional data \citep{yao2005functional}, while skewed processes have been developed for dense functional data \citep{staicu2012modeling}. These models mainly capture low-order moments such as the mean, covariance, and skewness.

More generally, processes of the form 
\begin{eqnarray}\label{latent_model}
    X(t):=F_{t}^{-1}\circ F_{t,\text{base}}(Z(t)),\ t\in \mathcal{S}
\end{eqnarray}
are also copula processes with a base $Z$, where \( F_t \) can be any continuous and strictly increasing distribution function on \( \supp(X(t)) \), \( t \in \mathcal{S} \). 
Through a general pointwise transport  \(\{F_t^{-1}\circ F_{t,\mathrm{base}}: t\in \mathcal{S}\}\) on the latent space, the copula process can capture rich marginal distributional features beyond low-order moments.

\subsection{Pointwise Flow for Functional Data}\label{sec:functional-flow}

In this subsection, we introduce the basic framework for generating copula functional data via a pointwise flow.  
Following \eqref{latent_model}, new functional data can be generated in two steps:  
first, sample a base function \( Z \) from the joint distribution \( H_{\text{base}}(\cdot) \);  
then, generate a new functional sample by applying the transformation \( F_t^{-1} \circ F_{t,\text{base}}(Z(t)) \) for \( t \in \mathcal{S} \).
This approach relies on two key components: the joint distribution \( H_{\text{base}}(\cdot) \) and the transformation \( F_t^{-1} \circ F_{t,\text{base}} \). According to \eqref{copula_formula}, \( H_{\text{base}}(\cdot) \) is further determined by \( X \) after being transformed by \( F_{t,\text{base}}^{-1} \circ F_t \).
Therefore, both maps
\(
F_{t,\text{base}}^{-1} \circ F_t \ \text{and} \ F_t^{-1} \circ F_{t,\text{base}}
\)
play essential roles in generating copula functional data.

We propose to characterize both \( F_t^{-1} \circ F_{t,\text{base}} \) and \( F_{t,\text{base}}^{-1} \circ F_t \) within the framework of continuous normalizing flows \citep{chen2018neural}.
Specifically, we construct a flow map $\phi_{u,t}$ such that 
\begin{eqnarray}\label{data_generation}
\phi_{u,t}(Z_0(t))=Z_u(t),\ t\in \mathcal{S},\ u\in [0,1].
\end{eqnarray}
where $\phi_{0,t}(\cdot) = \text{id}(\cdot)$, $Z_u$, $u\in [0,1]$, is a family of random functions on $\mathcal{S}$ such that \( Z_0 \) is a base process, and \( Z_1 \) is the target function such that $Z_1(t)\overset{d}{=} X(t)$, $t\in \mathcal{S}$.
Simply put, \(\phi_{u,t}\) is constructed to transport the base samples to samples that follow the distribution \(F_t\), for each \(t\in\mathcal{S}\).
Under this, we require the following equation to hold:
\begin{eqnarray}\label{flow_modelling}
    \phi_{1,t}(\cdot)
    =
    F_t^{-1}\circ F_{t,\mathrm{base}}(\cdot),
    \qquad
    t\in\mathcal{S},
\end{eqnarray}
which facilitates the extraction of both \(F_t^{-1}\circ F_{t,\mathrm{base}}\) and its inverse through the flow.

\begin{remark}[Relation to Optimal Transport]\it
Optimal transport is an important topic that has received significant attention in modern statistics \cite{thorpe2018introduction,hutter2021minimax,mena2025statistical}.
In our setting,
\(
P_t := F_t^{-1}\circ F_{t,\mathrm{base}}
\)
is exactly the optimal transport map that pushes \(F_{t,\mathrm{base}}\) forward to \(F_t\), for each \(t\in\mathcal{S}\).
Formally, given a convex and continuous cost function \(c(\cdot)\) satisfying \(c(x)\ge 0\) for all \(x\in\mathbb{R}\), \(P_t\) solves the Monge problem (see Section~2.1 of \cite{thorpe2018introduction}):
\[
\inf_{H_t\in \mathcal{H}_t}
\int c \big(x- H_t(x)\big)\,\mathrm{d}F_{t,\mathrm{base}}(x).
\]
where \(\mathcal H_t\) denotes the collection of maps \(H_t\) such that
\(H_t(Z(t))\sim F_t\) whenever \(Z(t)\sim F_{t,\mathrm{base}}\).
As a result, \eqref{flow_modelling} essentially requires \(\phi_{1,t}(\cdot)\) to be the optimal transport from the base to the target.
\end{remark}

In the following, we construct \( \{\phi_{u,t};\, u \in [0,1]\} \) to accomplish the optimal transport.

\begin{D}[Vector Field for Continuous Normalizing Flow Construction]\label{def_flow}
Consider a $(u,t)$-dependent vector field $V_{u,t}(\cdot)$ from $\mathbb{R}$ to $\mathbb{R}$ and a map $\phi: [0,1]\times \mathcal{S} \times \mathbb{R}\rightarrow \mathbb{R}$ induced by $V_{u,t}$:
\begin{eqnarray}
\label{flow_def}
     &&\frac{\partial \phi_{u,t}(x)}{\partial u}=V_{u,t}(\phi_{u,t}(x)),\quad
    \text{subject to}\quad \phi_{0,t}(x)=x.
\end{eqnarray}
For a functional sequence $\{Z_u; u\in [0,1]\}$, denote \( p_{u,t}(\cdot) \) as the probability density function of \( Z_u(t) \). 
We say the vector field $V_{u,t}$ generates $\{Z_u; u\in [0,1]\}$ or the probability path $\{p_{u,t}; u\in [0,1],t\in \mathcal{S}\}$ if the map $\phi_{u,t}$, $u\in [0,1]$ and $t\in \mathcal{S}$, satisfy
\begin{eqnarray}\label{equ_flow}
    \phi_{u,t}(Z_0(t))=Z_u(t),\quad \text{or equivalently,}\quad p_{u,t}(x)=p_{0,t}(\phi_{u,t}^{-1}(x))\cdot \bigg|\frac{\partial\phi_{u,t}^{-1}(x) }{\partial x}\bigg|.
\end{eqnarray}
\end{D}

In Definition~\ref{def_flow}, the vector field \( V_{u,t} \) establishes a flow map \( \phi_{u,t}(\cdot) \) for each \( t \), transporting \( Z_0(t) \) to \( Z_u(t) \) via the differential equation~\eqref{flow_def}. 
This definition guarantees that \( \phi_{1,t}(Z_0(t)) \overset{d}{=} F_t^{-1} \circ F_{t,\text{base}}(Z_0(t)) \), \( t \in \mathcal{S} \), provided that \( Z_0 \) and \( Z_1 \) have marginal distributions \( F_{t,\text{base}} \) and \( F_t \), respectively.
The following theorem establishes the existence and uniqueness of the solution to \eqref{flow_def}, which ensures that \eqref{flow_modelling} holds.
\begin{Theorem}[Existence and Uniqueness]\label{inverse_formula}
Assume that the vector field \( V_{u,t}(x) \) is uniformly bounded for \( u \in [0,1] \), \( t \in \mathcal{S} \), and \( x \in \mathbb{R} \), and is continuous in \( u \in [0,1] \). Moreover, suppose that for any \( x,y\in \mathbb{R} \) and \( t\in \mathcal{S} \),
\begin{eqnarray*}
    \big|V_{u,t}(x) - V_{u,t}(y)\big| \leq L \cdot |x - y|
\end{eqnarray*}
for some Lipschitz constant \( L \) independent of \( u \), \( x \), \( y \), and \( t \). Then:
\begin{itemize}
  \item[(a)] 
There exists a unique map \( \phi_{u,t} \) that generates \( \{Z_u : u \in (0,1]\} \) given \( Z_0 \) and \( V_{u,t} \).
    \item[(b)] 
The map \( \phi_{u,t}(x) \) in \eqref{flow_def} is a strictly increasing diffeomorphism for Lebesgue-a.e.\ \(x\in \mathbb{R}\), for each \(u\in[0,1]\) and \(t\in\mathcal{S}\).
Moreover, the inverse of \( \phi_{1,t} \) satisfies \( \phi_{1,t}^{-1} = \psi_{1,t} \) for each \( t \in \mathcal{S} \), where 
\( \psi_{1,t}\) is the solution to  
\begin{equation}\label{inver_flow_def}
    \frac{\partial}{\partial u}\, \psi_{u,t}(z)
    \;=\;
    -\, V_{1-u,t}\big(\psi_{u,t}(z)\big),
    \quad
    \psi_{0,t}(z) = z,\ \text{for}\ z \in \{\phi_{1,t}(x) : x \in \mathbb{R}\}.
\end{equation}
    \item[(c)] Suppose \( Z_0(t) \) and \( Z_1(t) \) are continuous random variables, \( t\in \mathcal{S} \). Let the marginal distributions of \( Z_0(t) \) and \( Z_1(t) \) be \( F_{t,\text{base}} \) and \( F_t \), respectively. 
    Then, for each \( t \in \mathcal{S} \), \( \phi_{1,t}(\cdot) \) and \( \psi_{1,t}(\cdot) \) satisfy
    \[
        \phi_{1,t}(z) = F_t^{-1} \circ F_{t,\text{base}}(z) \quad \text{and} \quad 
        \psi_{1,t}(x) = F_{t,\text{base}}^{-1} \circ F_t(x),
    \]
    for Lebesgue-almost every \( z \in \operatorname{supp}(Z_0(t)) \) and \( x \in \operatorname{supp}(Z_1(t)) \).  

    \item[(d)] Under the conditions in (c), suppose that \( X \) is a copula process with marginals \( F_t\), \( t \in \mathcal{S} \), and \( Z_0 \) is the base of \( X \). Then the processes \( X \) and \( Z_1 \) are equal in distribution.
\end{itemize}
\end{Theorem}

Theorem~\ref{inverse_formula} indicates that, under the Lipschitz condition, \( \phi_{1,t}^{-1} = \psi_{1,t} \) exists and can be constructed via \eqref{inver_flow_def} using the vector field \( V_{u,t} \). 
As a result, optimal transport \eqref{flow_modelling} holds, and both maps \( F_t^{-1} \circ F_{t,\text{base}} \) and \( F_{t,\text{base}}^{-1} \circ F_t \) can be generated by \( V_{u,t} \) under the continuous normalizing flow framework (Theorem~\ref{inverse_formula}(c)).
This property leads to valid generation of \( Z_1 \) with the same distribution as \(X\), as ensured by Theorem~\ref{inverse_formula}(d).

\begin{remark}[Flow Modelling via Continuous Normalizing Flows]\it
  Continuous normalizing flows are a flexible subclass of normalizing flow models capable of representing highly complex data distributions \citep{papamakarios2021normalizing}.  
  They offer several key advantages for flow modeling.  
  First, the map \( \phi_{u,t}(\cdot) \) induced by the differential equation is a diffeomorphism, ensuring invertibility throughout the transformation. In our case, the diffeomorphism leads to the ``two-maps-from-one-vector-field'' property, which offers the convenience of modeling both \(F_t^{-1}\circ F_{t,\mathrm{base}}\) and its inverse under the same ODE.
  Second, the use of infinitesimal vector fields in \eqref{flow_def} allows for an effectively infinite sequence of nonlinear transformations. This multi-step transport provides greater expressive power for modelling the nonlinear transformations \( F_t^{-1} \circ F_{t,\text{base}} \) and \( F_{t,\text{base}}^{-1} \circ F_t \).
\end{remark}

\paragraph*{Advantages of Pointwise Flows}
Existing flow-based methods \citep{liu2022flow,lipman2022flow,tong2023improving,kerrigan2023functional} mainly focus on object-level flow transport, that is, designing flows that map the entire function \(Z_0\) to \(Z_1\). In contrast, we construct a pointwise flow that transports \(Z_0(t)\) to \(Z_1(t)\) for each \(t \in \mathcal{S}\). This structure leverages a key property of functional data: although \( \{Z_1(t): t\in \mathcal{S}\} \) is an infinite-dimensional object, it differs from high-dimensional objects in that it may vary smoothly over \(t\).
This ``blessing of infinite dimensionality'' allows us to construct three-dimensional flows as in~\eqref{flow_def}, offering several key benefits:
\begin{itemize}
\item \textbf{Parsimonious structure:}
The flow generated by pointwise flows is substantially simpler than existing 
object-level flows \citep{liu2022flow,lipman2022flow,tong2023improving,kerrigan2023functional}: 
it reduces to a three-dimensional vector field rather than a high-dimensional one (see 
Remarks~\ref{remark_CFM} below for detailed discussions). 
Therefore, we do not encounter issues that usually arise in high-dimensional settings, 
such as instability in differential equation solving \citep{gao2024gaussian,gao2024convergence}.
In addition, the low dimensionality also reduces the computational cost of flow training compared with object-level methods.

   \item \textbf{Accommodation of irregularity and smoothness:} 
In general, the functional nature of the generated data can be captured through the smoothness of \(V_{u,t}(x)\) with respect to \(t\) and \(x\).
This smooth structure, as shown in the following sections, enables \(V_{u,t}(x)\) to be directly estimated from sparse functional observations while ensuring that the generated data exhibit realistically smooth characteristics---a setting that existing object-level flows generally cannot accommodate.

\item \textbf{Copula Preservation}: The pointwise flow also accommodates copula-based functional data generation. Specifically, recall that $\phi_{u,t}\{Z_0(t)\}=Z_u(t)$, the strict monotonicity of $\phi_{u,t}$ (Theorem~\ref{inverse_formula}(b)) implies that the copula of the process $Z_u$ remains the same across different $u$ (for a proof, see Lemma~S5 in the Supplementary Materials).
This property is the key to applying \eqref{flow_def} or \eqref{inver_flow_def} to move between the target process $Z_1$ and $Z_0$ for copula estimation and data generation, as demonstrated in Section~\ref{sec: SMO} below.
\end{itemize}

\subsection{Linking Pointwise Flow via Smoothness}\label{Sec: smooth}

In this subsection, we focus on generating smooth functions using continuous pointwise flows.  
To characterize the smoothness of such functions, we consider the Sobolev space
\[
\mathcal W_{q}^r(\Omega)\;:=\;\Bigl\{ f:\Omega\to\mathbb R\;\Bigm|\; D^{\alpha}f\in \mathcal{L}^{2}(\Omega)\ \text{for all multi-indices }\alpha\in\mathbb N_0^{d}\ \text{with }|\alpha|\le q \Bigr\},
\]
equipped with the norm
\(
\|f\|_{\mathcal W_{q}^{r}(\Omega)}\;:=\;\left(\sum_{|\alpha|\le q}\|D^{\alpha}f\|_{\mathcal{L}^{2}}^{r}\right)^{1/r}.
\)
Here, $\Omega\subseteq\mathbb R^{d}$ is the domain, and for a multi-index $\alpha=(\alpha_1,\ldots,\alpha_d)$ we define
\(
|\alpha|=\alpha_1+\cdots+\alpha_d,\ \text{and}\
D^{\alpha}f \;:=\; \frac{\partial^{|\alpha|} f}{\partial x_1^{\alpha_1}\cdots \partial x_d^{\alpha_d}}
\). When $r= \infty$, $\|f\|_{\mathcal W_{q}^\infty(\Omega)}$ becomes \(\max_{|\alpha|\le q}\sup_{\bm x\in \Omega}|D^{\alpha}f(\bm x)|
\).

Recall that the continuous normalizing flow model (Definition~\ref{def_flow}) solves the differential equation to generate new functional samples:
\begin{equation}\label{ode_ge}
    \frac{\partial Z_{u}(t)}{\partial u} = V_{u,t}(Z_{u}(t)), \quad t \in \mathcal{S},
\end{equation}
given the initial condition \( Z_0(t) \). We propose the following theorem to ensure that the solution to \eqref{ode_ge} lies in the Sobolev space \(\mathcal W^2_q(\mathcal{S}) \).

\begin{Theorem}[Smooth Generation via Smooth Flow]\label{theo: smooth}
Suppose the conditions of Theorem~\ref{inverse_formula}.
For \( u \in [0,1] \), the order-\( q \) partial derivatives of \( V_{u,t}(x) \) for \( t \) and \( x \) exist, and
\begin{equation}\label{smooth_condition}
\frac{\partial^{l+m} V_{u,t}(x)}{\partial t^{m} \partial x^{l}} \in \mathcal{L}^2(\mathcal{S}),
\end{equation}
for any \(l+m=q\), \(u\in[0,1]\), and \(x\in\mathbb R\). Then, for any \( h(\cdot) \in\mathcal W^2_q(\mathcal{S}) \),
we have \( V_{u,\cdot}(h(\cdot)) \in \mathcal W^2_q(\mathcal{S}) \).
Furthermore, if \( \|V_{u,\cdot}(h(\cdot))\|_{\mathcal W^2_q(\mathcal{S})} \) is uniformly bounded for all \( u \in [0,1] \) and \( h \in \mathcal W^2_q(\mathcal{S}) \); and if for any \( h_1, h_2 \in \mathcal W^2_q(\mathcal{S}) \),
\begin{equation}\label{Lipschitz_condition}
\|V_{u,\cdot}(h_1(\cdot)) - V_{u,\cdot}(h_2(\cdot))\|_{\mathcal W^2_q(\mathcal{S})}
\leq L_{\mathcal W^2_q} \cdot \|h_1 - h_2\|_{\mathcal W^2_q(\mathcal{S})},
\end{equation}
where \( L_{\mathcal W^{2}_{q}} \) is a constant independent of \(u\), \(h_{1}\), and \(h_{2}\).  
Then the theoretical solution \(Z_u\) to~\eqref{ode_ge} lies in \(\mathcal W^{2}_{q}(\mathcal{S})\), provided that \(Z_0 \in\mathcal W^{2}_{q}(\mathcal{S})\).
\end{Theorem}

In Theorem~\ref{theo: smooth}, condition~\eqref{smooth_condition} ensures that 
\( \frac{\partial Z_u}{\partial u} \in \mathcal W^2_q(\mathcal{S}) \) whenever 
\( Z_u \in \mathcal W^2_q(\mathcal{S}) \), which implies that 
\( Z_u + \Delta u \cdot \frac{\partial Z_u}{\partial u} \in \mathcal W^2_q(\mathcal{S}) \) 
for any sufficiently small increment \( \Delta u \). 
This is a necessary condition to avoid adding a non-smooth increment 
\( \Delta u \cdot \frac{\partial Z_u}{\partial u} \) to the current function.
Given this, the functional Lipschitz condition~\eqref{Lipschitz_condition} guarantees that \( \lim_{\Delta u \to 0} \left( Z_u + \Delta u \cdot \frac{\partial Z_u}{\partial u} \right) \) exists and lies in \(\mathcal W^2_q(\mathcal{S}) \), ensuring that the solution to~\eqref{ode_ge} remains in \( \mathcal W^2_q(\mathcal{S}) \).

In practice, we approximate the solution of \eqref{ode_ge} using numerical methods (e.g., Runge--Kutta; \citealp{butcher1996history}). 
Proposition~S1 in Part A.3 of the Supplementary Materials shows that condition~\eqref{smooth_condition} is sufficient to guarantee that the numerical solution to \eqref{ode_ge} lies in \(\mathcal W^{2}_{q}(\mathcal{S}) \). 
Under this, we only require that \(Z_0\in \mathcal{W}_q^2(\mathcal{S})\) for smooth function generation, as ensured by the following example.

\begin{Ex}[Gaussian Bases with Smooth Covariance Functions]\label{Gaussian-covariance}
\it 
If \( Z_0(\cdot) \) is a mean-zero Gaussian process with covariance function \( K \), where \( K \) is a positive-definite kernel such that \( \frac{\partial^{2q}K(t,s)}{\partial t^q \partial s^q} \) exists and is continuous on \( \mathcal{S} \times \mathcal{S} \), then the realizations of \( Z_0(\cdot) \) belong to the Sobolev space \(\mathcal W_2^q(\mathcal{S}) \) almost surely \citep{henderson2024sobolev}.   
\end{Ex}

\section{Pointwise Flow Matching for Functional Data}\label{sec: flow_matching}

Suppose \( X(t) \) and \( Z_u(t) \), for \( u \in [0,1] \) and \( t \in \mathcal{S} \), are continuous random variables. 
Here, \( X \) represents a random function, and \( \{Z_u; u \in [0,1]\} \) is a random functional sequence with the probability path \( \{p_{u,t}; u \in [0,1], t \in \mathcal{S}\} \). 
Denote the probability density function of \( X(t) \) by \( f_X(\cdot; t) \). 
In this section, we consider \( X(\cdot) \) to be a copula process on \( \mathcal{S} \) with some base \( Z_0 \), and aim to extract vector fields such that \( Z_1(t) \overset{d}{=} X(t) \), \( t \in \mathcal{S} \). This guarantees that $Z_1$ and $X$ follow the same distribution, as indicated by Theorem~\ref{inverse_formula}(d).

To achieve the above goal, we generalize flow matching \citep{lipman2022flow} to pointwise flow setting.
Specifically, given \( \{Z_u; u \in [0,1]\} \) satisfying \( Z_1(t) \overset{d}{=} X(t) \), \( t \in \mathcal{S} \), the objective of pointwise flow matching is to solve
\begin{eqnarray}\label{loss_FM}
   \min_{U} \int_0^1 \mathbb{E}\bigg\{\frac{\partial Z_u(T)}{\partial u} - U(u,T,Z_u(T))\bigg\}^2 \, \mathrm{d}u,
\end{eqnarray}
where \( U \) is a three-dimensional function on \( [0,1] \times \mathcal{S} \times \mathbb{R} \), and \( T \) is a random variable independent of \( Z_u \)'s. 
Since \( \frac{\partial Z_u(t)}{\partial u} = V_{u,t}(Z_u(t)) \) by \eqref{flow_def}, the minimizer \( U(u,t,x) \) in \eqref{loss_FM} is the vector field \( V_{u,t}(x) \), for \( u \in [0,1] \), \( t \in \mathcal{S} \), and \( x \in \supp(Z_u(t)) \). 
In the following, we propose solving the flow matching \eqref{loss_FM} by a construction of the sequence \( \{Z_u; u \in [0,1]\} \).

\subsection{Conditional Pointwise Flow Matching}\label{Sec: contional_flow}
In this section, we construct a probability path $\{Z_u:\,u\in[0,1]\}$ based on $X$ to facilitate the pointwise flow matching in \eqref{loss_FM}.
Specifically, given \( X(t) \), let \( q_{u,t}(\,\cdot \mid X(t)) \) denote a conditional probability density that satisfies
\begin{eqnarray}\label{flow_condition}
    q_{0,t}(\cdot\mid X(t)) = p_{0,t}(\cdot) \quad \text{and} \quad q_{1,t}(\cdot\mid X(t)) = \delta(\cdot; X(t)), \quad t \in \mathcal{S}.
\end{eqnarray}
Given \( q_{u,t}(\cdot \mid x) \) and \( f_{X}(x; t) \), we define a marginal probability path \( p_{u,t}(\cdot) \) as
\begin{eqnarray}\label{pro_path}
    p_{u,t}(\cdot) := \int_{\mathbb{R}} q_{u,t}(\cdot\mid x) \cdot f_{X}(x; t)\, \mathrm{d}x.
\end{eqnarray}
Then, \( p_{1,t}(\cdot) = f_{X}(\cdot; t) \) by combining \eqref{flow_condition} and \eqref{pro_path}, which means that \( \{p_{u,t};\, u \in [0,1],\, t \in \mathcal{S}\} \) pushes the density \( p_{0,t}(\cdot) \) toward \( f_{X}(\cdot; t) \) as \( u \) varies from 0 to 1.
We will specify the construction of \(q_{u,t}(\cdot\mid x) \) that satisfies \eqref{flow_condition} in the following.

The conditional path provides a way to generate \( Z_{u}(t) \) following the distribution \( p_{u,t} \), which proceeds by first sampling \( X(t) \) and then generating \( Z_u(t) \) based on \( q_{u,t}(\cdot \mid X(t)) \).
This path fulfills the requirement of flow matching if \( \{Z_u; u \in [0,1]\} \) can indeed be generated by a vector field.
The following theorem confirms the existence of such a vector field.

\begin{Theorem}\label{Theo: cond}
Given any random function \( X(t) \) with probability density \( f_{X}(x; t) \) and any initial probability density \( p_{0,t}(\cdot) \), let \( V_{u,t}(\cdot \mid X(t)) \) be a vector field that generates the conditional probability path \( \{q_{u,t}(\cdot \mid X(t)); u \in [0,1] \} \), \( t \in \mathcal{S} \). 
Denote the support of \( p_{u,t} \) in \eqref{pro_path} as \( \supp(p_{u,t}) \), and define
\begin{eqnarray}
\label{Mar_vectorfield}
 V_{u,t}(y) = \int_{\mathbb{R}} V_{u,t}(y \mid x) \cdot \frac{q_{u,t}(y \mid x) \cdot f_{X}(x; t)}{p_{u,t}(y)}\, \mathrm{d}x,\quad y \in \supp(p_{u,t}).
\end{eqnarray}
Suppose that \( V_{u,t} \) satisfies the conditions in Theorem~\ref{inverse_formula}, and \( \{\tilde Z_u(t); u \in [0,1] \} \) is generated by \( V_{u,t} \) in \eqref{Mar_vectorfield} with \( \tilde Z_0(t) \sim p_{0,t} \), then \( \tilde Z_u(t) \sim p_{u,t} \), \( u \in [0,1] \) and \( t \in \mathcal{S} \). Then, the minimizers \( U \) obtained from \eqref{loss_FM} based on the functional sequence \( \{\tilde Z_u(t); u \in [0,1] \} \), and from 
\begin{eqnarray}\label{Conditional_match}
   \min_{U} \int_0^1 \mathbb{E}\bigg\{\mathbb{E}\bigg(\frac{\partial Z_u(T)}{\partial u} - U(u,T,Z_u(T))\bigg)^2 \mid X,T\bigg\}\, \mathrm{d}u,
\end{eqnarray}
with \( Z_u(t) \mid X(t) \sim q_{u,t}(\cdot \mid X(t)) \), are the same.
\end{Theorem}

Two key points are worth noting regarding the optimization~\eqref{Conditional_match}. 
First, we require only the marginal density \(p_{0,t}\), \(t\in\mathcal{S}\), to construct the conditional path \(q_{u,t}(\cdot \mid X(t))\), so full knowledge of the joint distribution of \(Z_0\) is not necessary for vector field estimation. 
Second, the construction of \(Z_u\), \(u\in[0,1]\), in \eqref{Conditional_match} depends only on \(X\), \(p_{0,t}\), and \(q_{u,t}(\cdot \mid X(t))\), and is independent of the vector field \(V_{u,t}\). 
This avoids the construction of \(Z_u\) via \(V_{u,t}\) using the differential equation~\eqref{flow_def}, which bypasses computationally intensive equation solving for vector field estimation.
Similar strategy has also been used in the literature on differential equation estimation \citep{ramsay2017dynamic,tan2024green,li2025sparse}.

\paragraph*{Constructing a Conditional Path}
To implement conditional pointwise flow matching, we need to specify the path \( \{q_{u,t}(\cdot \mid X(t)) : u \in [0,1],\; t \in \mathcal{S} \} \). 
Notably, there are infinitely many possible conditional paths that satisfy the conditions in \eqref{flow_condition}. 
In this article, we adopt the rectified flow \citep{liu2022flow} to construct the conditional probability path as a straight-line interpolation between \( Z_0(t) \) and \( X(t) \):
\begin{eqnarray}
     q_{u,t}(\cdot\mid X(t)) = \delta(\cdot \mid (1-u)Z_0(t) + uX(t)), \quad Z_0(t)\sim p_{0,t},\ t\in \mathcal{S}. \label{tong_flow}
\end{eqnarray}
The flows induced by \eqref{tong_flow} permit the base density \( p_{0,t} \) to be any general distribution \citep{liu2022flow}. 

Using the rectified flow \eqref{tong_flow}, the 
conditional flow matching problem~\eqref{Conditional_match} becomes
\begin{eqnarray}\label{CFM_loss} 
V:&=&\argmin_{U}\int_0^1\mathbb{E}\bigg[\mathbb{E}\bigg\{X(T)- Z_0(T)-U\big(u,{T},Z_{u}(T)\big)\bigg\}^2\mid X,T\bigg]\ \mathrm{d}u\\
&=&\argmin_{U}\int_0^1\mathbb{E}\bigg\{X(T)- Z_0(T)-U\big(u,T,Z_{u}(T)\big)\bigg\}^2\ \mathrm{d}u,\nonumber
\end{eqnarray}
where $Z_0(t)\sim p_{0,t}$, $X(t)\sim f_X(\cdot; t)$, $t\in \mathcal{S}$, and $Z_u(t) = (1 - u) Z_0(t) + u X(t)$, $u \in [0,1]$; $Z_0$, $X$, and $T$ are independent.

\begin{remark}[Comparison to Existing Rectified Flow Matching]\label{remark_CFM}
\it
Existing rectified flow matching for functional data \citep{kerrigan2023functional} primarily focus on the optimization problem
\[
\min_{\mathcal{M}} \int_0^1 \mathbb{E} \big\| X - Z_0 - \mathcal{M}\big(u, Z_u\big) \big\|_{\mathcal{L}^2}^2 \mathrm{d}u,
\]
where \(X\), \(Z_0\), and \(Z_u\) are defined as in~\eqref{CFM_loss}, and \(\mathcal{M}\) is an unknown operator mapping from \([0,1] \times \mathcal{L}^2(\mathcal{S})\) to \(\mathcal{L}^2(\mathcal{S})\).  
In this formulation, \(\mathcal{M}\) aims to transport \(Z_0\) to \(Z_1\) such that \(Z_1\) shares the same distribution as \(X\), whereas \(V\) in~\eqref{CFM_loss} transports \(Z_0(t)\) to \(Z_1(t)\), ensuring that \( Z_1(t) \overset{d}{=} X(t) \), \( t \in \mathcal{S} \).  
Notice that the latter is an essentially simpler transport, and thus the structure of \(V\) is considerably less complex than that of \(\mathcal{M}\).  
As we show below, \(V\) can be consistently estimated from sparsely observed functional data. 
In contrast, \(\mathcal{M}\) can only be estimated from fully observed functional data since it is defined over entire functions.
\end{remark}

\section{Smooth Flow Matching}\label{sec: SMO}

In real-world applications, random functions are only observed at discrete time points rather than being directly measured over the entire continuum \citep{yao2005functional,chiou2007functional,wang2016functional,tan2024functional}. 
This scenario is referred to as sparsely observed functional data \citep{yao2005functional,chiou2007functional,wang2016functional,tan2024functional}. 
It is conventional to model such data as \( X_{i}(T_{ij}) \), where \( i = 1, \ldots, n \) and \( j = 1, \ldots, J_i \), with \( X_{i}(T_{ij}) \) denoting the observation of the \( i \)th function \( X_i \) at time \( T_{ij} \). Here, \( n \) denotes the number of observed functions (sample size), \( J_i \) represents the number of observation times for subject \( i \), and the time points \( \{ T_{ij} : j \in [J_i] \} \) may vary across subjects.

Our objective in this section is to develop a flow matching framework for generating new functional samples from irregularly and sparsely observed functional data.

\subsection{Matching Flows for Discretely Observed Functional Data}

We first consider the noiseless case for the observation \(X_i(T_{ij})\) and then discuss the noisy case in Remark~\ref{Re: noise} below.
Here, we adopt the flow matching method introduced in Section~\ref{sec: flow_matching}. 
First, we take \( Z_0(\cdot) \) to be a stochastic process with known marginal distributions \( F_{t,\mathrm{base}} \), \( t \in \mathcal{S} \), where the copula structure of \( Z_0(\cdot) \) can be chosen arbitrarily at this stage.
Accordingly, we construct an empirical loss \(\mathcal{L}_n(U)\) based on \eqref{CFM_loss}:
\begin{eqnarray}\label{Empirical_FM} 
\frac{1}{n}\sum_{i=1}^n\frac{1}{J_i}\sum_{j=1}^{J_i}\int_0^1\mathbb{E}\bigg[\bigg\{X_i(T_{ij}) - Z_0(T_{ij}) - U\big(u, T_{ij}, Z_{u,i}(T_{ij})\big)\bigg\}^2 \mid X_i, T_{ij}\bigg]\ \mathrm{d}u.
\end{eqnarray}
Here, \( U:[0,1]\times\mathcal S\times\mathbb R\to\mathbb R \) is a three-dimensional function to be estimated; \( Z_{u,i} = (1-u) Z_0 + u X_i \) represents the transported functional sample at time \( u \), initialized with \( Z_0 \), which approaches \( X_i \) at \( \{T_{ij}; j \in [J_i]\} \) as \( u \) increases to 1. 
Let \( \mathcal{X} \) denote a bounded interval in \( \mathbb{R} \) containing all values \( Z_{u,i}(T_{ij}) \).

We propose to restrict \( U \) within a smooth functional space to ensure smooth function generation. To achieve this, we employ spline regression to estimate the vector field.
Let \( \mathcal{B}_{L,q}(\mathbb{D}) \) denote the \( L \)-dimensional B-spline functional space of order \( q \), with equally spaced knots over a domain \( \mathbb{D} \). Denote \( \mathcal{B}_{L,q}(\mathbb{D}_1, \mathbb{D}_2, \mathbb{D}_3) \) as the tensor product of the spaces \( \mathcal{B}_{L,q}(\mathbb{D}_1) \), \( \mathcal{B}_{L,q}(\mathbb{D}_2) \), and \( \mathcal{B}_{L,q}(\mathbb{D}_3) \), which consists of multivariate spline functions defined on \( \mathbb{D}_1 \times \mathbb{D}_2 \times \mathbb{D}_3 \).
Note that the vector field \( U \in \mathcal{B}_{L,4}([0,1], \mathcal{S}, \mathcal{X}) \) satisfies \eqref{smooth_condition} for \( q = 2 \), ensuring the generation of functions from \( \mathcal W_2^2(\mathcal{S}) \).

We consider solving the following optimization problem to estimate the vector field \(V\):
\begin{align}\label{est_vecfil}
    \hat{V}:=\operatorname{argmin}_{U\in \mathcal{B}_{L_V,4}([0,1],\mathcal{S},\mathcal{X})} \big\{{\mathcal{L}}_n(U)&+ \mathcal{J}(U)\big\},\ \\
   \text{with} \quad  \mathcal{J}(U)= \int_{\mathcal{X}}\int_{\mathcal{S}} \int_0^1 \lambda_u\left( \frac{\partial^2 U}{\partial u^2}\right)^2
    +\lambda_t\left( \frac{\partial^2 U}{\partial t^2}\right)^2&+\lambda_x\left( \frac{\partial^2 U}{\partial x^2}\right)^2 \ \mathrm{d}u\mathrm{d}t\mathrm{d}x,\nonumber
\end{align}
where \(\mathcal{B}_{L_V,4}([0,1],\mathcal{S},\mathcal{X})\) denotes the space of cubic splines, with $L_V$ basis functions in each of the three directions; \(\mathcal{J}(U)\) is a penalty on the smoothness of the tensor product of spline spaces \citep{wood2006low}; and \(\lambda_u\), \(\lambda_t\), and \(\lambda_x\) are tuning parameters used to control the smoothness of the vector field in different directions. To ensure that the solutions to \eqref{flow_def} and \eqref{inver_flow_def} with 
\( V = \hat{V} \) exist uniquely, we restrict the class 
\( \mathcal{B}_{L_V,4}([0,1],\mathcal{S},\mathcal{X}) \) in \eqref{est_vecfil} 
to include only vector fields that have bounded 
\( \mathcal{L}^{2}([0,1]\times\mathcal{S}\times\mathbb{R}) \)- and  
\( \mathcal{W}_1^{\infty}([0,1]\times\mathcal{S}\times\mathbb{R}) \)-norms. 
This restriction is imposed for technical convenience but is not required in implementation. 
We refer to this approach as {\it smooth flow matching} (SFM).

\paragraph*{Practical Implementation} We apply Monte Carlo sampling and discrete integration to approximate the expectation and integration in \eqref{est_vecfil}. In detail, we first take a regularly spaced time grid \(\mathcal{S}_{r}\subset \mathcal{S}\) such that \(\cup_{i=1}^n\big\{T_{ij}; j\in [J_i]\big\}\subset \mathcal{S}_{r}\). Then, we generate \( H \) independent and identically distributed functions \( Z_0^{(h)} \), for \( h = 1, \ldots, H \), sampled on the time grid \( \mathcal{S}_r \), where each has marginal distribution \( F_{t,\text{base}} \), \( t \in \mathcal{S} \). Following this, we approximate the expectation with respect to \( Z_0 \) in~\eqref{est_vecfil} by the samples \( Z_0^{(h)} \), constructing the empirical loss $\hat{\mathcal{L}}_n(U)$:
\begin{eqnarray}
\label{Empirical_FM_appro} 
\frac{1}{nHF}\sum_{i=1}^n\frac{1}{J_i}\sum_{j=1}^{J_i}\sum_{h=1}^H\sum_{f=1}^F\bigg\{X_i(T_{ij})- Z_0^{(h)}(T_{ij})-U\big(u_f,T_{ij},Z^{(h)}_{u_f,i}(T_{ij})\big)\bigg\}^2,
\end{eqnarray}
where \( Z_{u,i}^{(h)} = (1 - u) Z_0^{(h)} + u X_i \) is a realization of \( Z_{u,i} \), and \( \{u_f : f = 1, \ldots, F\} \) denotes a fixed, regularly spaced grid in \([0,1]\), with \(F\) being the number of grid points in \([0,1]\).
We then replace \( \mathcal{L}_n(U) \) in \eqref{est_vecfil} with \( \hat{\mathcal{L}}_n(U) \) for the minimization, resulting in a standard nonparametric regression problem.

\begin{figure}[ht!]
    \centering
    \includegraphics[width=0.9\linewidth]{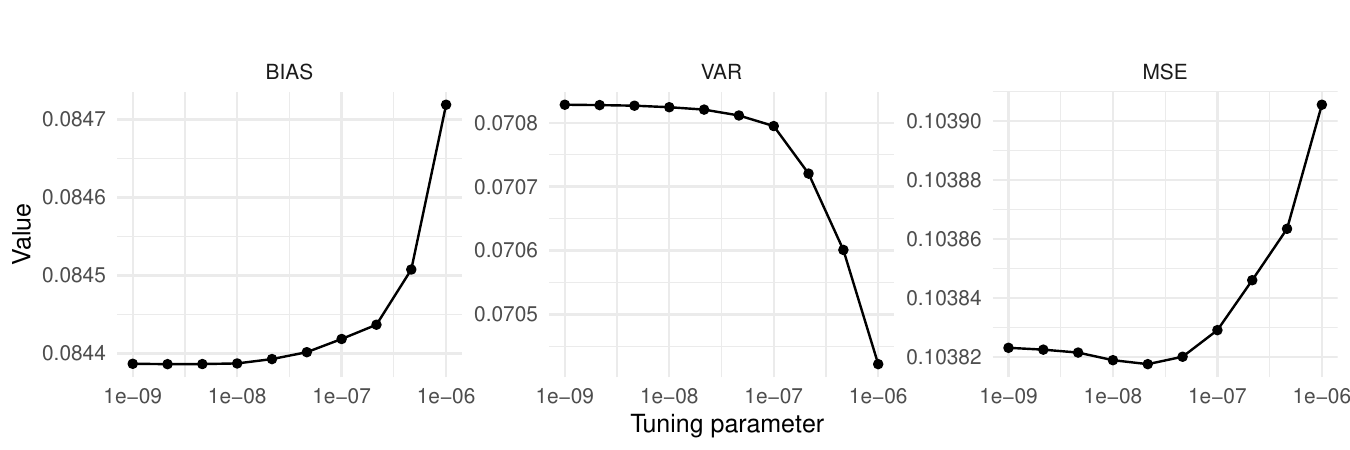}
    \caption{The bias, variance, and mean squared error (MSE) of the estimated vector field $\hat V$, computed from $100$ simulation replications. The estimator is obtained from \eqref{est_vecfil} by replacing $\mathcal{L}_n(U)$ with $\widehat{\mathcal{L}}_n(U)$, with $\lambda_u=\lambda_t=\lambda_x$ taking different values.
    The simulation setting and the definitions of bias, variance, and MSE are given in Part~C.1.1 of the Supplementary Materials.}
    \label{fig:mse_vector}
\end{figure}

{In Figure~\ref{fig:mse_vector}, we conduct a simulation study to examine how the tuning parameters \( \lambda_u \), \( \lambda_t \), and \( \lambda_x \) affect the estimated vector field. 
Overall, we find that as the tuning parameters increase, the bias of the vector field increases while the estimation variance decreases; consequently, the resulting mean squared error first decreases and then increases.} 
To select values for the tuning parameters, we adopt the restricted maximum likelihood (REML) method \citep{paul2009consistency,wood2011fast}.
This approach treats the nonparametric regression as a Bayesian regression problem, and the tuning parameters are estimated via maximum likelihood under the Bayesian framework. 
See Part~B.5 of the Supplementary Materials for a detailed derivation.

To efficiently compute \( \hat{V} \) from \eqref{est_vecfil}, we use the \texttt{R} package \texttt{mgcv} \citep{wood2015package}, which supports both optimization and REML-based tuning for \eqref{est_vecfil}. 
The estimated vector field is then used to estimate the map \( F_t^{-1} \circ F_{t,\text{base}} \) and its inverse \( F_{t,\text{base}}^{-1} \circ F_t \) in the next section.

\subsection{Pullback Estimation and Forward Generation}\label{sec: copula_estimation}

In addition to the vector field \( V \), generating copula functional data requires specifying the joint distribution of the base process \( Z_0 \).
In this subsection, we develop a practical approach to estimate this joint distribution based on the observed data \( \{(X_i(T_{ij}), T_{ij}) : i = 1, \ldots, n,\; j = 1, \ldots, J_i\} \). We first illustrate our procedure using a Gaussian base, i.e., taking \( F_{t,\text{base}} \) in \eqref{latent_model} to be \( \Phi \), \( t \in \mathcal{S} \). Accordingly, we extend the methodology to the Student-\(t\) copula and to more general copula settings in Parts~B.1 and~B.2 of the Supplementary Materials, respectively.

To generate the Gaussian base process, we require estimating the latent correlation function \( \rho \). We achieve this by the following procedures. 
\begin{itemize}
\item 
Based on Theorem~\ref{inverse_formula} (b), we define \(\hat{\psi}_{1,t}(z)\) as the solution to the pullback equation:
\begin{eqnarray}\label{est_forward_ode}
\frac{\partial \hat{\psi}_{u,t}(z)}{\partial u} = -\hat{V}(1-u,t,\hat{\psi}_{u,t}(z))\quad
    \text{subject to} \quad \hat{\psi}_{0,t}(z) = z.
\end{eqnarray}
Then, \( \hat{\psi}_{1,t} \) is an estimate of \( \Phi^{-1} \circ F_t \) in \eqref{latent_correlation}.
Accordingly, we compute
\begin{eqnarray}
\label{empirical_latent}
    G_i(T_{ij_1},T_{ij_2})=\hat{\psi}_{1,T_{ij_1}}\big( X_{i}(T_{ij_1})\big)\cdot \hat{\psi}_{1,T_{ij_2}}\big( X_{i}(T_{ij_2})\big)
\end{eqnarray}
for each \(i, j_1, j_2\), serving as an empirical estimate of \(\rho(T_{ij_1},T_{ij_2})\) in \eqref{latent_correlation}.

\item Second, we propose to apply surface smoothing methods with a penalized bivariate spline on \( \{G_i(T_{ij_1}, T_{ij_2}); i = 1, \dots, n, j_1, j_2 = 1, \dots, J_i\} \) to estimate the correlation function \( \rho \):
\begin{align}\label{est_rho}
    \hat{\rho}
:=&\operatorname{argmin}_{f\in  \mathcal{B}_{L_{\rho},4}(\mathcal{S},\mathcal{S})} \frac{1}{n}\sum_{i=1}^n\frac{1}{J_i^2}\sum_{j_1,j_2=1}^{J_i}\big\{G_i(T_{ij_1},T_{ij_2}) - f(T_{ij_1},T_{ij_2})\big\}^2\nonumber\\
+& \int_{\mathcal{S}}\int_{\mathcal{S}} \left\{\lambda_1\left( \frac{\partial^2 f(t_1,t_2)}{\partial t_1^2}\right)^2
+\lambda_2\left( \frac{\partial^2 f(t_1,t_2)}{\partial t_2^2}\right)^2\right\}\ \mathrm{d}t_1\mathrm{d}t_2.
\end{align}
We adopt a similar strategy as in \eqref{est_vecfil} for optimizing \eqref{est_rho} and tuning the parameters \( \lambda_1 \) and \( \lambda_2 \) using the REML method.
The estimated latent correlation function then ensures that the generated Gaussian processes belong to \(\mathcal W_2^2(\mathcal{S}) \), as guaranteed by Example~\ref{Gaussian-covariance}. 

\item Third, given the estimated vector field and latent correlation function, we generate \( \tilde{Z}_0 \) as a mean-zero Gaussian process with covariance \( \hat{\rho} \), and then apply the transformation \( \widehat{F_t^{-1} \circ \Phi}(\tilde{Z}_0(t)) \), \( t \in \mathcal{S} \), to produce new functional samples. 
Here, the flow map \( \widehat{F_t^{-1} \circ \Phi}(x) \) corresponds to the solution \( \hat{\phi}_{1,t}(x) \) of the forward differential equation
\begin{equation}\label{est_backward_ode}
    \frac{\partial \hat{\phi}_{u,t}(x)}{\partial u} = \hat{V}(u,t,\hat{\phi}_{u,t}(x)), 
    \quad \text{subject to} \quad \hat{\phi}_{0,t}(x) = x,
\end{equation}
in accordance with Definition~\ref{def_flow}.
\end{itemize}

To ensure that \(\hat{\rho}\) is a valid covariance function, we restrict \(\mathcal{B}_{L_{\rho},4}\) to  include only positive semidefinite kernels. In practice, this constraint may be relaxed by first solving~\eqref{est_rho} without the positive-definiteness condition and then applying the nearest positive-definite (NPD) approximation \citep{higham2002computing} to \(\hat{\rho}\) on a chosen time grid. Since the true covariance \(\rho\) is always positive semidefinite, the NPD step can be viewed as projecting \(\hat{\rho}\) back onto the parameter space and thus does not introduce bias.

We summarize the complete SFM generation procedure for Gaussian copula processes in Algorithm~\ref{algo:total_algorithm}, which includes the estimations of vector fields and latent correlation functions. Here, \(\mathcal{S}_r\) is chosen as a dense grid in \(\mathcal{S}\) to approximately cover the observed times; in our implementation, we use 30 equally spaced points. 

Note that we set the covariance of \(Z_0^{(h)}\) to \(\mathbb{I}(t=s)\) in Algorithm~\ref{algo:total_algorithm}. Meanwhile, we take \(F=30\) and \(H=1000\) in~\eqref{Empirical_FM_appro} when computing \(\hat{\mathcal{L}}_n(U)\). In Part~C.1 of the Supplementary Materials, we conduct comprehensive sensitivity analyses for these choices, which show that the generated functions are not sensitive to these settings.

\begin{algorithm}[ht!]
\caption{Smooth Flow Matching for Gaussian Copula Processes}
\label{algo:total_algorithm}
\footnotesize
\begin{algorithmic}[1]
\State \textbf{Input:} Observed data \(\{(X_i(T_{ij}), T_{ij});\ i=1,\ldots,n,\ j=1,\ldots,J_i\}\); number of samples \(H\); number of samples \(M\) for generation; and a regularly spaced time grid \(\mathcal{S}_{r}\) such that \(\cup_{i=1}^n\{T_{ij};\ j\in [J_i]\}\subset \mathcal{S}_{r}\).
\State Generate \(H\) independent mean-zero Gaussian processes \(Z_0^{(h)}\), \(h=1,\ldots,H\), sampled on \(\mathcal{S}_r\), with covariance function \(\mathbb{I}(t=s)\).
\State Obtain the vector field \(\hat{V}\) from
\[
\hat{V} := \argmin_{U\in \mathcal{B}_{L_V,4}([0,1],\mathcal{S},\mathcal{X})} \big\{\hat{\mathcal{L}}_n(U)+ \mathcal{J}(U)\big\}.
\]
\State Compute \(G_i(T_{ij_1},T_{ij_2}) = \hat{\psi}_{1,T_{ij_1}}\big(X_i(T_{ij_1})\big)\cdot \hat{\psi}_{1,T_{ij_2}}\big(X_i(T_{ij_2})\big)\), where \(\hat{\psi}_{1,t}(z)\) is obtained by solving
\[
\frac{\partial \hat{\psi}_{u,t}(z)}{\partial u} = -\hat{V}(1-u,t,\hat{\psi}_{u,t}(z)), \quad
\text{subject to} \quad \hat{\psi}_{0,t}(z) = z.
\]
\State Apply surface smoothing \eqref{est_rho} on \(\{G_i(T_{ij_1}, T_{ij_2});\ i = 1, \dots, n,\ j_1,j_2 = 1, \dots, J_i\}\) to obtain \(\hat{\rho}(t,s)\), and apply the NPD approximation to \(\hat{\rho}\) on \(\mathcal{S}_r \times \mathcal{S}_r\).
\State Generate \(M\) independent mean-zero Gaussian processes \(\tilde{X}_0^{(l)}\), \(l=1,\ldots,M\), sampled on \(\mathcal{S}_r\), with covariance given by the NPD approximation of \(\hat{\rho}(t,s)\).
\State For each \(l=1,\ldots,{M}\) and \(t\in \mathcal{S}_r\), solve
\[
\frac{\partial \tilde{X}_u^{(l)}(t)}{\partial u} = \hat{V}(u,t,\tilde{X}_u^{(l)}(t)), \quad
\text{with initial condition } \tilde{X}_0^{(l)}(t),
\]
to obtain \(\tilde{X}_1^{(l)}(t)\) for each \(t\in \mathcal{S}_r\).
\State \textbf{Output:} Synthetic functions \(\tilde{X}_1^{(l)}(t)\), \(t\in \mathcal{S}_r\), \(l=1,\ldots,M\).
\end{algorithmic}
\end{algorithm}

\begin{remark}[Smooth Flow Matching for Noisy Data]\label{Re: noise}
\it
Smooth flow matching is generally applicable to noisy observations \( Y_{ij} = X_i(T_{ij}) + \varepsilon_{ij} \), where \( \varepsilon_{ij} \) represents white noise. For this case, we apply smoothing splines \citep{gu2013smoothing} to the paired data \( \{(Y_{ij},T_{ij}) : j = 1, \ldots, J_i\} \) for each subject \( i \), yielding smoothed trajectories \( \hat{X}_i \).
This procedure removes observational noise, and  
we then replace \( X_{i}(T_{ij}) \) with the denoised values \( \hat{X}_i(T_{ij}) \), and the dataset \( \{(\hat{X}_i(T_{ij}), T_{ij}) : i = 1, \ldots, n,\ j = 1, \ldots, J_i\} \) is used as input for SFM in Algorithm~\ref{algo:total_algorithm}. 
\end{remark}

\subsection{Consistency of Smooth Flow Matching}\label{sec: con_smo}

In this subsection, we establish the statistical consistency of the proposed smooth flow matching method. 
Suppose that the time points \( \{T_{ij};\ i = 1,\ldots,n,\ j = 1, \ldots, J_i\} \) and the functional data \( \{X_i;\ i = 1,\ldots,n\} \) are independent. 
The number of observations \( J_i \), for \( i = 1,\ldots,n \), is deterministic and bounded. 
We also introduce the following assumptions.

\begin{asum}\label{asum:timepoint}
The time points \( \{T_{ij} \,;\, i = 1, \ldots, n,\; j = 1, \ldots, J_i\} \) are independently drawn from a density \( p_T \), where \( p_T \) is positive and continuously differentiable on \( \mathcal{S} \).
\end{asum}

\begin{asum}\label{asum:X}
The processes \( X_i \) are independent and identically distributed, and satisfy
\(
\sup_{t\in \mathcal{S}} \mathbb{E}\bigl|X_i(t)\bigr|^{4} < \infty.
\)
Similarly, \( Z_0 \) in \eqref{Empirical_FM} satisfies
\(
\sup_{t\in \mathcal{S}} \mathbb{E}\bigl|Z_0(t)\bigr|^{4} < \infty.
\)
\end{asum}

\begin{asum}\label{asum:density}
The densities $f_X(x; t)$ of $X_i(t)$ and $p_{0,t}(x)$ of $Z_0(t)$ are positive and uniformly bounded for
$t \in \mathcal{S}$ and $x \in \mathbb{R}$.
In addition, the maps $(t,x)\mapsto f_X(x;t)$ and $(t,x)\mapsto p_{0,t}(x)$ have continuous partial derivatives
up to order~2. 
\end{asum}

Assumptions~\ref{asum:timepoint} and \ref{asum:X} are standard conditions on the observed time grid and the moments of functional data 
\citep{yao2005functional,hsing2015theoretical,xiao2020asymptotic}, accommodating settings with irregular and sparse observations and guaranteeing the identifiability of the vector field \(V\).
Assumption~\ref{asum:density} imposes smoothness conditions on the marginal densities of the base and target, ensuring controllable errors for vector field estimation. 

\begin{Theorem}[Consistency of the Estimated Vector Field]\label{theo:consist_vec}
Suppose that Assumptions~\ref{asum:timepoint}--\ref{asum:density} hold. 
Then the true vector field \( V(u,t,x) \) in \eqref{CFM_loss} exists uniquely and is contained in
\( \mathcal{W}_2^2\big([0,1]\times \mathcal{S}\times \mathcal{X}\big) \)
for any bounded set \( \mathcal{X} \).
Let \(\tilde  Z_u(t)\sim p_{Z_u(t)} \) be independent of \( X_i \)s, 
\( T_{ij} \)s, and \( Z_0 \), where $p_{Z_u(t)}$ is the density of \( Z_{u}(t) = (1-u) Z_0(t) + u X_i(t) \).
The tuning parameters 
\( L_V, \lambda_u, \lambda_t, \lambda_x \) in \eqref{est_vecfil} satisfy  
\( L_V \to \infty \), \( L_V = o(n^{1/3}) \), and 
\( \max\{\lambda_u, \lambda_t, \lambda_x\} = o(1) \) as \( n \to \infty \).
Then, the estimator \( \hat{V} \) is consistent in the sense that
\[
\mathcal{R}(\hat V, V)
:=\bigg\{\int_0^1\int_{\mathcal{S}} 
\mathbb{E}\Big(\hat V(u,t,\tilde Z_u(t))-V(u,t,\tilde Z_u(t))\Big)^2\,\mathrm{d}t\mathrm{d}u\bigg\}^{1/2}=o_p(1),
\quad n \to \infty.
\]
\end{Theorem}

Note that Theorem~\ref{theo:consist_vec} does not require 
\( \min_{i=1,\ldots,n} J_i \to \infty \) as \( n \to \infty \). 
This is because \eqref{est_vecfil} can borrow strength across subjects due to the temporal smoothness of marginal densities $f_X(\cdot;t)$ and $p_{0,t}$, similar to conventional functional data approaches \citep{yao2005functional,hsing2015theoretical}. 
Consequently, SFM remains applicable in realistic settings where data are 
irregularly and sparsely observed.

Here, we do not provide a convergence rate for vector field estimation via \eqref{Empirical_FM} for discretely observed functional data.
While such discrete settings are common in functional data regression analysis \citep{cai2011optimal,hsing2015theoretical}, a key difference in \eqref{Empirical_FM} is that the time variable $T_{ij}$ not only appears as an explicit argument of the regression function $U$, but is also embedded in another dimension through a random process. This entangled structure complicates the convergence analysis under discrete observation setting.
Besides, minimizing $U$ in \eqref{Empirical_FM} is a typical uncoupled regression problem \citep{rigollet2019uncoupled,hutter2021minimax}, that is, for an observed sample $(X_i(T_{ij}),T_{ij})$ in \eqref{Empirical_FM}, there is no uniquely matched sample $Z_0(T_{ij})$ to pair with $(X_i(T_{ij}),T_{ij})$; instead, any draw of $Z_0(T_{ij})$ can be coupled with $(X_i(T_{ij}),T_{ij})$ through the loss. This uncoupled structure further introduces complications in the convergence analysis.

To quantify the discrepancy between the true distribution of functional data \( X \) and the generated functional data \( \tilde{X} \) from SFM, we use the Wasserstein distance:
\begin{equation}\label{wass}
    W_2(X, \tilde{X}) = \inf_{\gamma \in \Gamma(\nu, \tilde{\nu})}
    \left( \int \|X - \tilde{X}\|_{\mathcal{L}^2}^2 \, \mathrm{d}\gamma(X, \tilde{X}) \right)^{1/2},
\end{equation}
where \( \nu \) and \( \tilde{\nu} \) denote the distributions of \( X \) and \( \tilde{X} \), respectively, and \( \Gamma(\nu, \tilde{\nu}) \) represents the set of all couplings between \( \nu \) and \( \tilde{\nu} \). Under the Gaussian copula framework, we have the following theorem.

\begin{Theorem}[Consistency of the Latent Correlation Function and Generated Functional Data]
\label{theo:consist}
Suppose that Assumptions~\ref{asum:timepoint}--\ref{asum:density} hold and \( J_i \ge 4 \) for all \( i \). Denote 
\(
\bar J := \frac{1}{n}\sum_{i=1}^n J_i.
\)
Assume that \( X_i \) is a Gaussian copula process with latent correlation function \( \rho \), where 
\( \rho\) has continuous partial derivatives up to total order \(2\) on \(\mathcal S\times\mathcal S\).
The base \( p_{0,t}\) is a standard normal density, $t\in \mathcal{S}$, and the true vector field \( V \) satisfies the conditions in Theorem~\ref{inverse_formula}.
Let \(L_\rho\), \(\lambda_1\), and \(\lambda_2\) in \eqref{est_rho} satisfy regularity conditions (Assumption~S1 in Part~A.6 of the Supplementary Materials).
Then the estimator \( \hat{\rho} \) is consistent:
\[
\|  \hat{\rho} - \rho \|^2_{\mathcal{L}^2}
=
O_p\!\left(
\mathcal{R}^2(\hat V, V) + (n\bar J^2)^{-2/3} + n^{-1}
\right),\qquad n \to \infty.
\]
provided that \( \lambda_1 \) and \( \lambda_2 \) are taken at the optimal order:
\(
\min\{(n\bar J^{\,2})^{-2/3}, \bar J^{-4}
\}\;\le\;
\lambda_1,\lambda_2
\;\le\;
\max\{(n\bar J^{\,2})^{-2/3}, n^{-1}\}.
\)
As a result, the Wasserstein distance \( W_2(X, \tilde{X})=o_p(1) \) as \( n \to \infty \).
\end{Theorem}

The rate for estimating \(\rho\) consists of three terms. The term \(\mathcal{R}^2(\hat V, V)\) captures the error from estimating the vector field \(V\). The term \((n\bar J^{\,2})^{-2/3}\) corresponds to the nonparametric smoothing error from bivariate spline fitting, and \(n^{-1}\) represents the estimation error from averaging over \(n\) independent subjects. These terms have the same form as those in covariance function estimation for functional data \citep{hsing2015theoretical,xiao2020asymptotic}, except that the mean function error term in covariance estimation is replaced by the vector field error term.

\begin{remark}[Semiparametric Generation for Sparse Functional Data]
\it
Theorem~\ref{theo:consist} establishes generation consistency for discretely observed functional data under the Gaussian copula framework.  
Here, the latent Gaussian assumption is commonly used for sparse functional data \citep{hall2008modelling,zhou2023functional}, because sparsity hinders fully nonparametric modeling of the dependence structure of random functions.
To address this, our framework provides a semiparametric approach for handling sparse time sampling, which utilizes a parametric copula model for the latent base while not imposing parametric assumptions on the marginal density \(f_X(\cdot; t)\) for \(t\in \mathcal{S}\).
Since vector field estimation (see Theorem~\ref{theo:consist_vec}) does not rely on the Gaussian copula assumption, our method can be extended beyond the Gaussian copula to more general copula-process latent structures when richer dependence is needed.    
\end{remark}

\section{Simulation Studies}\label{sec:simu}

In this section, we evaluate the performance of the proposed SFM method for generating smooth functional data. 
All reported results for each setting are based on 100 replications.

\paragraph*{Data Generation and Comparison Methods} 
We consider both Gaussian and non-Gaussian settings.
For the Gaussian case, we generate sparsely observed functional data using the KL expansion:
\begin{equation*}
X_i(T_{ij}) = \mu(T_{ij}) + \sum_{k=1}^K \xi_{ik} \psi_k(T_{ij}),
\quad i = 1, \ldots, n;\ j = 1, \ldots, J_i,
\end{equation*}
where \( \mu(t) \) is a fixed mean function constructed as a linear combination of degree-4 B-spline basis functions with coefficients independently drawn from \( \mathcal{N}(0,1) \); 
\( \psi_k \), \( k = 1,\ldots,K \), are the first \( K=20 \) non-constant Fourier basis functions; 
\( \xi_{ik} \sim \mathcal{N}(0, \sigma_k^2) \), with \( \sigma_k = \|\mu\|_{\mathcal L^2} \cdot \exp((1 - k)/2) \); 
\( J_i \) is randomly sampled from either \( \{2, \ldots, 6\} \) or \( \{6, \ldots, 10\} \); 
and the time points \( T_{ij} \) are drawn independently from a uniform distribution over an equally spaced 50 point grid on \( [0,1] \).
For the non-Gaussian case, we transform the latent Gaussian process above using a Gamma transformation:
\begin{eqnarray*}
X_i(T_{ij}) = \Gamma^{-1}\!\left(
\Phi_{T_{ij}}\!\left(\mu(T_{ij}) + \sum_{k=1}^K \xi_{ik} \psi_k(T_{ij})\right);
0.5,\ 1\right),
\end{eqnarray*}
where \( \Phi_t \) is the distribution function of the latent Gaussian variable at time \( t \), and \( \Gamma^{-1}(\cdot;\, 0.5, 1) \) denotes the quantile function of the Gamma distribution with shape \( 0.5 \) and rate \( 1 \). 
All other components (\( \mu \), \( \xi_{ik} \), \( \psi_k \), \( K \), \( T_{ij} \), \( J_i \)) are defined as in the Gaussian case.

We compare SFM with four baseline methods. DSM (Denoising Score Matching) is based on score-based diffusion models in function space \citep{lim2023score}, using neural operator techniques to estimate score functions for functional generation. FM (Flow Matching) is a flow based approach that employs neural operators to estimate functional vector fields \citep{kerrigan2023functional}. GP (Gaussian Process Sampling) generates data as a Gaussian process, with the mean and covariance functions estimated using smoothing techniques from functional data analysis \citep{yao2005functional, hsing2015theoretical}. 
Finally, KL (KL Expansion with FPCA) estimates eigenfunctions and eigenvalues via functional principal component analysis \citep[FPCA;][]{yao2005functional, hsing2015theoretical}, and generates new functions by sampling Gaussian FPCA scores, scaled by the estimated eigenvalues, and reconstructing the curves using the estimated eigenfunctions.
{For the Gaussian-based generation methods, see Part~B.6 of Supplementary Materials.}

The DSM and FM methods are implemented in \texttt{Python}, while GP, KL, and SFM are implemented in \texttt{R}. SFM is implemented using Algorithm~\ref{algo:total_algorithm}, adopting smooth estimation \eqref{est_vecfil} for the vector field using the tensor product spline \( \mathcal{B}_{6,4}([0,1],\mathcal{S},\mathcal{X}) \). For modeling the scores or vector fields in DSM and FM, we employ the Fourier Neural Operator (FNO; \citealp{li2020fourier}) architecture for modeling, with 20 Fourier modes, 3 layers, and a channel width of 30. Because DSM and FM require fully observed functions, we first interpolate each sparse trajectory using cubic splines on the 50 point time grid before training and sampling. All experiments are conducted on a high performance machine with 40 cores and 208 GB RAM.

\paragraph*{Evaluation Measure} We adopt the Wasserstein distance \eqref{wass} to quantify the discrepancy between the true distribution of functional data and that of the generated functional data. We draw 100 samples from each distribution and compute the empirical Wasserstein distance using the \texttt{transport} package in \texttt{R} \citep{gottschlich2014shortlist}.

\begin{figure}[htbp]
    \centering
    \includegraphics[width=.95\linewidth]{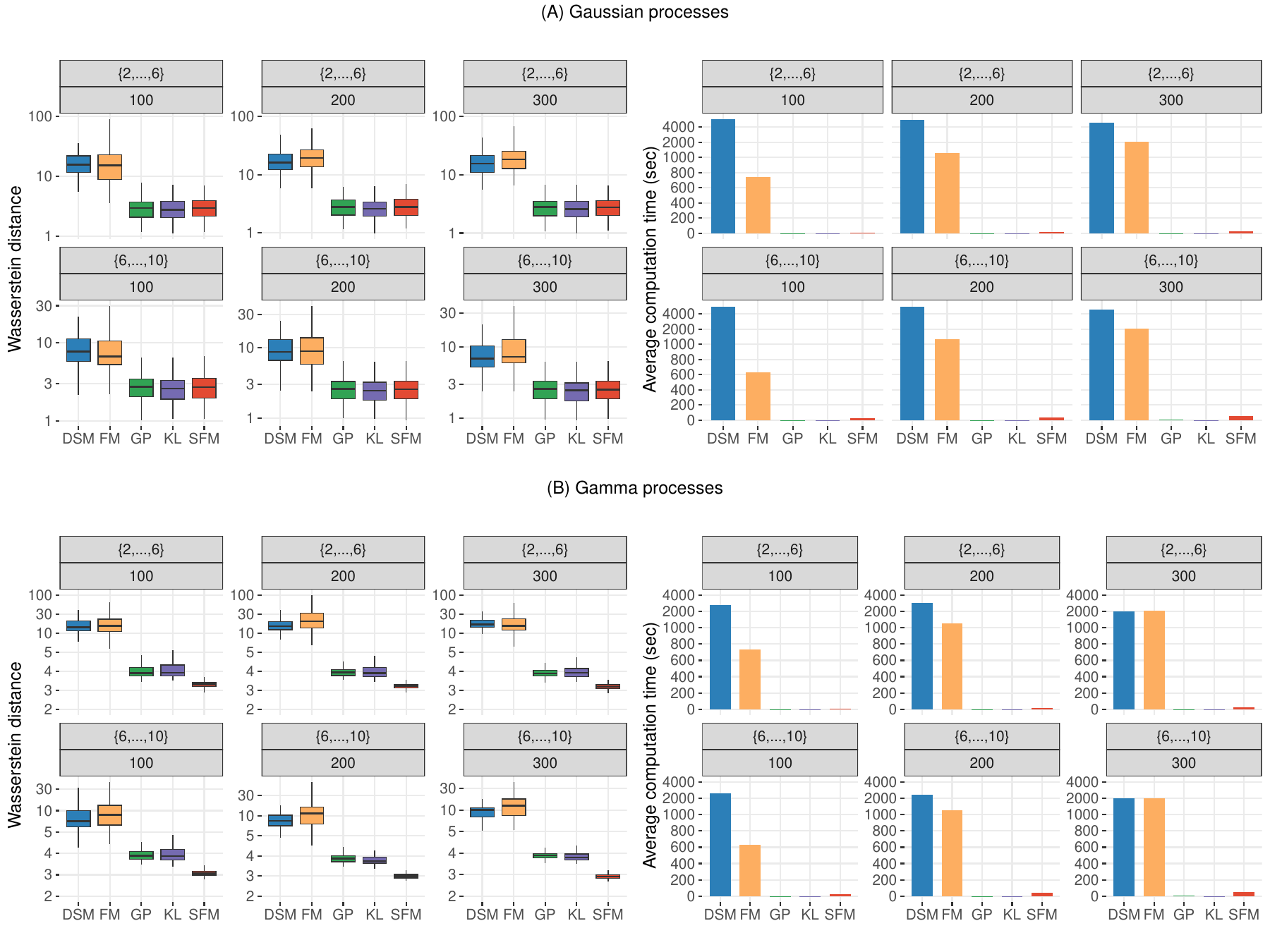}
    \caption{Left panel: Box plots of Wasserstein distances from 100 simulation replications under varying sample sizes \( n \) (subtitles) and numbers of observed time points \( J_i \) (main titles). Right panel: average computation times.}
    \label{fig: Diserror}
\end{figure}

\paragraph*{Results} Figure~\ref{fig: Diserror} summarizes the results across different combinations of sample sizes \( n \) and sparsity levels \( J_i \). 
It reports both the Wasserstein distance between the true and generated distributions from replicated simulations and the average computation time required to estimate the data generator for each method. 
We further compare the estimated mean functions and eigenfunctions between the real and synthetic data in Part~C.2 of the Supplementary Materials.

For the Gaussian case, SFM performs comparably to GP and KL, even though it does not explicitly assume Gaussianity in its modeling procedure. 
In contrast, DSM and FM yield the poorest performance when the observed functional data are relatively sparse. 
In Part~C.3 of the Supplementary Materials, we further evaluate the performance of DSM and FM in a densely observed case, where the interpolation step in DSM and FM has little impact for data generation. We observe that the performance of DSM and FM is similar to that of SFM, suggesting that the unsatisfactory behavior of DSM and FM in sparse cases (Figure~\ref{fig: Diserror}) is attributed to the uncertainty caused by the interpolation step.

For the non-Gaussian (Gamma) case, DSM and FM continue to underperform, while GP and KL now exhibit the largest errors, reflecting their strong reliance on the Gaussian assumption that is violated in this setting.
Across both scenarios, SFM consistently produces the most accurate or at least competitive functional samples compared to all other methods. In addition, SFM is substantially more computationally efficient than DSM and FM because it bypasses the need for neural operator estimation during training. 

In Figure~S2 of the Supplementary Materials, we compare the SFM-estimated latent correlation function $\hat{\rho}$ with the true function, with the number of observed time points set to $\{2,\ldots,6\}$. The results show that $\hat{\rho}$ is consistent as the sample size $n$ increases, in accordance with Theorem~\ref{theo:consist}.

\begin{figure}[htbp]
    \centering
    \includegraphics[width=0.9\linewidth]{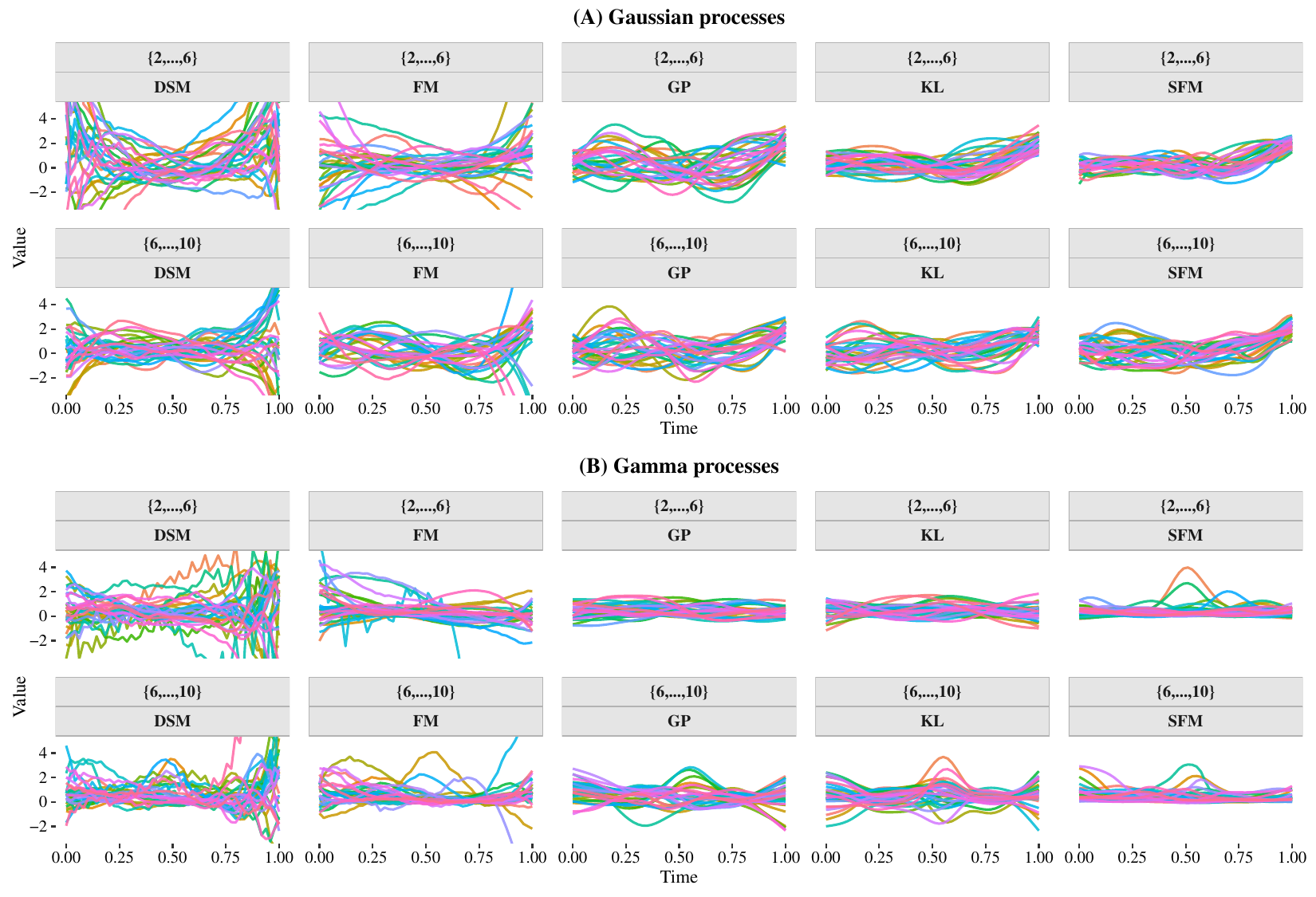}
    \caption{Illustration of generated functional data under different numbers of observed time points (main titles) and different generation methods (subtitles). The sample size is set to \( n = 100 \) for data generation.}
    \label{fig: generated_fda}
\end{figure}

We illustrate the functional data generated by different methods in Figure~\ref{fig: generated_fda} and provide smoothness diagnostics on the generated curves in Figure~S8 of the Supplementary Materials. 
In both the Gaussian and Gamma settings, we find that the roughness of the curves generated by DSM and FM is significantly larger than that of the curves generated by SFM. 
This demonstrates the superior ability of SFM to produce smooth, realistic functional trajectories.

Note that GP and KL yield visually plausible samples in the Gaussian case; however, in the Gamma case, their outputs frequently take negative values (Figure~\ref{fig: generated_fda}), contradicting the true data’s positive support. 
In contrast, our proposed SFM method generates smooth functional data that largely align with the correct support in both the Gaussian and Gamma settings.

Finally, we provide a sensitivity analysis to evaluate the impact of measurement noise on SFM in Part~C.5 of the Supplementary Materials. To this end, we simulate  functional data by adding Gaussian noise. SFM is then applied both directly to the noisy data and to denoised data obtained following Remark~\ref{Re: noise}, and we evaluate the generated synthetic data based on Wasserstein distances. 
We observe that the generation performance of SFM generally deteriorates as the noise level increases. 
Nonetheless, applying a denoising step prior to generation can improve the performance of SFM, making it less sensitive to the added noise.

\section{Real Data Analysis}\label{sec:real}

In this section, we apply the proposed SFM method to the MIMIC-IV EHR database \citep{johnson2020mimic}, which contains de-identified ICU records from the Beth Israel Deaconess Medical Center collected between 2008 and 2019. The dataset includes longitudinal clinical measurements recorded during ICU stays, offering valuable insights into the temporal dynamics of patient health. Such data are critical for monitoring clinical progression, informing medical decisions, and developing predictive models in healthcare.

Our goal is to synthesize realistic surrogate patient trajectories that preserve the underlying temporal patterns of the observed data. We compare the performance of SFM with two baseline methods, DSM and FM, as introduced in the simulation section. DSM and FM were implemented in \texttt{Python}, while SFM was implemented in \texttt{R}. 

We focus on ICU patients who were eventually discharged. To mitigate the influence of outliers, we remove the top 1\% of measurements across all patient trajectories for each clinical feature. Each patient’s data are then modeled as a random function over a time domain, where time zero corresponds to ICU admission. The endpoint of the time domain is chosen such that 80\% of all patient observations fall within this interval. We retain patient level trajectories with at least four time points. To reduce observational noise, we apply smoothing splines to each patient’s functional data, consistent with the denoising procedure described in Remark~\ref{Re: noise}. Due to substantial variability in measurement times across patients, the resulting functional data remain irregularly sampled for each feature after preprocessing.

\begin{figure}[htbp]
    \centering
    \includegraphics[scale = 0.35]{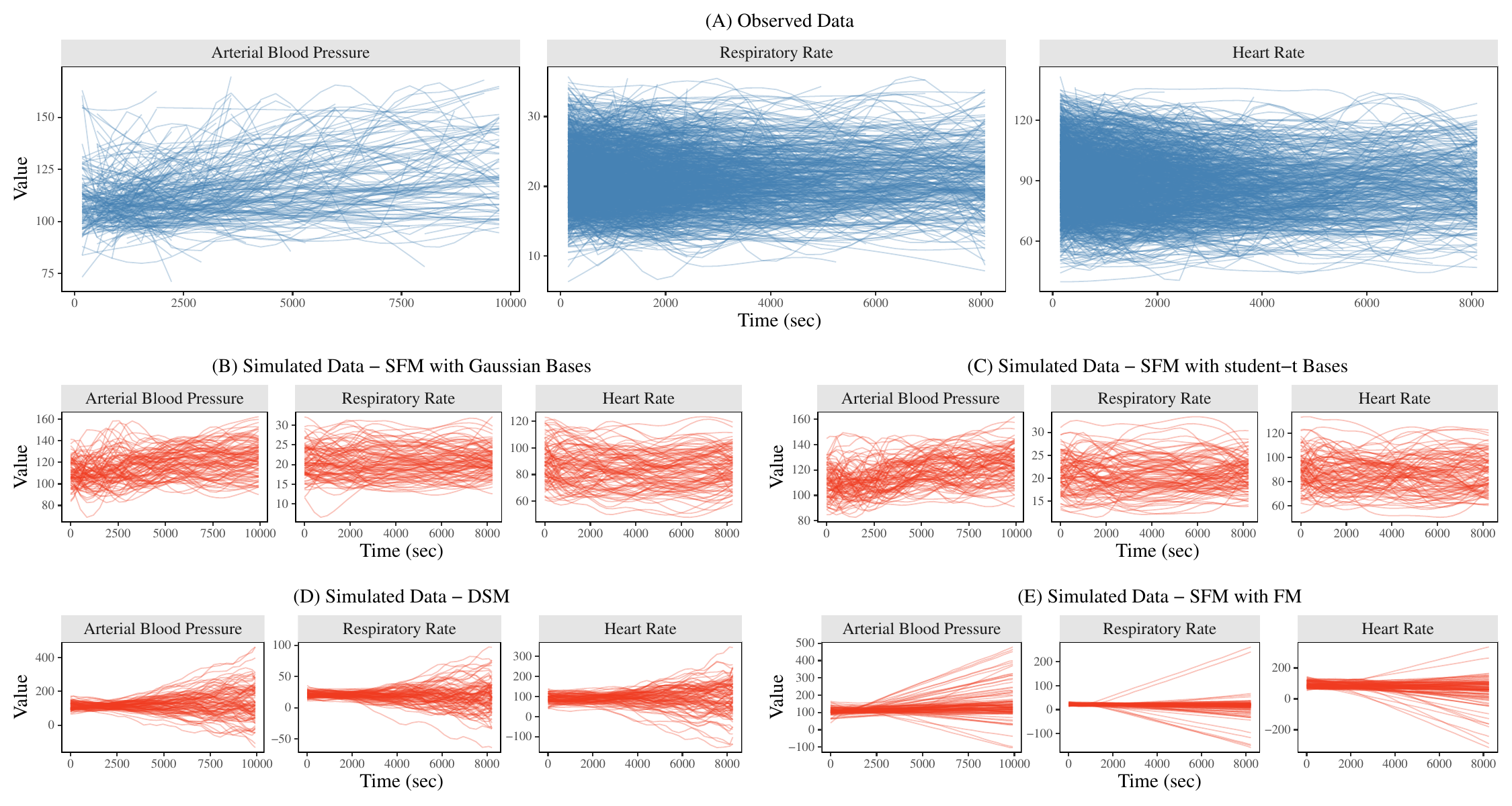}
    \caption{\textbf{(A)} Irregularly observed functional data for each clinical feature; 
    \textbf{(B)} Synthetic functional data generated by SFM with Gaussian bases; 
    \textbf{(C)} Synthetic functional data generated by SFM with Student-\(t\) bases; 
    \textbf{(D)} Synthetic functional data generated by DSM; 
    \textbf{(E)} Synthetic functional data generated by FM.}
    \label{dat:real_simulated}
\end{figure}

Our analysis focuses on three clinical features: arterial blood pressure, respiratory rate, and heart rate. The corresponding numbers of patients are \( n = 256 \), \(1446\), and \(1446\), respectively. The preprocessed and denoised functional data for these features are shown in Figure~\ref{dat:real_simulated}~(A).

We apply SFM with both Gaussian and Student-\(t\) bases (Algorithm~\ref{algo:total_algorithm} and Algorithm~S1 in the Supplementary Materials) to each clinical feature to synthesize new functional samples. 
To evaluate performance, we compare SFM with DSM and FM, using the same neural network architecture as in the simulation study. 
Here, we further utilize GPU acceleration to speed up the computation of DSM and FM.
Because DSM and FM require fully observed functional trajectories, we impute missing values by applying smoothing splines, as previously described, over a fixed evaluation grid \( \mathcal{S}_r \).
This grid consists of 30 equally spaced time points spanning the observed time domain of each clinical feature, approximately covering all time points at which any patient was measured.

Figure~\ref{dat:real_simulated}~(B)--(E) displays synthetic functional samples generated by SFM, DSM, and FM, evaluated on \( \mathcal{S}_r \). 
For each method and clinical feature, we generate 100 synthetic samples. 
The computational training times for SFM, DSM, and FM are summarized in Table~\ref{tab:time}. 
These results show that SFM achieves substantially shorter training times than DSM and FM, highlighting its superior computational efficiency.

\begin{table}[H]
\centering
\setlength{\tabcolsep}{10pt}
\footnotesize
\renewcommand{\arraystretch}{1}
\caption{Computational training time (in seconds) for different data generation methods. All experiments are conducted on a high performance machine with 40 CPU cores, 208 GB RAM, and a single NVIDIA RTX A5000 GPU.}
\begin{tabular}{lcccc}
\hline
 & SFM (Gaussian base) & SFM (Student-$t$ base) & DSM & FM \\
\hline
Arterial blood pressure & 30.28  & 40.92  & 410.90 & 188.28  \\
Respiratory rate        & 37.05 & 52.94 & 597.67 & 210.36 \\
Heart rate              & 37.71 & 54.57 & 574.73 & 200.01 \\
\hline
\end{tabular}
\label{tab:time}
\end{table}

In Figure~\ref{dat:real_simulated}, the functional samples generated by DSM and FM differ substantially from the observed data. 
This discrepancy primarily arises from the irregular sampling in the original dataset, which leads to a high proportion of missing values that must be imputed prior to model training. 
The imputation process not only denoises the observed values but also extrapolates data at time points outside the range of actual observations for certain patients. 
Such extrapolation can introduce considerable error, which may poorly capture the true underlying dynamics and consequently degrade the quality of the synthesized samples.

\paragraph*{Evaluation on Dimension Reduction} To evaluate the quality of the synthesized data, we estimate the mean function and the first two eigenfunctions from both the observed and synthesized functional data using FPCA. 
For each clinical feature, the synthetic sample size is set equal to the number of patients. 
The results are presented in Figure~S10 of Supplementary Materials.
We find that the mean and eigenfunctions estimated from DSM- and FM-generated samples differ substantially from those of the observed data. 
In contrast, the estimates derived from SFM-synthesized data closely match the corresponding patterns from the real data, demonstrating SFM’s ability to recover key temporal dynamics in EHR trajectories. 

\begin{figure}[ht!]
    \centering
    \includegraphics[width=0.8\linewidth]{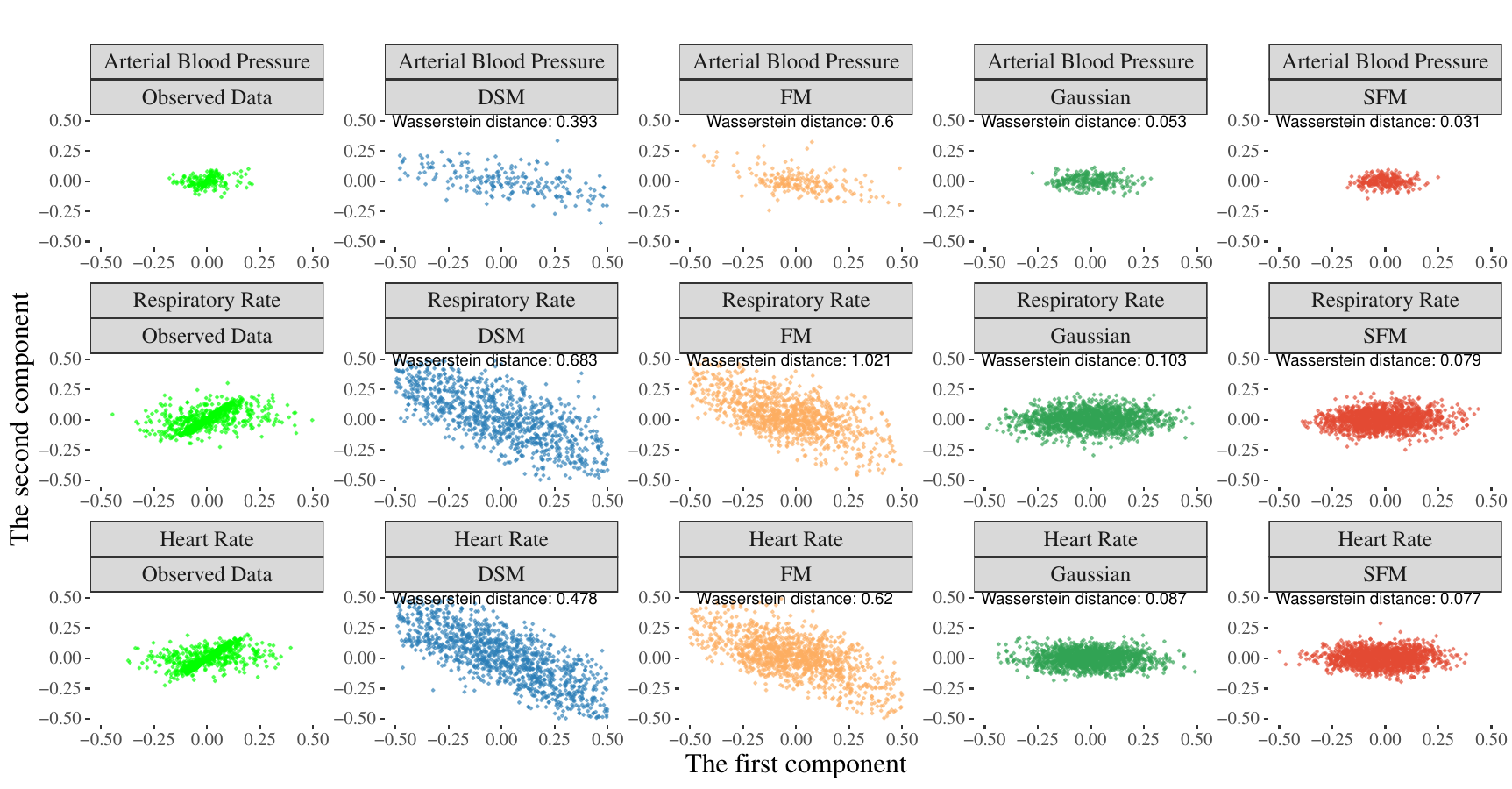}
    \caption{
    Projected scores for the observed data, synthetic data, and Gaussian-based samples. 
    The Wasserstein distances in the last four columns are computed between the corresponding samples and those from the observed data.}
    \label{fig:functional_pattern}
\end{figure}

To further assess performance, we project both the observed and synthesized functional data onto the eigenfunctions estimated from the observed data (green curves labeled EF1 and EF2 in Figure~S10 of Supplementary Materials). 
For the synthesized data, projections are computed via numerical integration over the fixed grid \( \mathcal{S}_r \), yielding two-dimensional score vectors for each subject. 
To handle irregular sampling in the observed data, we apply a shrinkage-based score estimation approach following \cite{yao2003shrinkage}. 
These scores represent the dominant modes of variation in the data, with variances determined by the corresponding eigenvalues.

As an additional baseline, we include a Gaussian score-generation method that samples each score dimension independently from a mean-zero Gaussian distribution, with variances equal to the corresponding eigenvalues estimated from observed data \citep{yao2005functional}. 
The score distributions for the observed data, SFM (Gaussian base), DSM, FM, and the Gaussian-score method are shown in Figure~\ref{fig:functional_pattern}, with synthetic sample sizes matched to the observed data.

We observe that the scores generated by DSM and FM deviate markedly from those of the observed data, reflecting poor alignment with the underlying data structure. 
In contrast, the scores from SFM and the Gaussian-score method both exhibit distributions similar to those of the observed data, with SFM showing even closer alignment. 
This suggests that SFM may capture higher-order and non-Gaussian characteristics present in the EHR trajectories—features that are not well represented by purely Gaussian score-generation approaches.

\paragraph*{Privacy Preservation Assessment}
To assess privacy preservation, we test whether the synthetic data are excessively similar to the training data, which indicates data memorization and potential privacy leakage. Following prior works \citep{yale2020generation,tian2024reliable}, we randomly split the observed data into two equal halves (training and testing), generate synthetic data based only on the training set using SFM, and compare the similarity between training--synthetic and testing--synthetic pairs. 

Because the observed trajectories are irregularly sampled, we project both the observed and synthetic data onto eigenfunctions estimated from the observed data when calculating similarity. Here, the number of projected components for the observed data is chosen such that the fraction of variance explained exceeds \(95\%\), and we use the same number for the synthetic data. 
We then quantify privacy loss by computing
\[
\text{Privacy.loss}(\text{synthetic}) 
= d(\text{training}, \text{synthetic}) 
- d(\text{testing}, \text{synthetic}),
\]
where \(d(\cdot,\cdot)\) is taken as the nearest-neighbor distinctness metric \citep{yale2020generation}, measuring the similarity between the projected score distributions of two datasets. Under this, a privacy-loss value that deviates from zero suggests potential memorization and privacy leakage \citep{yale2020generation}.

\begin{figure}[htbp]
    \centering
    \includegraphics[width=0.9\linewidth]{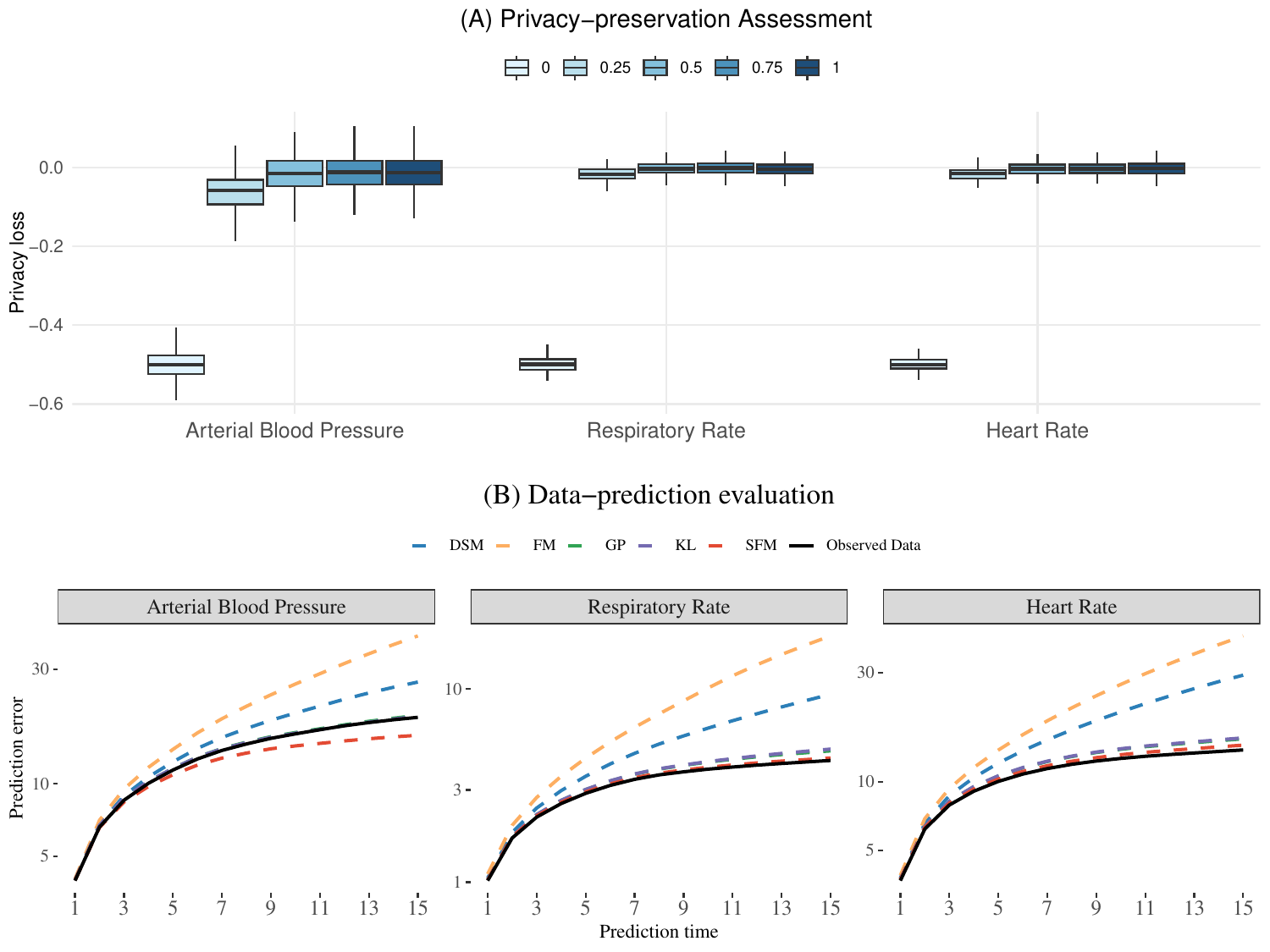}
    \caption{
    \textbf{(A)} Boxplots of privacy-loss values for different mixture proportions $a$ (see legend) for datasets of three clinical features. 
    For each setting, the privacy-loss values are computed over 200 replicated simulations. 
    \textbf{(B)} Prediction errors of different methods across various prediction horizons \( g \).
    }
    \label{fig: real_prediction}
\end{figure}

For illustration, we compute
\(
\text{Privacy.loss}(a \cdot \text{synthetic} + (1-a) \cdot \text{training}) 
\),
which evaluates privacy loss under different mixture proportions \(a\) of synthetic and training data. 
Here, \(a = 0\) corresponds to using only training data, a baseline representing full privacy leakage.  
We repeat the above experiment 200 times and report the resulting privacy.loss values in Figure~\ref{fig: real_prediction} (A).
We observe that, as \(a\) increases, the privacy loss tends toward \(0\) for all three clinical features, indicating that a higher proportion of synthetic data generally reduces privacy loss relative to the baseline. 
In particular, when \(a = 1\), most of the privacy loss values are centered around \(0\), which means that the synthetic data do not differ significantly from either the training or testing distributions. 
This result provides evidence of privacy preservation.

\paragraph*{Data Prediction Evaluation}

To assess the practical utility of the synthesized data, we consider a one-step-ahead prediction task:
\begin{eqnarray}\label{reg_mol}
    X_i(T_{j+1}) = f(X_i(T_j), T_j) + \tau_{ij}, \quad i = 1, \ldots, n,\ j = 1, \ldots, 29,
\end{eqnarray}
where \( X_i(T_j) \) denotes the synthesized or observed denoised functional data for subject \( i \) at time \( T_j \);  
\(\mathcal{S}_r = \{T_j;\ j = 1, \ldots, 30\}\) is an equally spaced time grid;  
\( \tau_{ij} \) are mean-zero noise terms;  
and \( f(\cdot, t) \) is the unknown prediction function at time \( t \).  
We compare the prediction functions estimated from synthesized and observed functional data to evaluate the predictive utility of each dataset. 
Estimation is carried out using nonparametric regression based on \eqref{reg_mol}; see Part~B.7 of the Supplementary Materials for details.

We estimate \( f \) separately from the original observed data and from synthetic datasets generated by SFM, FM, and DSM. 
For comparison, we also include synthetic datasets produced by the GP and KL methods described in the simulation study. 
Predictive performance is evaluated using the \( g \)-step-ahead prediction error on the observed validation data:
\[
\text{Error}_g =
\sqrt{
\sum_{i\in \mathcal{V}} \sum_{j=1}^{30-g}
\left\{ X_i(T_{j+g}) - \hat{f}^g_{T_j}(X_i(T_j)) \right\}^2
\cdot \mathcal{I}_{T_j, T_{j+g}}(X_i)
},
\]
where \( \mathcal{I}_{T_j,T_{j+g}}(X_i) = 1 \) if \( X_i \) is observed at both \( T_j \) and \( T_{j+g} \), and \( 0 \) otherwise;  
and \(
\hat f^1_{T_j}(x):=\hat f(x,T_j),
\
\hat f^g_{T_j}(x)
:=
\hat f\!\left(
\hat f^{g-1}_{T_j}(x),
T_{j+g-1}
\right),
\ g\ge2,\)
is the recursively defined \( g \)-step-ahead predictor.
Here, \( \mathcal{V} \) indexes the validation set.  

For the observed data, we randomly split the dataset into training (\( \mathcal{E} \)) and validation (\( \mathcal{V} \)) subsets of equal size.  
We train \(\hat{f}\) on \( \mathcal{E} \) and evaluate performance on \( \mathcal{V} \).  
For the generative methods, we synthesize a dataset \( \mathcal{E}_{\text{syn}} \) matching the size of \( \mathcal{E} \), train \( \hat{f} \) on \( \mathcal{E}_{\text{syn}} \), and evaluate performance on \( \mathcal{V} \).  
This process is repeated 200 times, and the average \( \text{Error}_g \) values across different prediction horizons \( g \) are reported in Figure~\ref{fig: real_prediction}~(B).

Figure~\ref{fig: real_prediction}~(B) shows that SFM achieves the lowest \( \text{Error}_g \) among all generative methods, indicating that it produces synthetic data with superior predictive power. 
Notably, its prediction error is comparable to that obtained from the observed data and even lower when the sample size is smaller (e.g., arterial blood pressure with \( n = 256 \), compared to \( n = 1446 \) for other features). 
These results suggest that SFM can generate reliable, densely and regularly sampled functional data, thereby reducing prediction uncertainty---particularly in settings with limited or irregularly observed trajectories.

\section{Discussions}\label{sec:discussions}

In this article, we introduce a novel smooth flow matching (SFM) method for generating new functional samples from sparsely and irregularly observed functional data. 
Our approach proposes a semiparametric framework that learns general marginal distributions of random functions while enforcing smoothness in the synthesized samples. 
It offers an interpretable, flexible, and computationally efficient data generator that does not rely on assumptions such as Gaussianity, low-rank structure, or dense and regular observations, nor does it require computationally intensive operator learning. 
Simulation studies and real data analysis demonstrate the superior performance of SFM in terms of the quality of generated samples and computational efficiency, providing high-quality surrogate datasets that preserve both the temporal dynamics and predictive power of the underlying data under privacy constraints.

Generative modeling for unaligned data, such as the functional data considered in this article, is an important topic in synthetic data analysis. In finite-dimensional tabular settings, unalignedness can arise due to missing data \citep{givens2025score}, and our method may still be applicable in this setting if a meaningful notion of ``neighborhood'' across tabular features can be defined. This neighborhood structure is crucial for smooth-flow-type matching, as it allows us to borrow information across neighboring features for estimation even under unaligned observations.
Furthermore, this article mainly focuses on univariate functional data generation from irregular observations. Extending the SFM framework to irregularly observed multivariate functional data is also of great interest in many real-world applications, such as EHR data \citep{johnson2020mimic}, spatio-temporal data \citep{tan2024graphical}, and epidemiological data \citep{luo2024functional}, where capturing interdependencies among random functions is crucial. We leave these directions for future investigation.

\bibliographystyle{apalike}

\bibliography{refbib}

\appendix

\section{Theoretical Proofs}

\subsection{Proof of Proposition~\ref{Pro_copula}}
\begin{proof}
The distributions of \(X(t)\) and \(Z(t)\) are denoted by \(F_t\) and \(F_{t,\text{base}}\), respectively.
Define \( g_t(x) = F_t^{-1} \circ F_{t,\text{base}}(x) \). 
Thus, \( g_t \) is a continuous and strictly increasing function from \({\operatorname{supp}\{Z(t)\}}\) to \({\operatorname{supp}\{X(t)\}}\).

If \( X(\cdot) \) is a copula process with base process \( Z(\cdot) \), define
\[
\tilde{Z}(t) = F^{-1}_{t,\text{base}}\!\big(F_t(X(t))\big).
\]
For any \(t_1,\ldots,t_m\in \mathcal{S}\) and \(x_1\in {\operatorname{supp}\{Z(t_1)\}},\ldots, x_m\in {\operatorname{supp}\{Z(t_m)\}}\),
\begin{eqnarray*}
   && \mathbb{P}\big\{ \tilde{Z}(t_1)\leq x_1, \ldots,  \tilde{Z}(t_m)\leq x_m\big\}\\
    &=& \mathbb{P}\big\{F_{t_1}(X(t_1))\leq F_{t_1,\text{base}}(x_1), \ldots,  F_{t_m}(X(t_m))\leq F_{t_m,\text{base}}(x_m)\big\}\\
    &=& c_{t_1,\ldots,t_m}\big(F_{t_1,\text{base}}(x_1),\ldots,F_{t_m,\text{base}}(x_m)\big)\\
    &=& H_{\text{base}}\big(x_1,\ldots,x_m; t_1,\ldots,t_m\big)\\
    &=& \mathbb{P}\big\{ Z(t_1)\leq x_1, \ldots,  Z(t_m)\leq x_m\big\}.
\end{eqnarray*}
Then the stochastic processes \( Z(\cdot) \) and \( \tilde{Z}(\cdot) \) are equal in distribution.

Notice that
\begin{eqnarray*}
    g_t\big(\tilde{Z}(t)\big)
    &=& F_t^{-1} \circ F_{t,\text{base}} \left( F^{-1}_{t,\text{base}} \left( F_t(X(t)) \right) \right) \\
    &=& F_t^{-1} \circ F_t\big(X(t)\big) = X(t), \quad t \in \mathcal{S}.
\end{eqnarray*}
Thus, \( X(t) = g_t(\tilde{Z}(t)) \), almost surely, for all \( t \in \mathcal{S} \). 
Therefore, \(\{X(t)\}_{t\in \mathcal{S}}\) and \(\{g_t\big(\tilde{Z}(t)\big)\}_{t\in \mathcal{S}}\) are equal in distribution.
Since \(\tilde{Z}(\cdot)\overset{d}{=}Z(\cdot)\), it follows that \(\{X(t)\}_{t\in \mathcal{S}}\) and \(\{g_t(Z(t))\}_{t\in \mathcal{S}}\) are equal in distribution.

Conversely, assume that there exists a process \( \tilde{Z}(\cdot)\) with the same distribution as \(Z(\cdot)\), and for each \( t \in \mathcal{S} \) there exists a continuous and strictly increasing function \( g_t \) such that
\( X(t) = g_t\big(\tilde{Z}(t)\big) \), \( t \in \mathcal{S} \).
We first claim that
\begin{eqnarray*}
    \tilde{Z}(t)=F_{t,\text{base}}^{-1}\!\big(F_t(X(t))\big),\ \text{almost surely},\ \forall\, t\in \mathcal{S}.
\end{eqnarray*}
This can be {shown as follows. Since \(X(t)=g_t(\tilde Z(t))\) and \(g_t\) is strictly increasing, we have \(\tilde Z(t)=g_t^{-1}(X(t))\). Moreover, because \(\tilde Z(t)\) has distribution \(F_{t,\text{base}}\), the random variable \(g_t^{-1}(X(t))\) also has distribution \(F_{t,\text{base}}\). On the other hand, \(F_{t,\text{base}}^{-1}\!\big(F_t(X(t))\big)\) has distribution \(F_{t,\text{base}}\) as well.}
Since \(g_t^{-1}\) and \(F_{t,\text{base}}^{-1}\circ F_t\) are both continuous and strictly increasing functions on \(x\in {\operatorname{supp}\{X(t)\}}\) for all \(t\in \mathcal{S}\),
Lemma~\ref{variable_equa} implies that \(g_t^{-1}(X(t))=F_{t,\text{base}}^{-1}\!\big(F_t(X(t))\big)\) almost surely. The claim holds.

Notice that for any \(t_1,\ldots,t_m\in \mathcal{S}\) and \(u_1,\ldots,u_m\in [0,1]\),
\begin{eqnarray*}
&&c_{t_1,\ldots,t_m}(u_1,\ldots,u_m)\\
&=& \mathbb{P}\big(F_{t_1}(X(t_1)) \leq u_1, \ldots, F_{t_m}(X(t_m)) \leq u_m\big)\\
&=&\mathbb{P}\big(F_{t_1,\text{base}}^{-1}\!\circ F_{t_1}(X(t_1)) \leq F_{t_1,\text{base}}^{-1}(u_1), \ldots, F_{t_m,\text{base}}^{-1}\!\circ F_{t_m}(X(t_m)) \leq F_{t_m,\text{base}}^{-1}(u_m)\big)\\
&=& \mathbb{P}\big\{ \tilde{Z}(t_1)\leq F_{t_1,\text{base}}^{-1}(u_1), \ldots,  \tilde{Z}(t_m)\leq F_{t_m,\text{base}}^{-1}(u_m)\big\}\\
   &=& \mathbb{P}\big\{ Z(t_1)\leq F_{t_1,\text{base}}^{-1}(u_1), \ldots,  Z(t_m)\leq F_{t_m,\text{base}}^{-1}(u_m)\big\}.
\end{eqnarray*}
Then \(X(\cdot)\) is a copula process with base process \(Z(\cdot)\).
This completes the proof.
\end{proof}

\

\subsection{Proof of Theorem~\ref{inverse_formula}}

\begin{proof}
Part (a) is a direct result of Lemma~\ref{le:PL}. 
We now prove parts (b), (c), and (d).

\medskip
\noindent\textbf{(b)}
We first claim that, for any fixed \(t\in\mathcal S\) and \(u\in[0,1]\), the map \(x\mapsto \phi_{u,t}(x)\) is strictly increasing. 
In particular, \(\phi_{{u},t}(\cdot)\) is injective and hence a bijection onto its image.

Suppose for contradiction that \(x_1<x_2\) but \(\phi_{{u},t}(x_1)\ge \phi_{{u},t}(x_2)\).
Since \(r\mapsto \phi_{r,t}(x)\) is continuous and
\[
\phi_{0,t}(x_1)=x_1<x_2=\phi_{0,t}(x_2),
\]
the intermediate value theorem implies that there exists a first time \(u_0\in(0,{u}]\) such that the trajectories meet:
\[
\phi_{u_0,t}(x_1)=\phi_{u_0,t}(x_2).
\]
By Lemma~\ref{le:PL}, two solutions that coincide at time \(u_0\) must coincide for all \(r\in[0,u_0]\); in particular,
\(
\phi_{0,t}(x_1)=\phi_{0,t}(x_2)
\),
i.e., \(x_1=x_2\), which is a contradiction.
Therefore \(\phi_{{u},t}(x_1)<\phi_{{u},t}(x_2)\), and the claim follows.

Next, we show that \(\phi_{u,t}\) is {invertible on its image, and that both \(\phi_{u,t}\) and \(\phi_{u,t}^{-1}\) are differentiable for Lebesgue-a.e.\ points in their respective domains,} for each \(u\in[0,1]\) and \(t\in\mathcal S\), and that its inverse satisfies~\eqref{inver_flow_def}.

\begin{itemize}
    \item Fix \(t\in\mathcal S\). Since \(\phi_{1,t}\) is strictly increasing, \(\phi_{1,t}^{-1}\) is well-defined on the image set \(\{\phi_{1,t}(x):x\in\mathbb R\}\). We next prove that \(x\mapsto \phi_{u,t}(x)\) is Lipschitz and hence differentiable a.e.\ in \(x\). Using the integral form of the flow,
\[
\phi_{u,t}(x)=x+\int_0^u V_{s,t}\big(\phi_{s,t}(x)\big)\,\mathrm{d}s,
\]
fix \(t\in\mathcal S\), \(x\in\mathbb R\), and \(h\neq 0\). Then
\begin{eqnarray*}
    \bigg|\frac{\phi_{u,t}(x+h)-\phi_{u,t}(x)}{h}\bigg|
&=&
\bigg|1+\int_0^u
\frac{V_{s,t}(\phi_{s,t}(x+h))-V_{s,t}(\phi_{s,t}(x))}{h}\,\mathrm{d}s\bigg|\\
&\leq & 1 +  L\cdot\int_0^u
\bigg|\frac{\phi_{s,t}(x+h)-\phi_{s,t}(x)}{h}\bigg|\,\mathrm{d}s,
\end{eqnarray*}
where we used the Lipschitz condition on \(V_{s,t}(\cdot)\) in \(x\).
Applying Gr\"onwall's inequality, we have
\begin{eqnarray}\label{Gr_inequality}
    \bigg|\frac{\phi_{u,t}(x+h)-\phi_{u,t}(x)}{h}\bigg| \leq  e^{L u}.
\end{eqnarray}
Thus, for each fixed \((u,t)\), the map \(x\mapsto \phi_{u,t}(x)\) is Lipschitz continuous, and hence absolutely continuous. By Rademacher's theorem, \(\partial_x\phi_{u,t}(x)\) exists for Lebesgue-a.e.\ \(x\in\mathbb R\).

\item We also have a lower Lipschitz bound. For \(x_1<x_2\), define \(d(u):=\phi_{u,t}(x_2)-\phi_{u,t}(x_1)>0\).
Using \eqref{flow_def} and the Lipschitz property of \(V_{u,t}\),
\[
d'(u)=V_{u,t}\big(\phi_{u,t}(x_2)\big)-V_{u,t}\big(\phi_{u,t}(x_1)\big)\ge -L\,d(u),
\]
so Gr\"onwall's inequality yields
\[
\phi_{u,t}(x_2)-\phi_{u,t}(x_1)=d(u)\ge e^{-Lu}(x_2-x_1).
\]
Consequently, for a.e.\ \(x\),
\[
\partial_x\phi_{u,t}(x)\geq e^{-Lu},
\]
and hence \(\partial_x\phi_{u,t}(x)\neq 0\) for Lebesgue-a.e.\ \(x\in\mathbb R\).

\item Since \(\phi_{u,t}\) is strictly increasing, it is injective and admits an inverse \(\phi_{u,t}^{-1}\) on the image set \(\{\phi_{u,t}(x):x\in\mathbb R\}\). Moreover, at points where \(\partial_x\phi_{u,t}(x)\) exists and is nonzero, the inverse function theorem implies that \(\phi_{u,t}^{-1}\) is differentiable at \(z=\phi_{u,t}(x)\) and satisfies
\[
\partial_z \phi_{u,t}^{-1}(z)=\frac{1}{\partial_x\phi_{u,t}(x)}\quad \text{for Lebesgue-a.e.\ }z\in \{\phi_{u,t}(x):x\in\mathbb R\}.
\]
In particular, \(\phi_{u,t}\) is invertible and both \(\phi_{u,t}\) and \(\phi_{u,t}^{-1}\) are differentiable Lebesgue-a.e.\ with nonzero derivatives.

\item  Define, for \(z\in\{\phi_{1,t}(x):x\in\mathbb R\}\),
\[
\psi_{u,t}(z):=\phi_{1-u,t}\big(\phi_{1,t}^{-1}(z)\big),\qquad u\in[0,1].
\]
Then \(\psi_{0,t}(z)=z\) and \(\psi_{1,t}(z)=\phi_{1,t}^{-1}(z)\). Moreover, by~\eqref{flow_def},
\begin{eqnarray*}
\frac{\partial}{\partial u}\,\psi_{u,t}(z)
&=&
\frac{\partial}{\partial u}\,\phi_{1-u,t}\big(\phi_{1,t}^{-1}(z)\big)
=
-\,V_{1-u,t}\!\left(\phi_{1-u,t}\big(\phi_{1,t}^{-1}(z)\big)\right)\\
&=&
-\,V_{1-u,t}\big(\psi_{u,t}(z)\big),
\end{eqnarray*}
which is exactly~\eqref{inver_flow_def}. This proves the inverse-flow representation {for \(\phi_{1,t}^{-1}\)} in (b).
\end{itemize}

(c) Fix \(t\in\mathcal S\). Let
\[
P_t(z):=F_t^{-1}\circ F_{t,\text{base}}(z),\qquad z\in {\operatorname{supp}}\!\big(Z_0(t)\big).
\]
Since \(F_{t,\text{base}}\) and \(F_t\) are continuous distribution functions, \(P_t\) is strictly increasing and continuous. Moreover,
\(P_t\big(Z_0(t)\big)\) has marginal distribution \(F_t\), and by \eqref{equ_flow} we also have
\(\phi_{1,t}\big(Z_0(t)\big)\overset{d}{=}Z_1(t)\), whose marginal distribution is \(F_t\). Hence
\[
\phi_{1,t}\big(Z_0(t)\big)\ \overset{d}{=}\ P_t\big(Z_0(t)\big).
\]
Since \(\phi_{1,t}(\cdot)\) and \(P_t(\cdot)\) are both strictly increasing and continuous, Lemma~\ref{variable_equa} yields
\[
\phi_{1,t}\big(Z_0(t)\big)=P_t\big(Z_0(t)\big)\quad \text{a.s.}
\]
Because \(Z_0(t)\) is a continuous random variable, Lemma~\ref{le: continu_inver} further implies that
\[
\phi_{1,t}(z)=F_t^{-1}\circ F_{t,\text{base}}(z),
\quad \text{for Lebesgue-a.e.\ } z\in {\operatorname{supp}}\!\big(Z_0(t)\big).
\]

Next, define
\[
Q_t(x):=F_{t,\text{base}}^{-1}\circ F_t(x),\qquad x\in {\operatorname{supp}}\!\big(Z_1(t)\big).
\]
For any {\(z\in \operatorname{supp}\!\big(Z_0(t)\big)\)},
\[
\mathbb P\!\left(\phi_{1,t}^{-1}\big(Z_1(t)\big)\le z\right)
=
\mathbb P\!\left(Z_1(t)\le \phi_{1,t}(z)\right)
=
{\mathbb P\!\left(\phi_{1,t}\big(Z_0(t)\big)\le \phi_{1,t}(z)\right)}
=
\mathbb P\!\left(Z_0(t)\le z\right),
\]
so \(\phi_{1,t}^{-1}\big(Z_1(t)\big)\overset{d}{=}Z_0(t)\). On the other hand,
\(Q_t\big(Z_1(t)\big)\overset{d}{=}Z_0(t)\) by the probability integral transform. Therefore,
\[
\phi_{1,t}^{-1}\big(Z_1(t)\big)\ \overset{d}{=}\ Q_t\big(Z_1(t)\big).
\]
Since \(\phi_{1,t}^{-1}(\cdot)\) and \(Q_t(\cdot)\) are strictly increasing and continuous on \({\operatorname{supp}\!\big(Z_1(t)\big)}\),
Lemma~\ref{variable_equa} gives
\[
\phi_{1,t}^{-1}\big(Z_1(t)\big)=Q_t\big(Z_1(t)\big)\quad \text{a.s.}
\]
and Lemma~\ref{le: continu_inver} implies
\[
\phi_{1,t}^{-1}(x)=F_{t,\text{base}}^{-1}\circ F_t(x),
\quad \text{for Lebesgue-a.e.\ } x\in {\operatorname{supp}}\!\big(Z_1(t)\big).
\]

(d) Fix any \(m\ge 1\) and \(t_1,\ldots,t_m\in\mathcal S\). For any
\(x_\ell\in{\operatorname{supp}}\!\big(X(t_\ell)\big)\), \(\ell=1,\ldots,m\), we have
\begin{eqnarray*}
&&\mathbb P\big(X(t_1)\le x_1,\ldots,X(t_m)\le x_m\big)\\
&=&\mathbb P\Big(F_{t_1}\!\big(X(t_1)\big)\le F_{t_1}(x_1),\ldots,
F_{t_m}\!\big(X(t_m)\big)\le F_{t_m}(x_m)\Big)\\
&=&c_{t_1,\ldots,t_m}\big(F_{t_1}(x_1),\ldots,F_{t_m}(x_m)\big).
\end{eqnarray*}
Since \(X\) is a copula process with base \(Z_0\), Definition~\ref{Def_copula_process} implies
\begin{eqnarray*}
&& c_{t_1,\ldots,t_m}\big(F_{t_1}(x_1),\ldots,F_{t_m}(x_m)\big)\\
&=&
H_{\text{base}}\!\Big(
F_{t_1,\text{base}}^{-1}\!\circ F_{t_1}(x_1),\ldots,
F_{t_m,\text{base}}^{-1}\!\circ F_{t_m}(x_m);
t_1,\ldots,t_m\Big)\\
&=&
\mathbb P\Big(
Z_0(t_1)\le F_{t_1,\text{base}}^{-1}\!\circ F_{t_1}(x_1),\ldots,
Z_0(t_m)\le F_{t_m,\text{base}}^{-1}\!\circ F_{t_m}(x_m)
\Big),
\end{eqnarray*}
where \(H_{\text{base}}(\cdot;t_1,\ldots,t_m)\) denotes the joint distribution function of
\(\big(Z_0(t_1),\ldots,Z_0(t_m)\big)\).

By Theorem~\ref{inverse_formula}(c), for each \(\ell=1,\ldots,m\),
\(
F_{t_\ell,\text{base}}^{-1}\circ F_{t_\ell}(x_\ell)=\psi_{1,t_\ell}(x_\ell)
=\phi_{1,t_\ell}^{-1}(x_\ell)
\)
for Lebesgue-a.e.\ \(x_\ell\in{\operatorname{supp}}\!\big(X(t_\ell)\big)\).
Since \(\phi_{1,t_\ell}\) is strictly increasing, the event
\(\{Z_0(t_\ell)\le \phi_{1,t_\ell}^{-1}(x_\ell)\}\) is equivalent to
\(\{\phi_{1,t_\ell}(Z_0(t_\ell))\le x_\ell\}\). Therefore,
\begin{eqnarray*}
&&\mathbb P\Big(
Z_0(t_1)\le F_{t_1,\text{base}}^{-1}\!\circ F_{t_1}(x_1),\ldots,
Z_0(t_m)\le F_{t_m,\text{base}}^{-1}\!\circ F_{t_m}(x_m)
\Big)\\
&=&
\mathbb P\Big(
\phi_{1,t_1}\big(Z_0(t_1)\big)\le x_1,\ldots,
\phi_{1,t_m}\big(Z_0(t_m)\big)\le x_m
\Big).
\end{eqnarray*}
Finally, by \eqref{equ_flow} with \(u=1\), \(\phi_{1,t}(Z_0(t))=Z_1(t)\) for each \(t\in\mathcal S\).
Hence,
\[
\mathbb P\big(X(t_1)\le x_1,\ldots,X(t_m)\le x_m\big)
=
\mathbb P\big(Z_1(t_1)\le x_1,\ldots,Z_1(t_m)\le x_m\big),
\]
which shows that \(\{X(t)\}_{t\in\mathcal S}\) and \(\{Z_1(t)\}_{t\in\mathcal S}\) have the same
finite-dimensional distributions, and therefore \(X(\cdot)\overset{d}{=}Z_1(\cdot)\).

\end{proof}

\

\subsection{Proof of Theorem~\ref{theo: smooth}}

\begin{proof}
Denote {\(V_{u,t}(x):=V(u,t,x)\).}
We claim that if \(h\in \mathcal W^2_q(\mathcal{S})\), then
\begin{eqnarray}\label{equ:vec}
&&    \frac{\mathrm{d}^q}{\mathrm{d} t^q} V\big(u, t, h(t)\big) \\
    =\sum_{m=0}^{q}&&\sum_{l=0}^{q-m}
    \sum_{\substack{k_1 + 2k_2 + \cdots + (q-m)k_{q-m} = q-m \\ k_1 + k_2 + \cdots + k_{q-m} = l}}
    C_{k_1,\ldots,k_{q-m},m,l}\,
    \frac{\partial^{l+m} V}{\partial t^{m} \partial x^{l}}\big(u,t,h(t)\big)\,
    \prod_{j=1}^{q-m} \big(h^{(j)}(t)\big)^{k_j},\nonumber
\end{eqnarray}
where \(C_{k_1,\ldots,k_{q-m},m,l}\) are some constants.

We first evaluate the formula for \( q = 1 \), which is
\begin{eqnarray*}
    \frac{\mathrm{d}}{\mathrm{d} t} V\big(u, t, h(t)\big)
    &=& \frac{\partial V}{\partial t}\big(u,t,h(t)\big)
      + \frac{\partial V}{\partial x}\big(u,t,h(t)\big)\, h'(t)\\
&=&C_{1,0,1}\,\frac{\partial V}{\partial x}\big(u,t,h(t)\big)\, h'(t)
  +C_{\emptyset,1,0}\,\frac{\partial V}{\partial t}\big(u,t,h(t)\big),
\end{eqnarray*}
where \( C_{1,0,1} = C_{\emptyset,1,0} = 1 \).

We then assume \eqref{equ:vec} holds for \(q=n\).
For \(q=n+1\), consider a generic term in \eqref{equ:vec} (with \(q=n\)):
\[
\frac{\partial^{l+m} V}{\partial t^m \partial x^l}\big(u,t,h(t)\big)\,
\prod_{j=1}^{n-m}\big(h^{(j)}(t)\big)^{k_j},
\]
and compute its derivative with respect to \(t\):
\[
\frac{\mathrm{d}}{\mathrm{d} t}\left(
\frac{\partial^{l+m} V}{\partial t^m \partial x^l}\big(u,t,h(t)\big)\,
\prod_{j=1}^{n-m}\big(h^{(j)}(t)\big)^{k_j}
\right).
\]
Define
\[
A(t)=\frac{\partial^{l+m} V}{\partial t^m \partial x^l}\big(u,t,h(t)\big),
\qquad
B(t)=\prod_{j=1}^{n-m}\big(h^{(j)}(t)\big)^{k_j}.
\]
By the product rule,
\[
\frac{\mathrm{d}}{\mathrm{d} t}\big\{A(t)B(t)\big\}
= A'(t)B(t)+A(t)B'(t).
\]
For \(A'(t)\), by the chain rule,
\begin{eqnarray*}
A'(t)
&=&
\frac{\partial^{l+m+1} V}{\partial t^{m+1}\partial x^{l}}\big(u,t,h(t)\big)
+
{\frac{\partial^{l+m+1} V}{\partial t^{m} \partial x^{l+1}}}\big(u,t,h(t)\big)\,h'(t).
\end{eqnarray*}
For \(B'(t)\), {by differentiating} one factor at a time,
\begin{eqnarray*}
B'(t)
&=&
\sum_{p=1}^{n-m}
\left\{\frac{\mathrm{d}}{\mathrm{d} t}\big(h^{(p)}(t)\big)^{k_p}\right\}
\prod_{j\neq p}\big(h^{(j)}(t)\big)^{k_j}\\
&=&
\sum_{p=1}^{n-m}
k_p\,\big(h^{(p)}(t)\big)^{k_p-1}\,h^{(p+1)}(t)\,
\prod_{j\neq p}\big(h^{(j)}(t)\big)^{k_j}.
\end{eqnarray*}
Combining \(A'(t)\) and \(B'(t)\), we obtain
\begin{eqnarray*}
&&\frac{\mathrm{d}}{\mathrm{d} t}\left(
\frac{\partial^{l+m} V}{\partial t^m \partial x^l}\big(u,t,h(t)\big)\,
\prod_{j=1}^{n-m}\big(h^{(j)}(t)\big)^{k_j}
\right)\\
&=&
\left\{
{\frac{\partial^{l+m+1} V}{\partial t^{m+1} \partial x^{l}}}\big(u,t,h(t)\big)
+
{\frac{\partial^{l+m+1} V}{\partial t^{m} \partial x^{l+1}}}\big(u,t,h(t)\big)\,h'(t)
\right\}
\prod_{j=1}^{n-m}\big(h^{(j)}(t)\big)^{k_j}\\
&&\quad+
\frac{\partial^{l+m} V}{\partial t^m \partial x^l}\big(u,t,h(t)\big)
\sum_{p=1}^{n-m}
k_p\,\big(h^{(p)}(t)\big)^{k_p-1}\,h^{(p+1)}(t)\,
\prod_{j\neq p}\big(h^{(j)}(t)\big)^{k_j}.
\end{eqnarray*}
In the following, we abbreviate $\frac{\partial^{l+m} V}{\partial t^{m} \partial x^{l}}(u,t,h(t))$ as $\frac{\partial^{l+m} V}{\partial t^{m} \partial x^{l}}$. We then have
\begin{eqnarray*}
  &&  \frac{\mathrm{d}^{n+1}}{\mathrm{d} t^{n+1}} V(u, t, h(t)) \\
    &=&  \sum_{m=0}^{n} \sum_{l=0}^{n-m} \sum_{\substack{k_1 + 2k_2 + \cdots + (n-m)k_{n-m} = n - m \\ k_1 + k_2 + \cdots + k_{n-m} = l}} C_{k_1,\ldots,k_{n-m},m,l} \frac{\mathrm{d}}{\mathrm{d} t} \left( 
    \frac{\partial^{l+m} V}{\partial t^{m} \partial x^{l}}\cdot
    \prod_{j=1}^{n-m} (h^{(j)}(t))^{k_j} \right)\\
    &=&\sum_{m=0}^{n} \sum_{l=0}^{n-m} \sum_{\substack{k_1 + 2k_2 + \cdots + (n-m)k_{n-m} = n - m \\ k_1 + k_2 + \cdots + k_{n-m} = l}} C_{k_1,\ldots,k_{n-m},m,l} \bigg\{\bigg( 
    \frac{\partial^{l+m+1} V}{\partial t^{m+1} \partial x^l}+  
    \frac{\partial^{l+m+1} V}{\partial t^m \partial x^{l+1}}\\
    &\cdot&
    h'(t) \bigg)\cdot\prod_{j=1}^{n-m} (h^{(j)}(t))^{k_j}\\
    &&+
    \frac{\partial^{l+m} V}{\partial t^m \partial x^l}\cdot
    \sum_{p=1}^{n-m} k_p (h^{(p)}(t))^{k_p - 1} h^{(p+1)}(t) \prod_{j \neq p} (h^{(j)}(t))^{k_j}\bigg\}\\
     &=&\sum_{m=0}^{n} \sum_{l=0}^{n-m} \sum_{\substack{k_1 + 2k_2 + \cdots + (n-m)k_{n-m} = n - m \\ k_1 + k_2 + \cdots + k_{n-m} = l}} C_{k_1,\ldots,k_{n-m},m,l} 
     \frac{\partial^{l+m+1} V}{\partial t^{m+1} \partial x^l}\cdot
     \prod_{j=1}^{n-m} (h^{(j)}(t))^{k_j} \\
    &&+ \sum_{m=0}^{n} \sum_{l=0}^{n-m} \sum_{\substack{k_1 + 2k_2 + \cdots + (n-m)k_{n-m} = n - m \\ k_1 + k_2 + \cdots + k_{n-m} = l}} C_{k_1,\ldots,k_{n-m},m,l} 
    \frac{\partial^{l+m+1} V}{\partial t^m \partial x^{l+1}}\cdot
    h'(t)  \\
    &\cdot &\prod_{j=1}^{n-m} (h^{(j)}(t))^{k_j}+ \sum_{m=0}^{n} \sum_{l=0}^{n-m} \sum_{\substack{k_1 + 2k_2 + \cdots + (n-m)k_{n-m} = n - m \\ k_1 + k_2 + \cdots + k_{n-m} = l}} C_{k_1,\ldots,k_{n-m},m,l} \\
    &&\cdot
    \frac{\partial^{l+m} V}{\partial t^m \partial x^l}\cdot
    \sum_{p=1}^{n-m} k_p (h^{(p)}(t))^{k_p - 1} h^{(p+1)}(t) \prod_{j \neq p} (h^{(j)}(t))^{k_j}\\
    &=&\sum_{m=0}^{n+1}\sum_{l=0}^{n+1-m}
    \sum_{\substack{k_1+2k_2+\cdots+(n+1-m)k_{n+1-m}=n+1-m\\
    k_1+k_2+\cdots+k_{n+1-m}=l}}
    \tilde C_{k_1,\ldots,k_{n+1-m},m,l}\,
    \frac{\partial^{l+m} V}{\partial t^{m} \partial x^{l}}\\
  &\cdot&  \prod_{j=1}^{n+1-m}(h^{(j)}(t))^{k_j},
\end{eqnarray*}
where \(\tilde C_{k_1,\ldots,k_{n+1-m},m,l}\) are some constants, since each term above is still of the same form.
Therefore, $\frac{\mathrm{d}^{n+1}}{\mathrm{d} t^{n+1}} V(u, t, h(t))$ can be represented by
\begin{eqnarray*}
   \sum_{m=0}^{n+1}\sum_{l=0}^{n+1-m} \sum_{\substack{k_1 + 2k_2 + \cdots + (n+1-m)k_{n+1-m} = n+1-m \\ k_1 + k_2 + \cdots + k_{n+1-m} = l}} \tilde{C}_{k_1,\ldots,k_{n+1-m},m,l} 
   \frac{\partial^{l+m} V}{\partial t^{m} \partial x^{l}}
   \prod_{j=1}^{n+1-m} (h^{(j)}(t))^{k_j},
\end{eqnarray*}
where 
$\tilde{C}_{k_1,\ldots,k_{n+1-m},m,l}$ are some constants determined by $C_{k_1,\ldots,k_{n-m},m,l}$.
{This proves the claim.}

Note that $\frac{\partial^{l+m} V}{\partial t^{m} \partial x^{l}}\in \mathcal{L}^2(\mathcal{S})$, $h^{(q)}\in \mathcal{L}^2(\mathcal{S})$, the exponent of the highest-order derivative term \(h^{(q)}\) in \eqref{equ:vec} is always 0 or 1, and $h^{(j)}$ is a continuous function for $j<q$, we have 
$\frac{\mathrm{d}^q}{\mathrm{d} t^q} V(u, t, h(t))$ is a function of $t$ 
belonging to
$\mathcal{L}^2(\mathcal{S})$. Therefore, 
\( V_{u,\cdot}(h(\cdot)) \in \mathcal W^2_q(\mathcal{S}) \) whenever \( h(\cdot) \in \mathcal W^2_q(\mathcal{S}) \).

Under conditions~\eqref{smooth_condition} and \eqref{Lipschitz_condition}, and \(Z_0 \in \mathcal W^2_q(\mathcal{S})\),
the solution \(Z_u\) to \eqref{ode_ge} lies in 
\(\mathcal W^2_q(\mathcal{S})\)
by Lemma~\ref{le:PL}.

\end{proof}

The following proposition is a direct result of Theorem~\ref{theo: smooth}.

\begin{PropositionS}[Runge--Kutta Method for Smooth Function Generation]\label{theo_smooth_numerical}
Let $\{u_1,\ldots,u_M\}$ be an equally spaced sequence of $[0,1]$ with a sufficiently small step size $\Delta u$. 
Given $Z_0(t)$ and $V_{u,t}$, define 
\begin{eqnarray}\label{RK4}
  Z_{u_{m+1}}(t)
  =
  Z_{u_{m}}(t)
  +\frac{\Delta u}{6}\big\{k_{1,m}(t)+2k_{2,m}(t)+2k_{3,m}(t)+k_{4,m}(t)\big\},
\end{eqnarray}
$t\in \mathcal{S}$,
where $k_{1,m}(t)=V_{u_m,t}(Z_{u_m}(t))$, 
$k_{2,m}(t)=V_{u_m+\Delta u /2,t}\big(Z_{u_m}(t)+\Delta u\cdot k_{1,m}(t)/2\big)$, 
$k_{3,m}(t)=V_{u_m+\Delta u /2,t}\big(Z_{u_m}(t)+\Delta u\cdot k_{2,m}(t)/2\big)$, and 
$k_{4,m}(t)=V_{u_m+\Delta u,t}\big(Z_{u_m}(t)+\Delta u\cdot k_{3,m}(t)\big)$. 
Suppose that 
$Z_0\in \mathcal W^2_q(\mathcal{S})$
and the vector field $V_{u,t}$ satisfies \eqref{smooth_condition}. Then
$$
Z_{u_{m}}\in \mathcal W^2_q(\mathcal{S}),
\quad m=1,\ldots,M.
$$
\end{PropositionS}

\

\subsection{Proof of Theorem~\ref{Theo: cond}}

\begin{proof}

We first show that \(\tilde Z_u(t)\), generated by the vector field \( V_{u,t} \) with initial condition \(\tilde Z_0(t) \sim p_{0,t} \), has the probability density \( p_{u,t} \). 
Our task is to prove that \( p_{u,t}(x) \) satisfies the continuity equation associated with the vector field \(V_{u,t}\), i.e.,
\[
\frac{\partial p_{u,t}(x)}{\partial u} 
= 
- \frac{\partial}{\partial x} \Big( p_{u,t}(x)\, V_{u,t}(x) \Big).
\]
This {implies that} \(p_{u,t}\) is the density of \({\tilde Z_u(t)} \) for each \(t\in\mathcal S\).

Notice that
\[
p_{u,t}(x)=\int_{\mathbb R} q_{u,t}(x \mid y)\, f_X(y; t) \, \mathrm{d}y,
\]
Hence
\[
\frac{\partial p_{u,t}(x)}{\partial u} 
= 
\int_{\mathbb R} \frac{\partial}{\partial u} q_{u,t}(x \mid y)\, f_X(y; t) \, \mathrm{d}y.
\]
Since \( V_{u,t}(\cdot \mid y) \) generates \( q_{u,t}(\cdot \mid y) \), \(q_{u,t}(\cdot \mid y)\) satisfies the continuity equation associated with the vector field \(V_{u,t}(\cdot\mid y)\):
\[
\frac{\partial q_{u,t}(x \mid y)}{\partial u} 
= 
- \frac{\partial}{\partial x} \Big( q_{u,t}(x \mid y)\, V_{u,t}(x \mid y) \Big).
\]
Thus,
\begin{eqnarray}\label{p_ut_1}
\frac{\partial p_{u,t}(x)}{\partial u} 
&=& 
\int_{\mathbb R} \left[ - \frac{\partial}{\partial x} \Big( q_{u,t}(x \mid y)\, V_{u,t}(x \mid y) \Big) \right] f_X(y; t) \, \mathrm{d}y \nonumber\\
&=& 
-\frac{\partial}{\partial x}\int_{\mathbb R} q_{u,t}(x \mid y)\, V_{u,t}(x \mid y)\, f_X(y; t) \, \mathrm{d}y.
\end{eqnarray}

Meanwhile, recall that
\[
V_{u,t}(x) 
= 
\int_{\mathbb{R}} V_{u,t}(x \mid y)\cdot 
\frac{q_{u,t}(x \mid y)\, f_X(y; t)}{p_{u,t}(x)} \, \mathrm{d}y,
\qquad \text{for } x\in \operatorname{supp}(p_{u,t}).
\]
Multiplying both sides by \(p_{u,t}(x)\) yields
\[
p_{u,t}(x)\,V_{u,t}(x)
=
\int_{\mathbb{R}} V_{u,t}(x \mid y)\, q_{u,t}(x \mid y)\, f_X(y; t)\, \mathrm{d}y,
\qquad \text{for } x\in \operatorname{supp}(p_{u,t}).
\]
Moreover, since \(q_{u,t}(x\mid y)f_X(y;t)\ge 0\) and
\[
\int_{\mathbb R} q_{u,t}(x \mid y)\, f_X(y; t)\, \mathrm{d}y = 0,
\]
for \(x\notin \operatorname{supp}(p_{u,t})\), {it follows that}
\[
q_{u,t}(x\mid y)\, f_X(y;t)=0 \quad \text{for a.e.\ }y,\ \text{whenever } x\notin \operatorname{supp}(p_{u,t}).
\]
Therefore, the identity
\[
p_{u,t}(x)\,V_{u,t}(x)
=
\int_{\mathbb R} V_{u,t}(x \mid y)\, q_{u,t}(x \mid y)\, f_X(y; t)\, \mathrm{d}y
\]
holds for all \(x\in\mathbb R\), and thus
\[
- \frac{\partial}{\partial x} \Big( p_{u,t}(x)\, V_{u,t}(x) \Big)
=
- \frac{\partial}{\partial x} 
\int_{\mathbb R} V_{u,t}(x \mid y)\, q_{u,t}(x \mid y)\, f_X(y; t) \, \mathrm{d}y.
\]
Combining this with \eqref{p_ut_1}, we obtain
\[
\frac{\partial p_{u,t}(x)}{\partial u} 
=
- \frac{\partial}{\partial x} \Big( p_{u,t}(x)\, V_{u,t}(x) \Big),
\]
as desired.

Since the vector field \(V_{u,t}\) is Lipschitz in \(x\), the solution \(\tilde Z_u\) to the ODE is unique. Hence,
\[
\tilde Z_u(t)\sim p_{u,t},\qquad t\in\mathcal S,\ u\in[0,1].
\]
In other words, \(\tilde Z_u(t)\) and \(Z_u(t)\) have the same distribution for each \(t\in\mathcal S\) and \(u\in[0,1]\).
In the sequel, we write \(Z_u(t)\) to denote either \(\tilde Z_u(t)\) or \(Z_u(t)\) appearing in \eqref{loss_FM} and \eqref{Conditional_match}, since they are equal in distribution.

Recall the objective of flow matching is to solve
\begin{eqnarray*}
   \min_{U}\ \int_0^1\ {\mathbb{E}\bigg[\bigg\{}\frac{\partial Z_u(T)}{\partial u}-U(u,T,Z_u(T)){\bigg\}^2\bigg]} \ \mathrm{d}u.
\end{eqnarray*} 
For $T=t$, 
\begin{eqnarray*}
&&\mathbb{E}\bigg\{\frac{\partial Z_u(t)}{\partial u}-U(u,t,Z_u(t))\bigg\}^2 \\
&=&  \mathbb{E}\big\{V_{u,t}(Z_u(t))-U(u,t,Z_u(t))\big\}^2 \\
&=&  {\mathbb{E}\big[\mathbb{E}\big[\big\{V_{u,t}(Z_u(t))-U(u,t,Z_u(t))\big\}^2\mid X(t)\big]\big]} \\
&=&   \mathbb{E}\bigg[\mathbb{E}\bigg[\bigg\{\int_{\mathbb{R}} V_{u,t}(Z_u(t)\mid x)\cdot \frac{q_{u,t}(Z_u(t)\mid x)\cdot f_{X}(x;t)}{p_{u,t}(Z_u(t))}\ \mathrm{d}x\\
&&-U(u,t,Z_u(t))\bigg\}^2\mid X(t)\bigg]\bigg] \\
&=&C- 2\mathbb{E}\bigg[\mathbb{E}\int_{\mathbb{R}} V_{u,t}(Z_u(t)\mid x)\cdot \frac{q_{u,t}(Z_u(t)\mid x)\cdot f_{X}(x;t)}{p_{u,t}(Z_u(t))}\ \mathrm{d}x\cdot U(u,t,Z_u(t))\mid X(t)\bigg]\\
&&+\mathbb{E}\big\{\mathbb{E}\big[U^2(u,t,Z_u(t))\mid X(t)\big]\big\}\\
&=&C- 2\int_{\mathbb{R}}\int_{\supp(p_{u,t})}\int_{\mathbb{R}} V_{u,t}(y\mid x)\cdot \frac{q_{u,t}(y\mid x)\cdot f_{X}(x;t)}{p_{u,t}(y)}\cdot U(u,t,y)\\
& &\cdot q_{u,t}(y\mid z)f_X(z;t)\ \mathrm{d}x\mathrm{d}y\mathrm{d}z+\mathbb{E}\big\{\mathbb{E}\big[U^2(u,t,Z_u(t))\mid X(t)\big]\big\}\\
&=&C- 2\int_{\mathbb{R}}\int_{\supp(p_{u,t})} V_{u,t}(y\mid x)\cdot q_{u,t}(y\mid x)\cdot f_{X}(x;t)\cdot U(u,t,y)\ \mathrm{d}x\mathrm{d}y\\
&&+\mathbb{E}\big\{\mathbb{E}\big[U^2(u,t,Z_u(t))\mid X(t)\big]\big\}\\
&=& C-2\mathbb{E}\big\{\mathbb{E}\big[V_{u,t}(Z_u(t)\mid X(t))\cdot U(u,t,Z_u(t))\mid X(t)\big]\big\} \ +\mathbb{E}\big\{\mathbb{E}\big[U^2(u,t,Z_u(t))\mid X(t)\big]\big\}\\
&=& {\mathbb{E}\big[\mathbb{E}\big[\big\{V_{u,t}(Z_u(t)\mid X(t))-U(u,t,Z_u(t))\big\}^2\mid X(t)\big]\big]+C} \\
&=& {\mathbb{E}\bigg[\mathbb{E}\bigg[\bigg(\frac{\partial Z_u(t)}{\partial u}-U(u,t,Z_u(t))\bigg)^2 \mid X(t)\bigg]\bigg]+C,}
\end{eqnarray*} 
where {$C$ denotes a generic constant that may vary from line to line and is independent of $U$.}
Let the density of $T$ be $p_T$. Then
\begin{eqnarray*}
&& \int_0^1{\mathbb{E}\bigg[\bigg\{}\frac{\partial Z_u(T)}{\partial u}-U(u,T,Z_u(T)){\bigg\}^2\bigg]} \ \mathrm{d}u\\
  &=&  \int_0^1\int_{\mathcal{S}}{\mathbb{E}\bigg[\bigg\{}\frac{\partial Z_u(t)}{\partial u}-U(u,t,Z_u(t)){\bigg\}^2\bigg]} p_T(t)\ \mathrm{d}t\mathrm{d}u\\
    &=&\int_0^1\int_{\mathcal{S}}{\mathbb{E}\bigg[\mathbb{E}\bigg[\bigg(\frac{\partial Z_u(t)}{\partial u}-U(u,t,Z_u(t))\bigg)^2 \mid X(t)\bigg]\bigg]}p_T(t) \ \mathrm{d}t\mathrm{d}u+C\\
     &=&\int_0^1{\mathbb{E}\bigg[\mathbb{E}\bigg[\bigg(\frac{\partial Z_u(T)}{\partial u}-U(u,T,Z_u(T))\bigg)^2 \mid X,T\bigg]\bigg]} \ \mathrm{d}u+C.
\end{eqnarray*}
The proof is complete.
\end{proof}

\subsection{Proof of Theorem~\ref{theo:consist_vec}}
For two sequences of non-negative real values $\{a_n\}$ and $\{b_n\}$, we say $a_n \lesssim b_n$ or $b_n\gtrsim a_n$ if there exists a constant $C>0$ such that $a_n \leq Cb_n$ for all $n$. 

Before proving, we impose the following condition.

\begin{ConS}\label{asum:Z} 
For \( u \in [0,1] \) and \( t \in \mathcal{S} \), the variable 
\( Z_{u,i}(t) = (1-u)\,Z_0(t) + u\,X_i(t) \) has a density \( p_{Z_u(t)} \) that is positive and uniformly bounded over \(u\in[0,1]\) and \(t\in\mathcal{S}\).
\end{ConS}

This condition can be obtained under Assumption~\ref{asum:density}. See Lemma \ref{assuption_bounded}.

\begin{proof}
We separate the proof into several parts.

\noindent\textbf{Existence and uniqueness of the vector field $V$.}
Let \(X\) have the same distribution as \(X_i\) but be independent of \(X_i\)s, and define
\(\ Z_u(t):=(1-u)\,Z_0(t)+u\,X(t)\).
Similarly, let \(T\) be identically distributed with \(T_{ij}\)s but be independent of \(T_{ij}\)s and independent of \(X\) and \(Z_0\).
Fix \(u\in[0,1]\). For any measurable function \(U\), consider
\[
\mathbb{E}\bigl(X(T)-Z_0(T)-U(u,T,Z_u(T))\bigr)^2.
\]
By the definition of conditional expectation, we have
\begin{eqnarray*}
&&\mathbb{E}\Bigl(X(T)-Z_0(T)
-
\mathbb{E}\bigl[X(T)-Z_0(T)\mid T,Z_u(T)\bigr]\Bigr)^2\\
   &\leq & \mathbb{E}\bigl(X(T)-Z_0(T)-U(u,T,Z_u(T))\bigr)^2,
\end{eqnarray*}
with equality if and only if
\begin{eqnarray}\label{con_U}
    U(u,T,Z_u(T))
=
\mathbb{E}\bigl[X(T)-Z_0(T)\mid T,Z_u(T)\bigr]
\quad \text{a.s.}
\end{eqnarray}

Define 
\[
V(u,t,z)
:=
\mathbb{E}\bigl[X(T)-Z_0(T)\mid T=t, Z_u(T)=z\bigr],
\]
so that, for every \(u\),
\[
V(u,T,Z_u(T))
=
\mathbb{E}\bigl[X(T)-Z_0(T)\mid T,Z_u(T)\bigr]
\quad \text{a.s.}
\]
For any \(U\), the pointwise inequality in \(u\) yields
\[
\mathcal{L}(V)
:=
\int_0^1
\mathbb{E}\Bigl(X(T)-Z_0(T)
-
\mathbb{E}\bigl[X(T)-Z_0(T)\mid T,Z_u(T)\bigr]\Bigr)^2\,\mathrm{d}u
\le
\mathcal{L}(U),
\]
hence \(V\) attains the minimum.

For uniqueness, suppose \(V_1\) and \(V_2\) are both minimizers. By the uniqueness of the \(\mathcal{L}^2\)-projection,
\[
V_1(u,T,Z_u(T)) = V_2(u,T,Z_u(T))
=
\mathbb{E}\bigl[X(T)-Z_0(T)\mid T,Z_u(T)\bigr]
\quad \text{a.s.},
\]
for each \(u\).

{To generalize almost sure equality to equality in the Lebesgue sense, we next prove that
\((T,Z_u(T))\) admits a joint density on \(\big\{(t,z): t\in \mathcal{S},\ z\in \supp\big(p_{Z_u(t)}\big)\big\}\) that is positive almost everywhere on its support, for each \(u\).}
Indeed, \(T\) has density \(p_T(t)\) that is positive and continuous on \(\mathcal{S}\), which means that \(0<c_T\le p_T(t)\le C_T\) for some constants \(c_T\) and \(C_T\).
Moreover, since \(T\) is independent of \(Z_u(\cdot)\), we have \(Z_u(T)\mid (T=t)\overset{d}{=}Z_u(t)\), so the conditional density of \(Z_u(T)\) given \(T=t\) is \(p_{Z_u(t)}(\cdot)\).
By Condition~\ref{asum:Z}, \(p_{Z_u(t)}(z)\) is positive and uniformly bounded on its support.
Hence, for each fixed \(u\in[0,1]\), the joint density of \((T,Z_u(T))\) is
\[
\ p_T(t)\,p_{Z_u(t)}(z),
\]
which is strictly positive almost everywhere on its support in
\(\big\{(t,z): t\in \mathcal{S},\ z\in \supp\big(p_{Z_u(t)}\big)\big\}\).

Thus, for a fixed \(u\), if \(V_1(u,t,z)\neq V_2(u,t,z)\) on a set of \((t,z)\) with positive Lebesgue measure, then
\(\mathbb{P}\bigl(V_1(u,T,Z_u(T))\neq V_2(u,T,Z_u(T))\bigr)>0\), contradicting the almost sure equality above.
Therefore, for each fixed \(u\in[0,1]\),
\[
V_1(u,t,z) = V_2(u,t,z)
\quad \text{for Lebesgue-a.e.\ }(t,z),
\]
which gives the uniqueness in the Lebesgue sense.

\

\noindent\textbf{The smoothness of the vector field \(V\).}
Let \(\Omega:=[0,1]\times\mathcal S\times\mathcal X\), where \(\mathcal X\subset\mathbb R\) is bounded.
Recall the endpoint definitions
\[
V(0,t,x):=\mu_X(t)-x,
\qquad
V(1,t,x):=x-\mu_0(t),
\]
and for \(u\in(0,1)\) define
\begin{eqnarray*}
    D(u,t,x)&:=&\int_{\mathbb R} f_X(z;t)\,p_{0,t}\!\left(\frac{x-uz}{1-u}\right)\,\mathrm{d}z,\\
M(u,t,x)&:=&\int_{\mathbb R} z\,f_X(z;t)\,p_{0,t}\!\left(\frac{x-uz}{1-u}\right)\,\mathrm{d}z,
\end{eqnarray*}
and
\begin{eqnarray*}
    \widetilde D(u,t,x)&:=&
\int_{\mathbb R} p_{0,t}(y)\, f_X\!\left(\frac{x-(1-u)y}{u};t\right)\,\mathrm{d}y,\\
\widetilde M(u,t,x)&:=&
\int_{\mathbb R} y\,p_{0,t}(y)\, f_X\!\left(\frac{x-(1-u)y}{u};t\right)\,\mathrm{d}y.
\end{eqnarray*}

Notice that
\begin{eqnarray}\label{exre_p}
    p_{Z_u(t)}(x)=\frac{1}{1-u}D(u,t,x)=\frac{1}{u}\widetilde D(u,t,x),
\end{eqnarray}
by the definition \(Z_u(t)=(1-u)Z_0(t)+uX(t)\). Then \(p_{Z_u(t)}(x)\) is continuous on
{\((0,1)\times \mathcal{S}\times\mathcal{X}\) by the continuity of \(f_{X}(x;t)\) and \(p_{0,t}(x)\). At the endpoints, \(p_{Z_0(t)}(x)=p_{0,t}(x)\) and \(p_{Z_1(t)}(x)=f_X(x;t)\), so \(p_{Z_u(t)}(x)\) is continuous on \([0,1]\times \mathcal{S}\times\mathcal{X}\).}
Combining this fact with Condition~\ref{asum:Z}, we have the uniform lower bound
\begin{eqnarray}\label{bound_p_density}
\inf_{(u,t,x)\in [0,1]\times\mathcal S\times\mathcal X}p_{Z_u(t)}(x)\ge c_D>0.
\end{eqnarray}

Fix \(u\in(0,1)\), \(t\in\mathcal S\), and \(x\in\mathbb R\).
Since \(Z_u(t)=(1-u)Z_0(t)+uX(t)\), conditioning on \((T=t,Z_u(t)=x)\) enforces
\[
(1-u)Z_0(t)+uX(t)=x
\quad\Longleftrightarrow\quad
Z_0(t)=\frac{x-uX(t)}{1-u}.
\]
Moreover, the conditional density of \(X(t)\) given \((T=t,Z_u(t)=x)\) is proportional to
\(z\mapsto f_X(z;t)\,p_{0,t}\!\bigl(\frac{x-uz}{1-u}\bigr)\), hence
\[
p_{X(t)\mid T=t,Z_u(t)=x}(z)
=
\frac{f_X(z;t)\,p_{0,t}\!\left(\frac{x-uz}{1-u}\right)}{D(u,t,x)},
\qquad z\in\mathbb R,
\]
and therefore
\[
\mathbb E\!\bigl[X(t)\mid T=t,Z_u(t)=x\bigr]
=
\frac{M(u,t,x)}{D(u,t,x)}.
\]
Plugging \(Z_0(t)=\frac{x-uX(t)}{1-u}\) into \eqref{con_U} yields
\[
V(u,t,x)
=
\mathbb E\!\bigl[\,X(t)-Z_0(t)\mid T=t,Z_u(t)=x\bigr]
=
\frac{1}{1-u}\left(\frac{M(u,t,x)}{D(u,t,x)}-x\right),
\quad u\in(0,1),
\]
where \(D(u,t,x)\ge c_D(1-u)\) due to \eqref{exre_p} and \eqref{bound_p_density}.
By symmetry (conditioning on \(Z_0(t)\) instead of \(X(t)\)), we also have
\[
V(u,t,x)
=
\frac{1}{u}\left(x-\frac{\widetilde M(u,t,x)}{\widetilde D(u,t,x)}\right),
\qquad u\in(0,1),
\]
where \(\widetilde D(u,t,x)\ge c_D\,u\) due to \eqref{exre_p} and \eqref{bound_p_density}.

Fix \(\varepsilon\in(0,1/2)\). On \([\varepsilon,1-\varepsilon]\times\mathcal S\times\mathcal X\),
both \(u\) and \(1-u\) are bounded away from \(0\).
Define the map
\[
\phi(u,x,z):=\frac{x-uz}{1-u}.
\]
{Then \(\partial_x\phi=(1-u)^{-1}\), \(\partial_u\phi=(x-z)(1-u)^{-2}\), and \(\partial_u^2\phi=2(x-z)(1-u)^{-3}\).}
Applying the chain rule, each mixed partial derivative \(\partial_u^a\partial_t^b\partial_x^c D(u,t,x)\) with
\(a+b+c\le 2\) can be written as a finite linear combination of integrals of the form
\[
\int_{\mathbb R} z^m\,
\partial_t^{b'} f_X(z;t)\,
\partial_t^{b''}\partial_x^{c'}p_{0,t}\bigl(\phi(u,x,z)\bigr)\,
\mathrm{d}z,
\qquad m\in\{0,1,2\},
\]
with coefficients bounded by a constant depending only on \(\varepsilon\) and \(\mathcal X\).
Under Assumption~\ref{asum:density} and the moment bounds
\(\sup_{t}\mathbb E|X(t)|^{4}<\infty\) and \(\sup_t\mathbb E|Z_0(t)|^{4}<\infty\),
these integrals admit an integrable envelope that does not depend on \((u,t,x)\in[\varepsilon,1-\varepsilon]\times\mathcal S\times\mathcal X\).
Hence dominated convergence implies that all mixed partial derivatives of \(D\) and \(M\) up to total order \(2\)
exist and are continuous on \([\varepsilon,1-\varepsilon]\times\mathcal S\times\mathcal X\).

Recall that \(D(u,t,x)\ge c_D(1-u)\ge c_D\varepsilon\) on this region, so \(D\) is bounded away from \(0\).
Therefore, by the quotient rule, the ratio \(M/D\) has mixed partial derivatives up to total order \(2\) that exist
and are continuous on \([\varepsilon,1-\varepsilon]\times\mathcal S\times\mathcal X\).
Since \(V(u,t,x)=\frac{1}{1-u}\bigl(\frac{M}{D}-x\bigr)\) and \((1-u)^{-1}\) is bounded on this region,
it follows that \(V\) has mixed partial derivatives up to total order \(2\) that exist and are continuous on
\([\varepsilon,1-\varepsilon]\times\mathcal S\times\mathcal X\).

{Control near \(u=0\):}
Consider \(u\in(0,1/2]\). Then \(1-u\ge 1/2\), so the factors \((1-u)^{-k}\) arising from the chain rule for \(\phi(u,x,z)\)
are uniformly bounded (for fixed \(k\le 4\)).
Repeating the argument as before on \((0,1/2]\times\mathcal S\times\mathcal X\) shows that \(D\) and \(M\), together with all
mixed partial derivatives up to total order \(2\), are continuous in \((u,t,x)\) on that set and admit limits as \(u\downarrow0\).
Using the endpoint definition \(V(0,t,x)=\lim_{u\downarrow0}V(u,t,x)\) (proved in the continuity part),
and defining the endpoint mixed derivatives (e.g., \(\partial_uV(0,t,x)\), \(\partial_u^2V(0,t,x)\)) as the corresponding limits
of the interior derivatives as \(u\downarrow0\), we conclude that \(V\) and its mixed partial derivatives up to total order \(2\)
{admit continuous one-sided extensions to \(u=0\) on \([0,1/2]\times\mathcal S\times\mathcal X\).}

Control near \(u=1\):
Consider \(u\in[1/2,1)\). Then \(u\ge 1/2\), so we use the symmetric representation
\[
V(u,t,x)=\frac{1}{u}\left(x-\frac{\widetilde M(u,t,x)}{\widetilde D(u,t,x)}\right),
\]
and define
\[
\psi(u,x,y):=\frac{x-(1-u)y}{u}.
\]
Applying the chain rule to \(\psi\), derivatives up to total order \(2\) produce factors involving \(u^{-k}\) (with \(k\le 4\))
and polynomials in \(|y|\) of degree at most \(2\), all uniformly controlled on \([1/2,1)\) up to constants depending on \(\mathcal X\).
Using Assumption~\ref{asum:density} and \(\sup_t\mathbb E|Z_0(t)|^4<\infty\) to dominate the \(|y|^2\) terms,
dominated convergence implies that \(\widetilde D\) and \(\widetilde M\) have mixed partial derivatives up to total order \(2\)
that exist and are continuous on \([1/2,1)\times\mathcal S\times\mathcal X\).
Recall that \(\widetilde D(u,t,x)\ge c_D u\ge c_D/2\) on this region, so \(\widetilde D\) is bounded away from \(0\),
and hence \(\widetilde M/\widetilde D\) has the same differentiability/continuity property by the quotient rule.
It follows that \(V\) has mixed partial derivatives up to total order \(2\) that exist and are continuous on
\([1/2,1)\times\mathcal S\times\mathcal X\).
Defining \(V(1,t,x):=\lim_{u\uparrow1}V(u,t,x)\) and defining endpoint mixed derivatives by limits as \(u\uparrow1\), {we obtain continuous one-sided extensions of \(V\) and all mixed partial derivatives up to total order \(2\) to \(u=1\) on
\([1/2,1]\times\mathcal S\times\mathcal X\).}

{The above shows} that \(V\) has mixed partial derivatives \(\partial_u^a\partial_t^b\partial_x^cV\) for all \(a+b+c\le 2\),
and these derivatives extend continuously to the compact set \(\overline\Omega\). Hence they are bounded on \(\overline\Omega\),
and since \(\Omega\) has finite Lebesgue measure,
\[
\int_{\Omega}\sum_{a+b+c\le 2}\bigl|\partial_u^a\partial_t^b\partial_x^cV(u,t,x)\bigr|^2\,
\mathrm{d}u\,\mathrm{d}t\,\mathrm{d}x
<\infty.
\]
Therefore all weak mixed derivatives up to total order \(2\) belong to \(\mathcal L^2(\Omega)\), and we conclude that
\(V\in\mathcal W_2^2(\Omega)\).

\

\noindent\textbf{Decomposition of error terms.} Let $\mathcal{X}=[-A, A]$ be a bounded set. Notice that
\begin{eqnarray*}
 &&   \int_{\mathbb{R}}\int_{\mathcal{S}}\int_0^1 
\mathbb{E} 
\left\{ 
\hat{V}\big(u, t, x\big) 
- 
V\big(u, t, x\big) 
\right\}^{2}p_{Z_u(t)}(x)\ \mathrm{d}u\,\mathrm{d}t\,\mathrm{d}x \\
&= &  \int_{\mathcal{X}}\int_{\mathcal{S}}\int_0^1 
\mathbb{E} 
\left\{ 
\hat{V}\big(u, t, x\big) 
- 
V\big(u, t, x\big) 
\right\}^{2}p_{Z_u(t)}(x)\ \mathrm{d}u\,\mathrm{d}t\,\mathrm{d}x  \\
&+& \int_{\mathbb{R}\setminus\mathcal{X}}\int_{\mathcal{S}}\int_0^1 
\mathbb{E} 
\left\{ 
\hat{V}\big(u, t, x\big) 
- 
V\big(u, t, x\big) 
\right\}^{2}p_{Z_u(t)}(x)\ \mathrm{d}u\,\mathrm{d}t\,\mathrm{d}x.
\end{eqnarray*}
Since $\hat{V}$ {has} bounded $\mathcal{L}^2$ norms on $[0,1]\times\mathcal{S}\times\mathbb{R}$ and Condition~\ref{asum:Z}, we have
\[
\int_{\mathbb{R}}\int_{\mathcal{S}}\int_0^1 \mathbb{E}\big\{\hat{V}(u,t,x)\big\}^2p_{Z_u(t)}(x)\,\mathrm{d}u\,\mathrm{d}t\,\mathrm{d}x<\infty.
\]
Meanwhile,
\begin{eqnarray*}
&&\int_{\mathbb{R}}\int_{\mathcal{S}}\int_0^1 \mathbb{E}\big\{V(u,t,x)\big\}^2p_{Z_u(t)}(x)\ \mathrm{d}u\,\mathrm{d}t\,\mathrm{d}x\\
&\lesssim&\int_0^1 \mathbb{E}\big\{V(u,T,Z_u(T))\big\}^2\ \mathrm{d}u\\
&=& \int_0^1 \mathbb{E}\Big|\mathbb E[X(T)-Z_0(T)\mid T,Z_u(T)]\Big|^2 \ \mathrm{d}u
 \\
&\le& \mathbb E\!\big(X(T)-Z_0(T)\big)^2\\
&<&\infty.
\end{eqnarray*}
Therefore,
\begin{eqnarray*}
    &&\int_{\mathbb{R}\setminus\mathcal{X}}\int_{\mathcal{S}}\int_0^1 
\mathbb{E}\big[\hat{V}(u,t,x)^2\big]{\,p_{Z_u(t)}(x)\,\mathrm{d}u\,\mathrm{d}t\,\mathrm{d}x} \to 0,\\
&&\int_{\mathbb{R}\setminus\mathcal{X}}\int_{\mathcal{S}}\int_0^1 
\mathbb{E}\big[V(u,t,x)^2\big]{\,p_{Z_u(t)}(x)\,\mathrm{d}u\,\mathrm{d}t\,\mathrm{d}x} \to 0,
\end{eqnarray*}
as $A\to\infty,$ and hence, by $\,(a-b)^2\le 2a^2+2b^2$,
\[
\int_{\mathbb{R}\setminus\mathcal{X}}\int_{\mathcal{S}}\int_0^1 
\mathbb{E} 
\left\{ 
\hat{V}\big(u, t, x\big) 
- 
V\big(u, t, x\big) 
\right\}^{2}{\,p_{Z_u(t)}(x)\ \mathrm{d}u\,\mathrm{d}t\,\mathrm{d}x} \to 0,
\quad A\to\infty.
\]
Let $A$ be sufficiently large such that
\[
\int_{\mathbb{R}\setminus\mathcal{X}}\int_{\mathcal{S}}\int_0^1 
\mathbb{E} 
\left\{ 
\hat{V}\big(u, t, x\big) 
- 
V\big(u, t, x\big) 
\right\}^{2}{\,p_{Z_u(t)}(x)\ \mathrm{d}u\,\mathrm{d}t\,\mathrm{d}x}
\]
is sufficiently small. It remains to prove that
\begin{eqnarray*}
\int_{\mathcal{X}}\int_{\mathcal{S}}\int_0^1 
\mathbb{E} 
\left\{ 
\hat{V}\big(u, t, x\big) 
- 
V\big(u, t, x\big) 
\right\}^{2}p_{Z_u(t)}(x)\ \mathrm{d}u\,\mathrm{d}t\,\mathrm{d}x
=o(1)
\end{eqnarray*}
as $n\to\infty$, which {implies that}
\begin{eqnarray}\label{target_error}
    \int_{\mathbb{R}}\int_{\mathcal{S}}\int_0^1 
\mathbb{E} 
\left\{ 
\hat{V}\big(u, t, x\big) 
- 
V\big(u, t, x\big) 
\right\}^{2}p_{Z_u(t)}(x)\ \mathrm{d}u\,\mathrm{d}t\,\mathrm{d}x
\to 0
\quad\text{as } n\to\infty.
\end{eqnarray}
In the following, we only focus on $x\in\mathcal{X}$ and abuse notation by writing {$\hat V$ and $V$} to denote {the} restrictions to the domain $\mathcal{X}$.

\

\noindent\textbf{Consistency of the vector field $V$.}
We proceed to prove the theorem under the simplifying assumption that \(\lambda_u = \lambda_t = \lambda_x = \lambda\). The proof for the more general case, where \(\lambda_u\), \(\lambda_t\), and \(\lambda_x\) differ, follows similarly.

Write $\mathbb{B}_{L_V}= \mathcal{B}_{L_V,4}([0,1],\mathcal{S},\mathcal{X})$.
Recall that $\hat{V}\in \mathbb{B}_{L_V}$ minimizes
\begin{eqnarray*}
  &&  \mathcal{L}_n(U) + \lambda\mathcal{J}(U) \\
    &=:&
\frac{1}{n}\sum_{i=1}^n\frac{1}{J_i}\sum_{j=1}^{J_i}\int_0^1\mathbb{E}\bigg[\bigg\{X_i(T_{ij})- Z_0(T_{ij})-U\big(u,T_{ij},Z_{u,i}(T_{ij})\big)\bigg\}^2\mid X_i,T_{ij}\bigg]\,\mathrm{d}u\\
&+&\lambda\mathcal{J}(U),
\end{eqnarray*}
where $\mathcal{J}(U)= \int_{\mathcal{X}}\int_{\mathcal{S}} \int_0^1 \left\{\left( \frac{\partial^2 U}{\partial u^2}\right)^2
    +\left( \frac{\partial^2 U}{\partial t^2}\right)^2+\left( \frac{\partial^2 U}{\partial x^2}\right)^2 \right\} \,\mathrm{d}u\,\mathrm{d}t\,\mathrm{d}x$.
Let $\mathcal{L}(U)=\mathbb{E}\{\mathcal{L}_n(U)\}$ denote the population loss; then $V=\argmin_{U}\mathcal{L}(U)$.

Define $V^*=\argmin_{U\in \mathbb{B}_{L_V}}\mathcal{L}(U)$.
As $V\in\mathcal{W}_2^{2}(\Omega)$, $V^*$ can approximate $V$ in the sense that \citep{de1983approximation}
\[
\|V^*-V\|_{\mathcal{L}^2}^2\;=\;o(1),
\]
as $L_V\rightarrow \infty$. Similarly, we have $\mathcal{J}(V^*)=O(1)$ and $\|V^*\|_{\mathcal{L}^2}=O(1)$ as $L_V\to \infty$.
Notice that $V^*$ is the minimizer of $\mathcal{L}$ on the closed convex set $\mathbb{B}_{L_V}$. By the Hilbert projection theorem (Theorem 2.5.1 in \cite{hsing2015theoretical}), we have
\begin{eqnarray}\label{hilbert_equ}
&&\mathcal{L}(U) - \mathcal{L}(V^*) \nonumber\\
&=&\int_0^1 \mathbb{E}_{T,Z_u}\Big\{
    V\big(u,T,Z_{u}(T)\big) - U\big(u,T,Z_{u}(T)\big)
  \Big\}^2 \, \mathrm{d}u\nonumber\\
  & & -\int_0^1 \mathbb{E}_{T,Z_u}\Big\{
    V\big(u,T,Z_{u}(T)\big) - V^*\big(u,T,Z_{u}(T)\big)
  \Big\}^2 \, \mathrm{d}u\nonumber\\
  &\geq& \int_0^1 \mathbb{E}_{T,Z_u}\Big\{
    U\big(u,T,Z_{u}(T)\big) - V^*\big(u,T,Z_{u}(T)\big)
  \Big\}^2 \, \mathrm{d}u,
\end{eqnarray}
for any $U \in \mathbb{B}_{L_V}$.

By the optimality of $\hat{V}$,
\[
\mathcal{L}_n(\hat{V})+\lambda \mathcal{J}(\hat{V})
\;\le\;
\mathcal{L}_n(V^*)+\lambda \mathcal{J}(V^*).
\]
Therefore,
\begin{eqnarray*}
  &&  \mathcal{L}(\hat{V})-\mathcal{L}(V^*) \\
    &\leq&
\mathcal{L}(\hat{V}) -\mathcal{L}(V^*) - \mathcal{L}_n(\hat{V})-\lambda \mathcal{J}(\hat{V})
+
\mathcal{L}_n(V^*)+\lambda \mathcal{J}(V^*)\\
 &\leq&
 \mathcal{L}_n(V^*) - \mathcal{L}(V^*)+ \mathcal{L}(\hat{V}) - \mathcal{L}_n(\hat{V})+\;\lambda \bigl\{\mathcal{J}(V^*)-\mathcal{J}(\hat{V})\bigr\}\\
&\lesssim &  \mathcal{L}_n(V^*) - \mathcal{L}(V^*)+ \mathcal{L}(\hat{V}) - \mathcal{L}_n(\hat{V}) + \lambda,
\end{eqnarray*}
where the final inequality is due to \( \mathcal{J}(V^*) =O(1) \) and \( \mathcal{J}(\hat{V}) \geq 0 \).
Combining the above inequality with \eqref{hilbert_equ}, we have
\begin{eqnarray*}
&&\int_0^1\mathbb{E}_{T,Z_u}\bigg\{\hat{V}\big(u,T,Z_{u}(T))-V\big(u,T,Z_{u}(T)\big)\bigg\}^2\,\mathrm{d}u \\
&\lesssim&
\int_0^1\mathbb{E}_{T,Z_u}\bigg\{\hat{V}\big(u,T,Z_{u}(T))-V^*\big(u,T,Z_{u}(T)\big)\bigg\}^2\,\mathrm{d}u\\
&&+
\int_0^1\mathbb{E}\bigg\{V^*\big(u,T,Z_{u}(T))-V\big(u,T,Z_{u}(T)\big)\bigg\}^2\,\mathrm{d}u\\
&\leq& \mathcal{L}(\hat{V})-\mathcal{L}(V^*)+
\int_0^1\mathbb{E}\bigg\{V^*\big(u,T,Z_{u}(T))-V\big(u,T,Z_{u}(T)\big)\bigg\}^2\,\mathrm{d}u \\
&\lesssim & \mathcal{L}_n(V^*) - \mathcal{L}(V^*)+ \mathcal{L}(\hat{V}) - \mathcal{L}_n(\hat{V})+ \lambda\\
&& +\int_0^1\mathbb{E}\bigg\{V^*\big(u,T,Z_{u}(T))-V\big(u,T,Z_{u}(T)\big)\bigg\}^2\,\mathrm{d}u.
\end{eqnarray*}

Since the densities of $T$ and $Z_{u}(t)$ are bounded, we have
\begin{eqnarray*}
  \int_0^1\mathbb{E}\bigg\{ V^*\big(u,T,Z_{u}(T)\big)-V\big(u,T,Z_{u}(T)\big)\bigg\}^2\,\mathrm{d}u\;\lesssim\;\|V^*-V\|^2_{\mathcal{L}^2}.
\end{eqnarray*}
The above two inequalities indicate
\begin{eqnarray*}
&&\int_0^1\mathbb{E}_{T,Z_{u}}\bigg\{\hat{V}\big(u,T,Z_{u}(T)\big)-V\big(u,T,Z_{u}(T)\big)\bigg\}^2\,\mathrm{d}u\nonumber\\
&\lesssim& |\mathcal{L}_n(V^*) - \mathcal{L}(V^*)|+ |\mathcal{L}(\hat{V}) - \mathcal{L}_n(\hat{V})| + \lambda + \|V^*-V\|^2_{\mathcal{L}^2}.
\end{eqnarray*}
By the assumption on the density of $T$, we have
\begin{eqnarray*}
&&\int_{\mathbb{R}}\int_{\mathcal{S}}\int_0^1 
\mathbb{E} 
\left\{ 
\hat{V}\big(u, t, x\big) 
- 
V\big(u, t, x\big) 
\right\}^{2}p_{Z_u(t)}(x)\ \mathrm{d}u\mathrm{d}t\mathrm{d}x\\
&=& {\int_0^1\int_{\mathcal{S}}
\mathbb{E}\left[
\left\{
\hat{V}\big(u, t, Z_u(t)\big)
-
V\big(u, t, Z_u(t)\big)
\right\}^{2}
\right] \,\mathrm{d}t\,\mathrm{d}u} \\
&\lesssim& \int_0^1
\mathbb{E}_{T,Z_{u}}
\bigg\{
\hat{V}\big(u,T,Z_{u}(T)\big)
-
V\big(u,T,Z_{u}(T)\big)
\bigg\}^2
\,\mathrm{d}u.
\end{eqnarray*}

As \( \lambda + \|V^* - V\|^2_{\mathcal{L}^2} = o(1) \), it remains to prove that
\begin{eqnarray}\label{o1}
  \mathbb{E}\sup_{U\in \mathbb{B}_{L_V}}|\mathcal{L}_n(U)-\mathcal{L}(U)| = o(1),
\end{eqnarray}
which implies
\begin{eqnarray}\label{vec_consistency}
   \int_{\mathcal{X}}\int_{\mathcal{S}}\int_0^1 
\mathbb{E} 
\left\{ 
\hat{V}\big(u, t, x\big) 
- 
V\big(u, t, x\big) 
\right\}^{2}p_{Z_u(t)}(x)\ \mathrm{d}u\mathrm{d}t\mathrm{d}x = o(1).
\end{eqnarray}

Define
\begin{eqnarray*}
h_n(U)
=
\frac{1}{n}\sum_{i=1}^n\bigg[X_i(T_{i})- \tilde{Z}_{0,i}(T_{i})-U\big(W_i,T_{i},\tilde{Z}_{W_i,i}(T_{i})\big)\bigg]^2  -\mathcal{L}(U),
\end{eqnarray*}
where {for each \(i\), \(W_i \sim \mathrm{Unif}([0,1])\), \(T_i \sim p_T\), and \(\tilde{Z}_{0,i}(t)\sim p_{Z_0(t)}\), and \(W_i\), \(T_i\), and \(\tilde{Z}_{0,i}\) are independent.}
For $u\in[0,1]$, define the transported process
\[
\tilde{Z}_{u,i}(t):=(1-u)\,\tilde{Z}_{0,i}(t)+u\,X_i(t),\qquad t\in\mathcal S,
\]
so that $\tilde{Z}_{W_i,i}(T_i)=\tilde{Z}_{u,i}(T_i)\big|_{u=W_i}$.
The triplet $(W_i, T_i, \tilde{Z}_{0,i})$ is coupled with the $i$th function $X_i$ and the triplets are independent across $i$.
Let $K_1,\dots,K_n$ be independent such that
$K_i\mid J_i \sim {\rm Unif}\{1,\dots,J_i\}$ and independent of everything else, and write
$T_i^\ast := T_{iK_i}$.
Then, by the definition of $\mathcal{L}_n(U)$,
\begin{align*}
&\mathcal{L}_n(U)\\
=&\frac{1}{n}\sum_{i=1}^n\frac{1}{J_i}\sum_{j=1}^{J_i}\int_0^1
\mathbb{E}\!\left[\Big\{X_i(T_{ij})- Z_0(T_{ij})-U\big(u,T_{ij},Z_{u,i}(T_{ij})\big)\Big\}^2\mid X_i,T_{ij}\right]\,\mathrm{d}u\\
=&\mathbb{E}\!\left[
\frac{1}{n}\sum_{i=1}^n\int_0^1
\mathbb{E}\!\left[\Big\{X_i(T_i^\ast)- \tilde Z_{0,i}(T_i^\ast)- U\big(u,T_i^\ast,\tilde Z_{u,i}(T_i^\ast)\big)\Big\}^2\mid X_i,T_i^\ast\right]\,\mathrm{d}u
\ \Big|\ \mathcal{G}
\right],
\end{align*}
where $\mathcal{G}=\{X_i,\{T_{ij}\}_{j=1}^{J_i}\}_{i=1}^n$, and we used that $Z_0$ is independent of $(X_i,T_{ij})$ and replaced $Z_0$ by an i.i.d.\ copy $\tilde Z_{0,i}$ inside the conditional expectation.
Since $\mathcal{L}(U)=\mathbb{E}\{\mathcal{L}_n(U)\}$, we have
\begin{eqnarray*}
&&\mathcal{L}_n(U)-\mathcal{L}(U)
\\
&=&
\mathbb{E}\!\left[
\frac{1}{n}\sum_{i=1}^n\int_0^1
\mathbb{E}\!\left[\Big\{X_i(T_i^\ast)- \tilde Z_{0,i}(T_i^\ast)- U\big(u,T_i^\ast,\tilde Z_{u,i}(T_i^\ast)\big)\Big\}^2\mid X_i,T_i^\ast\right]\,\mathrm{d}u
-\mathcal{L}(U)
\ \Big|\ \mathcal{G}
\right].
\end{eqnarray*}
Therefore, by Jensen's inequality,
\begin{align*}
&\sup_{U\in \mathbb{B}_{L_V}}|\mathcal{L}_n(U)-\mathcal{L}(U)|\\
\le&
\mathbb{E}\!\bigg\{
\sup_{U\in \mathbb{B}_{L_V}}\bigg|
\frac{1}{n}\sum_{i=1}^n\int_0^1
\mathbb{E}\!\left[\Big\{X_i(T_i^\ast)- \tilde Z_{0,i}(T_i^\ast)- U\big(u,T_i^\ast,\tilde Z_{u,i}(T_i^\ast)\big)\Big\}^2\mid X_i,T_i^\ast\right]\,\mathrm{d}u\\
-&\mathcal{L}(U)
\bigg| \Big| \ \mathcal{G} \bigg\}.
\end{align*}
Taking expectations gives
\begin{align*}
&\mathbb{E}\sup_{U\in \mathbb{B}_{L_V}}|\mathcal{L}_n(U)-\mathcal{L}(U)|\\
\le&
\mathbb{E}\sup_{U\in \mathbb{B}_{L_V}}\bigg|
\frac{1}{n}\sum_{i=1}^n\int_0^1
\mathbb{E}\!\left[\Big\{X_i(T_i^\ast)-\tilde Z_{0,i}(T_i^\ast)- U\big(u,T_i^\ast,\tilde Z_{u,i}(T_i^\ast)\big)\Big\}^2\mid X_i,T_i^\ast\right]\,\mathrm{d}u\\
-&\mathcal{L}(U)
\bigg|.
\end{align*}
Notice for any $t$,
\begin{eqnarray*}
 &&   \int_0^1
\mathbb{E}\!\left[\Big\{X_i(t)- {\tilde Z_{0,i}}(t)-U\big(u,t,{\tilde Z_{u,i}}(t)\big)\Big\}^2\mid X_i,t\right]\,\mathrm{d}u\\
&=&
\mathbb{E}\!\left(
\Big[X_i(t)-\tilde{Z}_{0,i}(t)-U\big(W_i,t,\tilde{Z}_{W_i,i}(t)\big)\Big]^2\ \Big|\ X_i,t
\right).
\end{eqnarray*}
Applying this with $t=T_i^\ast$ and averaging over $i$ shows that
\begin{align*}
&\frac{1}{n}\sum_{i=1}^n\int_0^1
\mathbb{E}\!\left[\Big\{X_i(T_i^\ast)- {\tilde Z_{0,i}}(T_i^\ast)-U\big(u,T_i^\ast,{\tilde Z_{u,i}}(T_i^\ast)\big)\Big\}^2\mid X_i,T_i^\ast\right]\,\mathrm{d}u-\mathcal{L}(U)\\
=&\ \mathbb{E}\!\left[{h_n^\ast(U)}\ \Big|\ \{X_i,T_i^\ast\}_{i=1}^n\right],
\end{align*}
where
\[
{h_n^\ast(U):=\frac{1}{n}\sum_{i=1}^n\Big[X_i(T_i^\ast)-\tilde Z_{0,i}(T_i^\ast)-U\big(W_i,T_i^\ast,\tilde Z_{W_i,i}(T_i^\ast)\big)\Big]^2-\mathcal L(U).}
\]
Hence, again by Jensen's inequality,
\[
\sup_{U\in \mathbb{B}_{L_V}}\left|\mathbb{E}\!\left[{h_n^\ast(U)}\mid \{X_i,T_i^\ast\}_{i=1}^n\right]\right|
\le
\mathbb{E}\!\left[\sup_{U\in \mathbb{B}_{L_V}}|{h_n^\ast(U)}|\mid \{X_i,T_i^\ast\}_{i=1}^n\right],
\]
and taking expectations yields
\[
\mathbb{E}\sup_{U\in \mathbb{B}_{L_V}}|\mathcal{L}_n(U)-\mathcal{L}(U)|
\le
\mathbb{E}\sup_{U\in \mathbb{B}_{L_V}}|{h_n^\ast(U)}|.
\]
{Since \(T_i^\ast \overset{d}{=} T_i\), we have}
\[
\mathbb{E}\sup_{U\in \mathbb{B}_{L_V}}|\mathcal{L}_n(U)-\mathcal{L}(U)|
\le
\mathbb{E}\sup_{U\in \mathbb{B}_{L_V}}|{h_n(U)}|.
\]

With the above inequality, it suffices to show that
\begin{equation}\label{o22}
    \mathbb{E}\,\sup_{U \in \mathbb{B}_{L_V}} \big| h_n(U) \big| = o(1)
\end{equation}
to obtain \eqref{o1}, which in turn implies \eqref{vec_consistency}.

Notice that \( \mathbb{E} X_i^4(t) \) and \( \mathbb{E} \tilde{Z}_{0,i}^4(t) \) are uniformly bounded for
\( i = 1,\ldots,n \) and $U$ has a bounded \(\mathcal{W}_1^{\infty}(\Omega)\)-norm.
Therefore, we have
\[
\sup_{n \ge 1} \mathbb{E}\,\sup_{U \in \mathbb{B}_{L_V}} h_n^2(U) < \infty,
\]
which implies that the collection
\(
\{ \sup_{U \in \mathbb{B}_{L_V}} |h_n(U)| : n \ge 1 \}
\)
is uniformly integrable.
By Theorem~2.20 of \cite{van2000asymptotic}, it follows that
\begin{equation}\label{GC}
    \sup_{U \in \mathbb{B}_{L_V}} |h_n(U)| = o_p(1)
\end{equation}
implies \eqref{o22}.

It therefore remains to establish \eqref{GC}.
Note that the dimension of the function class \( \mathbb{B}_{L_V} \) is of order \( L_V^3 \), and hence its Vapnik--Chervonenkis (VC) dimension is at most \( L_V^3 \); see Proposition~4.20 in \cite{wainwright2019high}.
By the Glivenko--Cantelli theorem for empirical processes (Theorems~2.4.3 and~2.6.7 in \cite{van2023weak}), \eqref{GC} holds provided that
\[
\frac{L_V^3}{n} \to 0 .
\]
This condition follows from the assumption \( L_V = o(n^{1/3}) \). 
This completes the proof.

\end{proof}

\subsection{Proof of Theorem~\ref{theo:consist}}
\begin{asumS}\label{asum:Llambda_rho}
The tuning parameters \( L_\rho, \lambda_1, \lambda_2 \) in \eqref{est_rho} satisfy the following conditions.  
Assume there exist constants \( \delta_1, \delta_2 > 0 \) and a sufficiently large constant \( C > 0 \) such that, as \( n \to \infty \), for \(k=1,2\),
\[
L_\rho \ge n^{\delta_1}, \qquad 
\lambda_k\, L_\rho^{4} \ge C, \qquad 
\lambda_k = o\!\left(n^{-\delta_2}\right).
\]
In addition, \(L_\rho\) satisfies
\[
L_\rho^2 = o\!\left(\frac{n}{\log n}\right), \qquad 
\frac{1}{n^2}\sum_{i=1}^n \frac{1}{J_i^2}
=
o\!\left(\frac{1}{L_\rho^{4}\log n}\right).
\]
\end{asumS}

Assumption~\ref{asum:Llambda_rho} specifies an asymptotic regime in which the spline space is rich enough to approximate \(\rho\) while the bivariate estimation problem remains well-posed. 
In particular, \(L_\rho\) grows with \(n\) but not too fast \(\big(L_\rho^2=o(n/\log n)\big)\), and the smoothing parameters \(\lambda_k\) vanish while remaining in the smoothing-spline regime \(\big(\lambda_k L_\rho^{4}\ge C\big)\), so the penalty controls roughness without introducing non-negligible asymptotic error. 
The additional condition \(\frac{1}{n^2}\sum_{i=1}^n J_i^{-2}=o\!\big((L_\rho^{4}\log n)^{-1}\big)\) ensures that irregular sampling across subjects does not dominate the stochastic error. 
This type of condition is conventional in penalized-spline estimation for covariance functions \citep{xiao2020asymptotic}.

\begin{proof}
We separate the proof into several parts.

\noindent\textbf{Consistency of the flow map $\hat{\psi}_{1,t}$.}
Let $\delta_t(z) = \hat{\psi}_{1,t}(z) - \psi_{1,t}(z)$. 
In the following, we aim to control the error rate between $\hat{\psi}_{1,t}$ and $\psi_{1,t}$. 

By the Alekseev--Gr\"obner formula,
\begin{eqnarray}\label{AG_fomula}
\delta_t(z) 
&=&  \hat{\psi}_{1,t}(z) - \psi_{1,t}(z)\nonumber\\
&=& \int_0^1 
{\frac{\mathrm{d}\hat \psi_{(u,1),t}(x)}{\mathrm{d}x}\bigg|_{x=\psi_{u,t}(z)}}
\cdot \bigg({V}(1-u,t,\psi_{u,t}(z))- \hat{V}(1-u,t,\psi_{u,t}(z))\bigg)\ \mathrm{d}u.  
\end{eqnarray}
where \( \hat\psi_{(u,v),t}(z) \) denotes the solution of
\begin{eqnarray*}
\frac{\partial}{\partial s}\,\hat{\psi}_{(u,s),t}(z) 
= -\hat{V}\!\big(1-s,t,\hat{\psi}_{(u,s),t}(z)\big),
\qquad 
\hat{\psi}_{(u,u),t}(z) = z,
\qquad s\in[u,v],
\end{eqnarray*}
starting at \( s=u \) and ending at \( s=v \), with initial value \( z \).
Notice that
\[
\hat \psi_{(u,w),t}(z)=z-\int_{u}^w \hat V\!\big(1-v,t, \hat\psi_{(u,v),t}(z)\big)\ \mathrm{d}v,
\]
then
\begin{eqnarray*}
\frac{\mathrm{d}\hat \psi_{(u,w),t}(z)}{\mathrm{d}z}
=   1-\int_{u}^w{\frac{\partial \hat V(1-v,t,x)}{\partial x}}\bigg|_{x=\hat\psi_{(u,v),t}(z)}\cdot \frac{\mathrm{d}\hat \psi_{(u,v),t}(z)}{\mathrm{d}z}\ \mathrm{d}v.
\end{eqnarray*}
{Notice that \(\hat{V}\in \mathcal{W}_1^{\infty}([0,1]\times\mathcal{S}\times\mathbb{R})\). Hence, there exists a constant \(L_1\) such that  
\(\frac{\partial}{\partial x}\hat V(u,t,x)\) is bounded by \(L_1\) for any \(u\), \(t\), and \(x\).}
Then
\begin{eqnarray*}
\bigg|\frac{\mathrm{d}\hat \psi_{(u,w),t}(z)}{\mathrm{d}z}\bigg|
\leq    1 +\int_{u}^wL_1\cdot \bigg|\frac{\mathrm{d}\hat \psi_{(u,v),t}(z)}{\mathrm{d}z}\bigg|\ \mathrm{d}v.
\end{eqnarray*}
Note that $\hat \psi_{(u,u),t}(z)=z$ and $\frac{\mathrm{d}\hat \psi_{(u,u),t}(z)}{\mathrm{d}z}=1$. By Gr\"onwall's inequality,
\begin{eqnarray*}
    \bigg|\frac{\mathrm{d}\hat \psi_{(u,1),t}(z)}{\mathrm{d}z}\bigg|\leq \exp\big((1-u)L_1\big).
\end{eqnarray*}
{Combining} the above inequalities with the Alekseev--Gr\"obner formula \eqref{AG_fomula},
\begin{eqnarray*}
 \int_{\mathcal{S}}\mathbb{E}\delta^2_t(X(t))\ \mathrm{d}t
\lesssim   \mathbb{E}\int_\mathcal{S}\int_0^1 \bigg|\hat{V}(1-u,t,\psi_{u,t}(X(t)))- {V}(1-u,t,\psi_{u,t}(X(t)))\bigg|^2\ \mathrm{d}u \mathrm{d}t,
\end{eqnarray*}
where \(X\) {is identically distributed as \(X_i\) and is independent of the \(X_i\)'s}.
Notice that \(V\) satisfies the conditions in Theorem~\ref{inverse_formula}. 
Therefore, $\psi_{u,t}(X(t))$ follows the same density as $Z_{1-u}(t)$ for each $t$ by Theorem~\ref{Theo: cond}. 
Then, by the change of variable \(u\mapsto 1-u\),
\begin{eqnarray}\label{error_delta}
  \int_{\mathcal{S}}\mathbb{E}\delta^2_t(X(t))\ \mathrm{d}t
&\lesssim& \mathbb{E}\int_{\mathcal{S}}\int_0^1 \bigg|\hat{V}(u,t,Z_u(t))- {V}(u,t,Z_u(t))\bigg|^2\ \mathrm{d}u\mathrm{d}t \\
&=& O\!\big(\mathcal{R}^2(\hat V, V)\big).\nonumber
\end{eqnarray}

\

\noindent\textbf{Consistency of the latent correlation functions $\hat{\rho}$.}
We prove that the estimator \(\hat{\rho}\) is consistent for the true function \(\rho\),
where the true function is defined as
\[
\rho(t, s) = \mathbb{E}\big[\psi_{1,t}(X(t))\, \psi_{1,s}(X(s))\big], \quad t, s \in \mathcal{S},
\]
and
\[
\hat{\rho} = \argmin_{f \in \mathcal{B}_{L_{\rho},4}(\mathcal{S},\mathcal{S})}
\frac{1}{n} \sum_{i=1}^n \frac{1}{J_i^2} \sum_{j_1, j_2 = 1}^{J_i}
\left[ G_i(T_{ij_1}, T_{ij_2}) - f(T_{ij_1}, T_{ij_2}) \right]^2 + \lambda\,\mathcal{J}(f),
\]
where
\[
G_i(T_{ij_1}, T_{ij_2})
=
\hat{\psi}_{1,T_{ij_1}}\!\big(X_i(T_{ij_1})\big)\cdot
\hat{\psi}_{1,T_{ij_2}}\!\big(X_i(T_{ij_2})\big).
\]
Define
\[
\tilde{\rho}(t, s) = \mathbb{E}_X\big[G_i(t, s)\big]
= \mathbb{E}_X\big[\hat{\psi}_{1,t}(X(t))\,\hat{\psi}_{1,s}(X(s))\big].
\]
By the triangle inequality,
\[
\|\hat{\rho} - \rho\|_{\mathcal{L}^2}
\leq
\|\hat{\rho} - \tilde{\rho}\|_{\mathcal{L}^2}
+
\|\tilde{\rho} - \rho\|_{\mathcal{L}^2}.
\]
In the following, we separately control the error bounds
\( \|\hat{\rho} - \tilde{\rho}\|_{\mathcal{L}^2} \) and
\( \|\tilde{\rho} - \rho\|_{\mathcal{L}^2} \).

\textbf{(a) Error bound for \( \|\tilde{\rho} - \rho\|_{\mathcal{L}^2} \):} Notice that
\begin{eqnarray*}
&&    \hat{\psi}_{1,t}(X(t)) \hat{\psi}_{1,s}(X(s)) \\
    &=& \psi_{1,t}(X(t)) \psi_{1,s}(X(s)) + \psi_{1,t}(X(t)) \delta_s(X(s)) \\
    & & + \psi_{1,s}(X(s)) \delta_t(X(t)) + \delta_t(X(t)) \delta_s(X(s)).
\end{eqnarray*}
Taking expectations:
\begin{eqnarray}\label{equ:rho}
    \tilde{\rho}(t, s) &=& \rho(t, s) + \mathbb{E}_{X}[\psi_{1,t}(X(t)) \delta_s(X(s))] + \mathbb{E}_{X}[\psi_{1,s}(X(s)) \delta_t(X(t))]\nonumber\\
& & + \mathbb{E}_{X}[\delta_t(X(t)) \delta_s(X(s))]\nonumber\\
&:=& \rho(t, s) + A(t,s) + B(t,s) + C(t,s).
\end{eqnarray}

By a Gr\"onwall-type argument similar to \eqref{Gr_inequality}, we can show that there exist constants $C_1$ and $C_2$ such that
\begin{eqnarray}\label{psi_b}
    |{\psi}_{1,t}(x)| \;\leq\; C_1 |x| + C_2,
\end{eqnarray}
which follows from the fact that $V(u,t,\cdot)$ is Lipschitz in $x$, uniformly over $(u,t)\in[0,1]\times\mathcal S$. Therefore, by Assumption~\ref{asum:X},
\begin{eqnarray}\label{phi_bound_h}
    \sup_{t\in\mathcal{S}} \mathbb{E}\Big\{\psi_{1,t}\!\big(X(t)\big)\Big\}^{4} < \infty.
\end{eqnarray}
{This implies that}
\begin{eqnarray*}
    \|A\|_{\mathcal{L}^2}^2&=&  {\int_{\mathcal{S}\times \mathcal{S}}\bigg(\mathbb{E}_{X}[\psi_{1,t}(X(t))\,\delta_s(X(s))]\bigg)^2\ \mathrm{d}t\mathrm{d}s}\\
    &\leq& \int_{\mathcal{S}\times \mathcal{S}}\mathbb{E}_{X}[\psi^2_{1,t}(X(t))] \mathbb{E}_{X}[\delta^2_s(X(s))]\ \mathrm{d}t\mathrm{d}s\\
    &=&O\!\bigg({ \int_{\mathcal{S}} \mathbb{E}_{X}\delta^2_s(X(s))\ \mathrm{d}s}\bigg).
\end{eqnarray*}
By \eqref{error_delta}, we have
\begin{eqnarray*}
    \|A\|_{\mathcal{L}^2}^2 = O_p\!\Big(\mathcal{R}^2(\hat V,V)\Big).
\end{eqnarray*}
We similarly prove that $\|B\|_{\mathcal{L}^2}^2=O_p\!\Big(\mathcal{R}^2(\hat V,V)\Big)$ and $\|C\|_{\mathcal{L}^2}^2=O_p\!\Big(\mathcal{R}^2(\hat V,V)\Big)$, then \eqref{equ:rho} leads to
\begin{eqnarray}\label{rho_mean}
    \|\tilde{\rho} - \rho\|^2_{\mathcal{L}^2} =O_p\!\Big(\mathcal{R}^2(\hat V,V)\Big).
\end{eqnarray}

\textbf{(b) Error bound for \( \|\tilde{\rho} - \hat\rho\|_{\mathcal{L}^2} \):} 
Similar to \eqref{psi_b}, we have
\(
|\hat{\psi}_{1,t}(x)| \le {\tilde C_1}|x|+{\tilde C_2}
\)
for all \(t\in\mathcal S\) and \(x\in\mathbb R\), {where \(\tilde C_1\) and \(\tilde C_2\) are constants independent of \(t\) and \(x\)}.
Hence, by Cauchy--Schwarz,
\[
|\tilde{\rho}(t,s)|
=
\Big|\mathbb{E}_X\big[\hat{\psi}_{1,t}\!\big(X(t)\big)\,\hat{\psi}_{1,s}\!\big(X(s)\big)\big]\Big|
\le
\Big(\mathbb{E}\hat{\psi}_{1,t}\!\big(X(t)\big)^2\Big)^{1/2}
\Big(\mathbb{E}\hat{\psi}_{1,s}\!\big(X(s)\big)^2\Big)^{1/2}.
\]
Moreover,
\(
\mathbb{E}_X\hat{\psi}^2_{1,t}\!\big(X(t)\big)
\le
2\tilde C_1^2\,\mathbb{E}X(t)^2+2\tilde C_2^2,
\)
so under Assumption~\ref{asum:X},
we obtain \(\sup_{t,s\in\mathcal S}|\tilde{\rho}(t,s)|<\infty\).

When \(t\neq s\), denote the joint density of \((X(t),X(s))\) by \(f_{X(t),X(s)}(x,z)\). Then
\begin{eqnarray*}
\tilde{\rho}(t, s)
&=&
\int_{\mathbb{R}}\int_{\mathbb{R}}
\hat{\psi}_{1,t}(x)\,\hat{\psi}_{1,s}(z)\,
f_{X(t),X(s)}(x,z)\ \mathrm{d}x\,\mathrm{d}z .
\end{eqnarray*}
Since \(X(\cdot)\) is a Gaussian copula process, Sklar’s theorem (see Part~\ref{sec: sklar}) yields
\begin{eqnarray}\label{hat_rho_exp}
f_{X(t),X(s)}(x,z)
=
c_{X}\big(F_t(x),F_s(z);\rho(t,s)\big)\,f_X(x;t)\,f_X(z;s),
\end{eqnarray}
where \(c_X\) is the Gaussian copula density determined by \(\rho(t,s)\).
Writing \(a=\Phi^{-1}(u)\) and \(b=\Phi^{-1}(v)\), we have
\[
c_X(u,v;\rho)
=
\frac{1}{\sqrt{1-\rho^{2}}}
\exp\!\left[
-\frac{1}{2(1-\rho^{2})}\bigl(a^{2}-2\rho\,ab+b^{2}\bigr)
+\frac12(a^{2}+b^{2})
\right].
\]
Therefore, for \(t\neq s\),
\begin{eqnarray*}
\tilde{\rho}(t, s)
&=&
\int_{\mathbb{R}}\int_{\mathbb{R}}
\hat{\psi}_{1,t}(x)\,\hat{\psi}_{1,s}(z)\,
c_{X}\big(F_t(x),F_s(z);\rho(t,s)\big)\,f_X(x;t)\,f_X(z;s)\ 
\mathrm{d}x\,\mathrm{d}z .
\end{eqnarray*}
When \(t=s\), the pair \((X(t),X(t))\) is supported on the diagonal \(\{x=z\}\) and does not admit a
two-dimensional density with respect to \(\mathrm{d}x\,\mathrm{d}z\). In this case,
\begin{eqnarray}\label{hat_rho_diag}
\tilde{\rho}(t,t)
=
\mathbb{E}_X\Big[\hat{\psi}^2_{1,t}\!\big(X(t)\big)\Big]
=
\int_{\mathbb{R}}\hat{\psi}_{1,t}^2(x)\, f_X(x;t)\,\mathrm{d}x.
\end{eqnarray}

Notice that:
(i) \(\rho(t,s)\) has continuous mixed partial derivatives up to total order \(2\) on \(\mathcal S\times\mathcal S\);
(ii) \((t,x)\mapsto f_X(x;t)\) has continuous mixed partial derivatives up to total order \(2\);
and (iii) \((t,x)\mapsto \hat{\psi}_{1,t}(x)\) has continuous partial derivatives in \(t\) up to order \(2\),
which follows from the smoothness of \(\hat V(u,t,x)\) in \((t,x)\) and the differential equation defining \(\hat\psi\).
Then we can differentiate under the integral sign in \eqref{hat_rho_exp} for \(t\neq s\), and in
\eqref{hat_rho_diag} on the diagonal, implying that \(\tilde{\rho}(t,s)\) has continuous mixed partial
derivatives up to total order \(2\) on \(\{(t,s)\in\mathcal S^2:t\neq s\}\), and \(\tilde{\rho}(t,t)\) has
continuous derivatives up to order \(2\) in \(t\) along the diagonal.

Under the above setting, {$\tilde{\rho}$ plays the role of the target covariance function, and the error rate for 
\( \|\tilde{\rho}-\hat{\rho}\|_{\mathcal{L}^2} \) is a classical covariance-function convergence rate that can be obtained directly from the results in \cite{xiao2020asymptotic}.}
We can verify that the covariance estimation here satisfies Assumptions~1,~6,~9--10, and~13 in \cite{xiao2020asymptotic}. 
Specifically, the second-order smoothness of $\tilde{\rho}$ implies Assumption~1 therein; 
Assumption~\ref{asum:timepoint} implies Assumption~6 therein; 
\eqref{phi_bound_h} implies Assumption~9 therein; 
and Assumption~\ref{asum:Llambda_rho} implies Assumptions~10 and~13 therein.
Therefore,
\[
\mathbb{E}\!\left(\|\hat\rho-\tilde\rho\|_{\mathcal L^2}^2\mid T_{ij},\ i=1,\ldots,n,\ j=1,\ldots,J_i,\ \hat{\psi}_{1,t},\ t\in\mathcal S\right)
{=
O_p\!\left((n\bar J ^2 )^{-2/3}+n^{-1}\right),}
\]
by Corollary~4.4 in \cite{xiao2020asymptotic}. 
Here, the rate holds provided that the tuning parameters $\lambda_1$ and $\lambda_2$ are chosen at the optimal order:
\[
\min\{(n\bar J^{\,2})^{-2/3}, \bar J^{-4}
\}\;\le\;
\lambda_1,\lambda_2
\;\le\;
\max\{(n\bar J^{\,2})^{-2/3}, n^{-1}\}.
\]
The unconditional rate of $\| \rho - \hat{\rho} \|^2_{\mathcal{L}^2}$ follows by iterated expectation and Markov's inequality. 
Let \(a_n:=(n\bar J^2)^{-2/3}+n^{-1}\). Since the conditional bound implies
\[
\mathbb{E}\!\left(\|\hat\rho-\tilde\rho\|_{\mathcal L^2}^2\mid T_{ij}, i= 1,\ldots, n, j = 1,\ldots,J_i,\hat{\psi}_{1,t},t\in \mathcal{S}\right)=O_p(a_n),
\]
{for some \(M>0\) and any $\varepsilon>0$,}
\begin{eqnarray*}
  && \mathbb{P}\!\left(\|\hat\rho-\tilde\rho\|_{\mathcal L^2}^2>Ma_n\right)\\
  &=& 
  \mathbb{E}\!\left[
  \mathbb{P}\!\left(\|\hat\rho-\tilde\rho\|_{\mathcal L^2}^2>Ma_n
  \,\middle|\, T_{ij},\ i=1,\ldots,n,\ j=1,\ldots,J_i,\hat{\psi}_{1,t},t\in \mathcal{S} \right)
  \right]\\
  &\le&
  \mathbb{P}(A_n^c)
  +\mathbb{E}\!\left[
  \mathbb{I}_{A_n}\,
  \mathbb{P}\!\left(\|\hat\rho-\tilde\rho\|_{\mathcal L^2}^2>Ma_n
  \,\middle|\, T_{ij},\ i=1,\ldots,n,\ j=1,\ldots,J_i,\hat{\psi}_{1,t},t\in \mathcal{S} \right)
  \right]\\
  &\le&
  \varepsilon
  +\mathbb{E}\!\left[
  \mathbb{I}_{A_n}\,
  \frac{\mathbb{E}\!\left(\|\hat\rho-\tilde\rho\|_{\mathcal L^2}^2
  \,\middle|\, T_{ij},\ i=1,\ldots,n,\ j=1,\ldots,J_i,\hat{\psi}_{1,t},t\in \mathcal{S} \right)}{Ma_n}
  \right]\\
  &\le& \varepsilon+\frac{C_\varepsilon}{M},
\end{eqnarray*}
where
\(
A_n:=\left\{
\mathbb{E}\!\left(\|\hat\rho-\tilde\rho\|_{\mathcal L^2}^2
\,\middle|\, T_{ij},\ i=1,\ldots,n,\ j=1,\ldots,J_i ,\hat{\psi}_{1,t},t\in \mathcal{S}\right)
\le C_\varepsilon a_n
\right\},
\)
and \(C_\varepsilon>0\) is chosen such that \(\mathbb{P}(A_n^c)\le \varepsilon\) for all sufficiently
large \(n\). Take \(M\) sufficiently large such that
\[
\mathbb{P}\!\left(\|\hat\rho-\tilde\rho\|_{\mathcal L^2}^2>Ma_n\right)\le 2\varepsilon,
\]
which proves that
\(\|\hat\rho-\tilde\rho\|_{\mathcal L^2}^2=O_p(a_n)\) unconditionally.
Combining the above result with \eqref{rho_mean}, we have
\begin{eqnarray}\label{rho_t}
    \|\hat\rho-\rho\|_{\mathcal L^2}^2
=
O_p\!\left(\mathcal{R}^2(\hat V,V)+(n\bar J^2)^{-2/3}+n^{-1}\right).
\end{eqnarray}

\

\noindent\textbf{Consistency of data generation.}
Define
\begin{itemize}
    \item \( X(t) = \phi_{1,t}(Z(t)) \), where \( Z(\cdot) \) is a mean-zero Gaussian process on \( \mathcal{S} \) with covariance function \( \rho(t,s) \).
    \item \( \tilde{X}(t) = \hat{\phi}_{1,t}(\tilde{Z}(t)) \), where \( \tilde{Z}(t) \) is a mean-zero Gaussian process on \( \mathcal{S} \) with covariance function \( \hat{\rho}(t,s) \).
\end{itemize}

To bound \( W_2(X, \tilde{X}) \), we introduce an intermediate process
\begin{eqnarray*}
    Y(t) = \hat{\phi}_{1,t}({Z}(t))
\end{eqnarray*}
and apply the triangle inequality in the Wasserstein metric:
\[
W_2(X, \tilde{X}) \leq W_2(X, Y)+ W_2(Y, \tilde{X}).
\]
In the following, we {show} that $W_2(X, Y)$ and $W_2(Y, \tilde{X})$ are $o_p(1)$, which completes our proof.

\begin{itemize}
\item Consider \( {X}(t) = \phi_{1,t}( Z(t)) \) and \( Y(t) = \hat{\phi}_{1,t}(Z(t)) \), which share the {same} underlying process \(  Z(t) \). {Using the coupling \( (X, Y)  = (\phi_{1,\cdot}( Z(\cdot)), \hat\phi_{1,\cdot}( Z(\cdot))) \), the squared Wasserstein distance satisfies}
\[
W_2^2({X}, Y) \leq \mathbb{E}_{{Z}}  \|X - Y\|^2_{\mathcal{L}^2}
= \mathbb{E}_{{Z}} \left[ \int_{\mathcal{S}} |\phi_{1,t}( Z(t)) - \hat{\phi}_{1,t}(Z(t))|^2 \ \mathrm{d}t \right].
\]
Similar to \eqref{AG_fomula}, we apply the Alekseev--Gr\"obner formula and obtain
    \begin{eqnarray}\label{AG_inq}
    &&   \int_{\mathcal{S}} |\phi_{1,t}( Z(t)) - \hat{\phi}_{1,t}(Z(t))|^2 \, \mathrm{d}t\\
       &\leq &  \int_{\mathcal{S}}\int_0^1 \bigg(\frac{\mathrm{d}\hat \phi_{(u,1),t}(x)}{\mathrm{d}x}\bigg|_{x={\phi}_{u,t}({Z}(t))}\bigg)^2\nonumber\\
       &&\cdot \bigg(\hat{V}(u,t,\phi_{u,t}({Z}(t)))- {V}(u,t,{\phi}_{u,t}({Z}(t)))\bigg)^2\ \mathrm{d}u\,\mathrm{d}t,\nonumber
    \end{eqnarray}
where \( \hat\phi_{(u,v),t}(x) \) denotes the solution of \eqref{est_backward_ode} starting at \( u \) and ending at \( v \), with initial value \( x \).    
Notice that $\hat\phi_{(u,w),t}(x)=x+\int_{u}^w \hat V(v,t, \hat\phi_{(u,v),t}(x))\ \mathrm{d}v$, 
\begin{eqnarray*}
\frac{\mathrm{d}\hat \phi_{(u,w),t}(x)}{\mathrm{d}x}
=   1+\int_{u}^w{\frac{\partial \hat V(v,t,\xi)}{\partial \xi}}\bigg|_{\xi=\hat\phi_{(u,v),t}(x)}\cdot \frac{\mathrm{d}\hat \phi_{(u,v),t}(x)}{\mathrm{d}x}\ \mathrm{d}v.
\end{eqnarray*}
{Since $\hat{V}\in \mathcal{W}_1^{\infty}([0,1]\times\mathcal{S}\times\mathbb{R})$, there exists a constant $L_1$ such that}  
\(\frac{\mathrm{d}}{\mathrm{d}x}\hat V(u,t,x)\) is bounded by \(L_1\) for any $u$, $t$, and $x$.
Therefore,
\begin{eqnarray*}
\bigg|\frac{\mathrm{d}\hat \phi_{(u,w),t}(x)}{\mathrm{d}x} \bigg|
\leq 1 + L_1  \cdot \int_{u}^w\bigg|\frac{\mathrm{d}\hat \phi_{(u,v),t}(x)}{\mathrm{d}x}\bigg|\ \mathrm{d}v, 
\end{eqnarray*}
By Gr\"onwall's inequality,
\begin{eqnarray*}
    \bigg|\frac{\mathrm{d}\hat\phi_{(u,w),t}(x)}{\mathrm{d}x}\bigg|\leq \exp((w-u)L_1).
\end{eqnarray*}
Combine the above inequalities with the previous Alekseev--Gr\"obner formula \eqref{AG_inq},
\begin{eqnarray*}
W_2^2(X, Y)
&\leq &\mathbb{E}_{{Z}}\int_{\mathcal{S}} |\phi_{1,t}( Z(t)) - \hat{\phi}_{1,t}( Z(t))|^2\, \mathrm{d}t\\
       &\lesssim&\mathbb{E}_{{Z}} \int_{\mathcal{S}}\int_0^1  \bigg(\hat{V}(u,t,\phi_{u,t}({Z}(t)))- {V}(u,t, \phi_{u,t}({Z}(t)))\bigg)^2\ \mathrm{d}u\,\mathrm{d}t\\
       &=&\int_{\mathcal{S}} \int_0^1  \mathbb{E}_{{Z_{u}(t)}}\bigg(\hat{V}(u,t,Z_u(t))- {V}(u,t, Z_u(t))\bigg)^2\ \mathrm{d}u\,\mathrm{d}t\\
       &=&o_p(1),
\end{eqnarray*}
where the final {bound follows from} \eqref{target_error}.

\item Consider \( \tilde X(t) = \hat{\phi}_{1,t}(\tilde{Z}(t)) \) and \( Y(t) = \hat{\phi}_{1,t}({Z}(t)) \). From the flow equation:
\[
\frac{\partial \hat{\phi}_{u,t}(x)}{\partial u} = \hat{V}_{u,t}(\hat{\phi}_{u,t}(x)), \quad \hat{\phi}_{0,t}(x) = x,
\]
where \( \hat{V}_{u,t}(x) \) is Lipschitz in \( x \) with constant \( L_1 \).
Similar to \eqref{Gr_inequality}, we have that
\begin{eqnarray*}
    |\hat{\phi}_{1,t}(x) - \hat{\phi}_{1,t}(y)| \leq e^{L_1} |x - y|.
\end{eqnarray*}

Consider \( \tilde X(t) = \hat{\phi}_{1,t}(\tilde{Z}(t)) \) and \( Y(t) = \hat{\phi}_{1,t}({Z}(t)) \). From the flow equation:
\[
\frac{\partial \hat{\phi}_{u,t}(x)}{\partial u} = \hat{V}_{u,t}(\hat{\phi}_{u,t}(x)), \quad \hat{\phi}_{0,t}(x) = x,
\]
where \( \hat{V}_{u,t}(x) \) is Lipschitz in \( x \) with constant \( L_1 \).
Similar to \eqref{Gr_inequality}, we have
\begin{eqnarray*}
    |\hat{\phi}_{1,t}(x) - \hat{\phi}_{1,t}(y)| \leq e^{L_1} |x - y|.
\end{eqnarray*}

Let \(\mathcal K\) and \(\hat{\mathcal K}\) denote the integral operators associated with the kernels \(\rho(t,s)\) and \(\hat{\rho}(t,s)\), respectively, i.e.,
\[
(\mathcal K f)(t)=\int_{\mathcal S}\rho(t,s)f(s)\,\mathrm{d}s,
\qquad
(\hat{\mathcal K}f)(t)=\int_{\mathcal S}\hat\rho(t,s)f(s)\,\mathrm{d}s.
\]
Consider an optimal coupling \((Z,\tilde Z)\) of the mean-zero Gaussian processes with covariance operators \(\mathcal K\) and \(\hat{\mathcal K}\), respectively. Using the induced coupling
\[
(\tilde X,Y)
=
\bigl(
\hat{\phi}_{1,\cdot}(\tilde Z(\cdot)),
\hat{\phi}_{1,\cdot}(Z(\cdot))
\bigr),
\]
we have
\begin{eqnarray*}
W_2^2(\tilde X,Y)
&\leq&
\mathbb E\|\tilde X-Y\|_{\mathcal L^2}^2\\
&\leq&
e^{2L_1}\,
\mathbb E\|\tilde Z-Z\|_{\mathcal L^2}^2\\
&=&
e^{2L_1}\,W_2^2(\tilde Z,Z)\\
&=&
e^{2L_1}
\bigg[
\operatorname{trace}(\mathcal K)
+\operatorname{trace}(\hat{\mathcal K})
-2\operatorname{trace}\!\Big(
(\mathcal K^{1/2}\hat{\mathcal K}\mathcal K^{1/2})^{1/2}
\Big)
\bigg].
\end{eqnarray*}
Define
\[
f(\hat{\mathcal K},\mathcal K)
:=
\operatorname{trace}(\mathcal K)
+\operatorname{trace}(\hat{\mathcal K})
-2\operatorname{trace}\!\Big(
(\mathcal K^{1/2}\hat{\mathcal K}\mathcal K^{1/2})^{1/2}
\Big).
\]
By Theorem~8 in \cite{mallasto2017learning}, \( f(\hat{\mathcal{K}}, \mathcal{K}) =o_p(1) \) if \( \| \mathcal{K} - \hat{\mathcal{K}} \|_{\mathrm{op}} =o_p(1) \), where
\[
\|\mathcal K\|_{\mathrm{op}}
=
\sup_{\{f:\,\|f\|_{\mathcal L^2}\leq 1\}}
\|\mathcal Kf\|_{\mathcal L^2}.
\]
Notice that
\begin{eqnarray*}
\| \mathcal{K} - \hat{\mathcal{K}} \|_{\mathrm{op}}^2
&=&
\sup_{\{f:\,\|f\|_{\mathcal{L}^2} \leq 1\}}
\bigg\|\int_{\mathcal{S}} \big(\rho(\cdot,s) -\hat \rho(\cdot,s)\big) f(s) \, \mathrm{d}s\bigg\|_{\mathcal{L}^2}^2\\
&\leq&
\|\rho-\hat{\rho}\|_{\mathcal{L}^2}^2
=
o_p(1),
\end{eqnarray*}
where the inequality follows from Cauchy--Schwarz and the final equality is obtained from \eqref{rho_t}.
Combining the above results, we finally obtain
\[
W_2(\tilde X,Y) = o_p(1).
\]

\end{itemize}

\end{proof}

\subsection{Additional Lemmas}

\begin{LemmaS}\label{variable_equa}
Let \( X \) be a random variable with a continuous and strictly increasing distribution function on an interval \(\supp(X) \subseteq \mathbb{R}\), with \(\mathbb{P}(X \in \supp(X)) = 1\). Suppose \( T \) and \( S \) are two continuous and strictly increasing functions on \(\supp(X)\). If the distributions of \( T(X) \) and \( S(X) \) are the same, then
\[
T(X) = S(X)
\]
almost surely.
\end{LemmaS}

\begin{proof}
Let \( F_X \) denote the CDF of \( X \), which is continuous and strictly increasing on \(\supp(X)\). Define the intervals
\( I_T = \{ T(x) \mid x \in \supp(X) \} \) and \( I_S = \{ S(x) \mid x \in \supp(X) \} \).
Since \( T \) and \( S \) are continuous and strictly increasing, and \(\supp(X)\) is an interval, \( I_T \) and \( I_S \) are intervals. Given \(\mathbb{P}(X \in \supp(X)) = 1\), it follows that \(\mathbb{P}(T(X) \in I_T) = \mathbb{P}(S(X) \in I_S) = 1\).

For \( y \in I_T \), since \( T \) is strictly increasing, its inverse \( T^{-1} \) exists, and the event \( \{ T(X) \leq y \} \) is equivalent to \( \{ X \leq T^{-1}(y) \} \). Thus,
\[
\mathbb{P}(T(X) \leq y) = \mathbb{P}(X \leq T^{-1}(y)) = F_X(T^{-1}(y)).
\]
Similarly, for \( y \in I_S \),
\[
\mathbb{P}(S(X) \leq y) = \mathbb{P}(X \leq S^{-1}(y)) = F_X(S^{-1}(y)).
\]

{Since \( T(X) \) and \( S(X) \) have the same distribution, their supports coincide, and hence \(I_T=I_S\).}
Let \( I = I_T = I_S \). Thus,
\[
F_X(T^{-1}(y)) = F_X(S^{-1}(y)) \quad \text{for all } y \in I.
\]

Since \( F_X \) is strictly increasing, it is injective. Hence, \( F_X(T^{-1}(y)) = F_X(S^{-1}(y)) \) implies
\[
T^{-1}(y) = S^{-1}(y) \quad \text{for all } y \in I.
\]

Let \(\Omega_0 = \{ \omega \in \Omega \mid X(\omega) \in \supp(X) \}\), with \(\mathbb{P}(\Omega_0) = 1\). For each \(\omega \in \Omega_0\), set \( x = X(\omega) \in \supp(X) \). Then, \( T(x) \in I \), and we have
\[
S^{-1}(T(x)) = T^{-1}(T(x)) = x.
\]
Applying \(S\) to both sides yields \(T(x)=S(x)\).
Thus, \( T(X(\omega)) = S(X(\omega)) \) for all \(\omega \in \Omega_0\). Since \(\mathbb{P}(\Omega_0)=1\),
\[
T(X) = S(X) \quad \text{almost surely.}
\]
The proof is complete.
\end{proof}

\

\begin{LemmaS}[Picard--Lindel\"of Theorem in a Banach space]\label{le:PL}
Let $(E,\|\cdot\|_E)$ be a Banach space, $t_0 \in \mathbb{R}$, $x_0 \in E$, and let $f: E \times \mathbb{R} \to E$ satisfy:
\begin{itemize}
    \item For each $x \in E$, the function $t \mapsto f(x, t)$ is continuous.
    \item For each compact set $K \subset E$ and each compact interval $I \subset \mathbb{R}$, there exists $L \geq 0$ such that for all $x, y \in K$ and $t \in I$,
    \[
    \| f(x, t) - f(y, t) \|_E \leq L \| x - y \|_E.
    \]
\end{itemize}
Then, there exists $\delta > 0$ such that the initial value problem
\[
\frac{\mathrm{d}x}{\mathrm{d}t} = f(x, t), \quad x(t_0) = x_0
\]
has a unique solution $x: [t_0 - \delta, t_0 + \delta] \to E$.
\end{LemmaS}

See Theorem~3.A in \cite{zeidler1985nonlinear} for the detailed proof.

\

\begin{LemmaS}\label{le: continu_inver}
Let \( X \) be a random variable with distribution \( \mu \), where \( \mu \) is absolutely continuous with respect to the Lebesgue measure \( \lambda \) (i.e., \( \mu \ll \lambda \)), and \( \mu \) has a density \(f\) such that $f(x)>0$ for $\lambda$-a.e.\ $x\in \operatorname{supp}(X)$.
If $T(X) = S(X)$ almost surely, then $T = S$ almost everywhere in $\operatorname{supp}(X)$ with respect to the Lebesgue measure.
\end{LemmaS}

\begin{proof}
Suppose \( T(X) = S(X) \) almost surely, meaning
\[
\mu(\{x : T(x) \neq S(x)\}) = 0. 
\]
Define \( A = \{x \in \operatorname{supp}(X) : T(x) \neq S(x)\} \). Since \( A \subset \{x : T(x) \neq S(x)\} \) and \( \mu(\{x : T(x) \neq S(x)\}) = 0 \), it follows that
\[
\mu(A) = 0.
\]
Given \( \mu \ll \lambda \), we have \( \mu(A) = \int_A f(x) \, \mathrm{d}\lambda(x) \). Notice that \( A \subset \operatorname{supp}(X) \) and $f(x)>0$ for $\lambda$-a.e.\ $x\in \operatorname{supp}(X)$.
If \( \lambda(A) > 0 \), then
\[
\mu(A) = \int_A f(x) \, \mathrm{d}\lambda(x) {\,>\,0},
\]
because $\lambda(A)>0$ and $f>0$ $\lambda$-a.e.\ on $\operatorname{supp}(X)$ imply $\int_A f\,\mathrm{d}\lambda>0$.
This contradicts \( \mu(A) = 0 \). Hence, \( \lambda(A) = 0 \), proving that
\[
\lambda(\{x \in \operatorname{supp}(X) : T(x) \neq S(x)\}) = 0.
\]
Thus, \( T = S \) almost everywhere on \( \operatorname{supp}(X) \) with respect to the Lebesgue measure. 
\end{proof}

\

\begin{LemmaS}\label{assuption_bounded}
    Assumption~\ref{asum:density} indicates Condition~\ref{asum:Z}.
\end{LemmaS}

\begin{proof}
Fix $u\in[0,1]$ and $t\in\mathcal S$. Let $f_X(\cdot;t)$ and $p_{0,t}(\cdot)$ denote the densities of $X_i(t)$ and $Z_0(t)$, respectively. 
For any measurable $g:\mathbb R\to\mathbb R$, define
\[
\|g\|_{\infty}:=\sup_{x\in\mathbb R}|g(x)|,\qquad 
\|g\|_{1}:=\int_{\mathbb R}|g(x)|\,dx .
\]
Define the uniform bounds
\[
M_X:=\sup_{t\in\mathcal S}\|f_X(\cdot;t)\|_{\mathcal{W}_0^{\infty}(\mathbb{R})}<\infty,\qquad 
M_0:=\sup_{t\in\mathcal S}\|p_{0,t}\|_{\mathcal{W}_0^{\infty}(\mathbb{R})}<\infty.
\]

For $u\in(0,1)$, write
\[
Z_{u,i}(t)=uX_i(t)+(1-u)Z_0(t).
\]
Let
\[
g_{u,t}(z):=\frac1u f_X(z/u;t),\qquad 
h_{u,t}(z):=\frac1{1-u}p_{0,t}\bigl(z/(1-u)\bigr),
\]
which are the densities of $uX_i(t)$ and $(1-u)Z_0(t)$, respectively. Since $uX_i(t)$ and $(1-u)Z_0(t)$ are independent, $Z_{u,i}(t)$ has density
\[
p_{Z_u(t)}(x)=(g_{u,t}*h_{u,t})(x)=\int_{\mathbb R}g_{u,t}(x-y)h_{u,t}(y)\,\mathrm{d}y.
\]
Because $g_{u,t}(z)>0$ and $h_{u,t}(z)>0$ for all $z$, we have $p_{Z_u(t)}(x)>0$ for all $x$.
For the boundary cases, $p_{Z_0(t)}=p_{0,t}$ and $p_{Z_1(t)}=f_X(\cdot;t)$, which are positive by assumption.

For $u\in(0,1)$,
\[
\|g_{u,t}\|_\infty\le \frac{M_X}{u},\qquad 
\|h_{u,t}\|_\infty\le \frac{M_0}{1-u},\qquad 
\|g_{u,t}\|_1=\|h_{u,t}\|_1=1.
\]
Since $p_{Z_u(t)}=g_{u,t}*h_{u,t}$, Young's convolution inequality
yields
\[
\|p_{Z_u(t)}\|_\infty
\le {\min\left\{\frac{M_X}{u},\frac{M_0}{1-u}\right\}}
\le {M_X+M_0},
\]
uniformly over $u\in(0,1)$ and $t\in\mathcal S$. For $u=0,1$,
\[
\|p_{Z_0(t)}\|_\infty=\|p_{0,t}\|_\infty\le M_0,\qquad 
\|p_{Z_1(t)}\|_\infty=\|f_X(\cdot;t)\|_\infty\le M_X,
\]
so the same uniform bound holds over all $u\in[0,1]$ and $t\in\mathcal S$.
\end{proof}

\

\begin{LemmaS}\label{pro_1}
Let $\{X(t):t\in\mathcal{S}\}$ be a random process, and let $\phi_t:\mathbb{R}\to\mathbb{R}$, $t\in\mathcal{S}$, be strictly increasing functions. Define
$$
Y(t)=\phi_t\{X(t)\}, \qquad t\in\mathcal{S}.
$$
Then for any finite collection
$t_1,\dots,t_m\in\mathcal{S}$, the random vectors
$$
\bigl(X(t_1),\dots,X(t_m)\bigr)
\quad \text{and} \quad
\bigl(Y(t_1),\dots,Y(t_m)\bigr)
$$
have the same copula.
\end{LemmaS}

\begin{proof}
Fix any $t_1,\dots,t_m\in\mathcal{S}$, and write
$$
X_j=X(t_j), \qquad Y_j=Y(t_j)=\phi_{t_j}(X_j), \qquad j=1,\dots,m.
$$
Let $F_j$ and $G_j$ denote the marginal distribution functions of $X_j$ and $Y_j$, respectively. Since $\phi_{t_j}$ is strictly increasing, we have
$$
G_j(y)
=
\mathbb{P}(Y_j\le y)
=
\mathbb{P}\bigl(\phi_{t_j}(X_j)\le y\bigr)
=
\mathbb{P}\bigl(X_j\le \phi_{t_j}^{-1}(y)\bigr)
=
F_j\bigl(\phi_{t_j}^{-1}(y)\bigr).
$$
Therefore,
$$
G_j(Y_j)
=
F_j\bigl(\phi_{t_j}^{-1}(Y_j)\bigr)
=
F_j(X_j).
$$
Hence the probability integral transforms of $X_j$ and $Y_j$ coincide:
$$
\bigl(G_1(Y_1),\dots,G_m(Y_m)\bigr)
=
\bigl(F_1(X_1),\dots,F_m(X_m)\bigr)
\quad \text{a.s.}
$$
Thus
$$
C_Y(u_1,\dots,u_m)
:=
\mathbb{P}\bigl(G_1(Y_1)\le u_1,\dots,G_m(Y_m)\le u_m\bigr)
$$
coincides with
$$
C_X(u_1,\dots,u_m):=\mathbb{P}\bigl(F_1(X_1)\le u_1,\dots,F_m(X_m)\le u_m\bigr).
$$
Therefore,
$$
C_Y=C_X.
$$
Since the choice of $t_1,\dots,t_m$ is arbitrary, all finite-dimensional copulas are preserved. 
\end{proof}

\

\section{Additional Technical Details}

\subsection{Smooth Flow Matching for Student-\texorpdfstring{$t$}{t} Copula Processes}\label{sec: t-copula}

We extend SFM to more general bases, enabling the model to capture more complex dependencies, such as tail dependence, i.e., the tendency for extreme values to occur simultaneously within a process \citep{staicu2012modeling}. In particular, we go beyond Gaussian bases by adopting the Student-$t$ copula framework described in the following example. Additional extensions of SFM to broader copula structures are discussed in Part~\ref{sec: extend_SFM}.

A \(t\)-copula process is characterized by its degrees of freedom \(\nu\) and a latent correlation function \(\rho\). 

\begin{ExS}[Student-\(t\) Copula Process]\label{Ex: T-copula}\it
Denote $\text{T}_\nu$ as the Student-\(t\) distribution with degrees of freedom~\(\nu>0\).
  Let \(\mathbf{T}_{\nu,\rho}(\cdot; t_1,\ldots,t_m)\) be the joint distribution
  function of a mean-zero multivariate Student-\(t\) process with
  degrees of freedom~\(\nu>0\) and correlation function~\(\rho\),
  evaluated at \(t_1,\ldots,t_m\).
  Set $F_{t,\text{base}}=\text{T}_\nu$, $t\in \mathcal{S}$, and
  $H_{\text{base}}(\cdot;t_1,\ldots,t_m)\;=\;
      \mathbf{T}_{\nu,\rho}(\cdot; t_1,\ldots,t_m),
      \quad t_1,\ldots,t_m\in\mathcal{S}$.
  A stochastic process \( X(\cdot) \) constructed with this base is called a Student-\(t\) copula process.
  By the {properties} of the multivariate $t$-distribution, we have
 \begin{equation}\label{latent_correlation_t}
      \rho(t,s)=\frac{\nu-2}{\nu}\cdot \mathbb{E}\!\bigl\{\text{T}_\nu^{-1}\circ F_t(X(t))\cdot \text{T}_\nu^{-1}\circ {F_s}(X(s))\bigr\},
      \qquad t,s\in\mathcal{S},\; \nu>2.
  \end{equation}
When \(\nu\to\infty\), the \(t\)-copula process converges to the Gaussian copula process.
\end{ExS}

We provide the full procedure for smooth flow matching of Student-$t$ copula processes in Algorithm~\ref{algo:total_algorithm_t} of the Supplementary Materials.
The key modifications include:
\begin{itemize}
  \item
  For any pair of observation times \((T_{ij_1},T_{ij_2})\) we estimate
  \begin{eqnarray}
      \label{eq:T_rho}
&&    G^{(t\text{-copula})}_i(T_{ij_1},T_{ij_2})\nonumber\\
      &=&
      \frac{\nu-2}{\nu}\;
      \Bigl\{\text{T}_{\nu}^{-1}\!\circ\!\Phi\!\circ\!
            \hat{\psi}_{1,T_{ij_1}}\!\bigl(X_i(T_{ij_1})\bigr)\Bigr\}
      \;
      \Bigl\{\text{T}_{\nu}^{-1}\!\circ\!\Phi\!\circ\!
            \hat{\psi}_{1,T_{ij_2}}\!\bigl(X_i(T_{ij_2})\bigr)\Bigr\}
  \end{eqnarray}
  based on \eqref{latent_correlation_t}.
  Surface‐smoothing the collection
  \(\bigl\{G^{(t\text{-copula})}_i(T_{ij_1},T_{ij_2})\bigr\}_{i,j_1,j_2}\)
  yields a smooth estimator \(\hat{\rho}(t,s)\) analogous to \eqref{est_rho}.

  \item
  Let \(\tilde{V}_0^{(l)}\sim\mathbf{T}_{\nu,\hat{\rho}}\) be a mean-zero \(t\)-process with correlation \(\hat{\rho}\).  It can be simulated via
  $$\tilde{V}_0^{(l)}(t)
      \;=\;
      \frac{W^{(l)}(t)}{\sqrt{u^{(l)}/\nu}},
      \quad
      t\in\mathcal{S},$$
  where \(W^{(l)}\) is a zero-mean Gaussian process with covariance \(\hat{\rho}\),
  \(u^{(l)}\sim\chi^2_{\nu}\), and \(W^{(l)}\) is independent of \(u^{(l)}\).

  \item {Set
  \[
  \tilde{Z}_0^{(l)}(t)=\Phi^{-1}\!\circ\!\text{T}_{\nu}\bigl(\tilde{V}_0^{(l)}(t)\bigr),
  \qquad t\in\mathcal S.
  \]
  Then \(\tilde{Z}_0^{(l)}\) is the Gaussianized version of the latent \(t\)-process \(\tilde V_0^{(l)}\), preserving the copula structure determined by \(\hat\rho\), and it is used as the latent input in the subsequent flow-generation step.}
\end{itemize}

When \(\nu\to\infty\), Algorithm~\ref{algo:total_algorithm_t} reduces to the Gaussian‐base version in Algorithm~\ref{algo:total_algorithm}.
Practical estimation of \(\nu\) from data is discussed in Part~\ref{sec: t-copula_nu}.

\begin{algorithm}[ht!]
\renewcommand{\thealgorithm}{S1}
\caption{Smooth Flow Matching for Student-$t$ Copula Processes}
\label{algo:total_algorithm_t}
\small
\begin{algorithmic}[1]
\State \textbf{Input:} Observed data $\{(X_i(T_{ij}), T_{ij}); i=1,\ldots,n, j=1,\ldots,J_i\}$, number of samples $H$ for estimation, number of samples $M$ for generation, and the degree of freedom $\nu>2$. A regularly spaced time grid $\mathcal{S}_{r}$ such that $\cup_{i=1}^n\big\{T_{ij}; j\in [J_i]\big\}\subset \mathcal{S}_{r}$.
\State Generate $H$ independent and identical mean-zero Gaussian processes, $Z_0^{(h)}$, $h=1,\ldots,H$, sampled at the time grid $\mathcal{S}_r$, with the covariance function being $\mathbb{I}(t=s)$.
\State Obtain the vector field $\hat{V}$ from the generated data $Z_0^{(h)}(t)$, \( h = 1, \ldots, H \), and \(t\in  \mathcal{S}_r \):
\begin{align*}
    \hat{V}:=\operatorname{argmin}_{U\in {\mathcal{B}_{L_V,4}([0,1],\mathcal{S},\mathcal{X})}} \big\{\hat{\mathcal{L}}_n(U)&+ {\mathcal{J}(U)}\big\}.
\end{align*}
\State  Calculate $G^{(t\text{-copula})}_i(T_{ij_1},T_{ij_2})$:
\begin{eqnarray*}
  G^{(t\text{-copula})}_i(T_{ij_1},T_{ij_2})
  = \frac{\nu-2}{\nu} \cdot 
  \text{T}_{\nu}^{-1}\circ\Phi\circ \hat{\psi}_{1,T_{ij_1}}\big( X_i(T_{ij_1})\big)
  \cdot 
  \text{T}_{\nu}^{-1}\circ\Phi\circ\hat{\psi}_{1,T_{ij_2}}\big( X_i(T_{ij_2})\big),
\end{eqnarray*}
where $\hat{\psi}_{1,t}(z)$ is the solution of
\[
\frac{\partial \hat{\psi}_{u,t}(z)}{\partial u} = -\hat{V}(1-u,t,\hat{\psi}_{u,t}(z)),
\qquad
\text{subject to}\quad \hat{\psi}_{0,t}(z) = z .
\]
\State Apply surface smoothing on \( \{G^{(t\text{-copula})}_i(T_{ij_1},T_{ij_2}); i = 1, \dots, n, j_1, j_2 = 1, \dots, J_i\} \) to obtain \( \hat{\rho}(t,s) \).
\State Regenerate $M$ {independent} mean-zero Gaussian processes, $\tilde{W}_0^{(l)}$, $l=1,\ldots,{M}$, sampled at the time grid $\mathcal{S}_r$, with the covariance function being $\hat{\rho}(t,s)$, and generate \( {\xi^{(l)}} \sim \chi^2_{\nu} \), $l=1,\ldots,{M}$.
\State For $l=1,\ldots,{M}$ and $t\in \mathcal{S}_r$, solve the equation
\begin{eqnarray*}
\frac{\partial \tilde{Z}_u^{(l)}(t)}{\partial u}
=
\hat{V}(u,t,\tilde{Z}^{(l)}_u(t))
\end{eqnarray*}
given the initial value
\[
\tilde{Z}_0^{(l)}(t)=\Phi^{-1}\circ \text{T}_{\nu}\bigg(\frac{\tilde{W}^{(l)}_0(t)}{\sqrt{{\xi^{(l)}} / \nu}}\bigg),
\]
obtaining $\tilde{Z}_1^{(l)}(t)$ for each $t\in \mathcal{S}_r$.
\State \textbf{Output:} $\tilde{Z}_1^{(l)}(t)$, $t\in \mathcal{S}_r$ and $l=1,\ldots,{M}$.
\end{algorithmic}
\end{algorithm}

\subsection{Extensions of Smooth Flow Matching to General Copula}\label{sec: extend_SFM}
We next show how SFM can accommodate a general class of copula processes.
By Sklar’s theorem \citep[][see Supplementary Section~\ref{sec: sklar}]{jaworski2010copula}, the joint density of the irregular observations
\(\bigl\{X_i(T_{i1}),\ldots,X_i(T_{iJ_i})\bigr\}\) satisfies
\begin{equation}\label{eq:Sklar}
  p_{X_i(T_{i1}),\ldots,X_i(T_{iJ_i})}(x_1,\ldots,x_{J_i})
  \;=\;
  c^{d}_{T_{i1},\ldots,T_{iJ_i}}\!\bigl(
    F_{T_{i1}}(x_1),\ldots,F_{T_{iJ_i}}(x_{J_i})
  \bigr)\,
  \prod_{j=1}^{J_i} {f_X(x_j;T_{ij})},
\end{equation}
where \(F_{T_{ij}}\) and {\(f_X(\cdot;T_{ij})\)} are the marginal CDF and density, while
\(c^{d}_{T_{i1},\ldots,T_{iJ_i}}\) is the copula density
\(
  \partial^{J_i}c_{T_{i1},\ldots,T_{iJ_i}}/\partial u_1\cdots\partial u_{J_i}.
\)
For any \(m\)-tuple \((t_1,\ldots,t_m)\subset\mathcal{S}\), we express the copula
\(c_{t_1,\ldots,t_m}\) via a base joint distribution
\(H_{\text{base}}\) based on Definition~\ref{Def_copula_process}:
\[
  c_{t_1,\ldots,t_m}(u_1,\ldots,u_m)
  \;=\;
  H_{\text{base}}\!\bigl(
    F^{-1}_{t_1,\text{base}}(u_1),\ldots,
    F^{-1}_{t_m,\text{base}}(u_m);\,t_1,\ldots,t_m
  \bigr),
\]
where \(F_{t,\text{base}}\) denotes the marginal base cumulative distribution function.

Let \(H_{\text{base},\theta}\) be a parametric family indexed by
\(\theta\in\Theta\).
Combining \eqref{eq:Sklar} with the above representation yields the log-likelihood of \(\theta\):
\begin{equation}\label{eq:lik}
\begin{aligned}
 & \ell\!\Bigl(\{X_i(T_{ij})\};\theta\Bigr)\\
  &=
  \sum_{i=1}^{n}
  \log\!
  \biggl[
     \frac{\partial^{J_i}
       H_{\text{base},\theta}\!\bigl(
         F^{-1}_{T_{i1},\text{base}}(u_1),\ldots,
         F^{-1}_{T_{iJ_i},\text{base}}(u_{J_i});\,
         T_{i1},\ldots,T_{iJ_i}
       \bigr)}
       {\partial u_1\cdots\partial u_{J_i}}
  \biggr]_{\,
        u_j = F_{T_{ij}}\bigl(X_i(T_{ij})\bigr)
      } \\
      &=
  \sum_{i=1}^{n}
  \log\!
  \biggl[
     \frac{\partial^{J_i}
       H_{\text{base},\theta}\!\bigl(
         z_1,\ldots,
         z_{J_i};\,
         T_{i1},\ldots,T_{iJ_i}
       \bigr)}
       {\partial z_1\cdots\partial z_{J_i}}
  \biggr]_{\,
        z_j = F^{-1}_{T_{ij},\text{base}}\circ F_{T_{ij}}\bigl(X_i(T_{ij})\bigr)
      }
      +C,
\end{aligned}
\end{equation}
where \(C\) contains the terms independent of \(\theta\).
Hence \(\ell\) depends on \(\theta\) and the unknown transformations
\(F^{-1}_{t,\text{base}}\!\circ F_t\), \(t\in\mathcal{S}\).
Estimating \(\theta\) while treating the latter nonparametrically constitutes a
semiparametric copula problem \citep{jaworski2010copula}.

Therefore, we introduce the following practical estimation steps.
\begin{itemize}
  \item[1.] Specify marginals: Fix base process \(F_{t,\text{base}}\) (e.g., Gaussian) for every \(t\in\mathcal{S}\).
  \item[2.] Estimate transformations: Using the smooth vector field from~\eqref{est_vecfil}, solve the backward ordinary differential equation in~\eqref{est_forward_ode} with
        initial value \(X_i(T_{ij})\) to obtain
        \(F^{-1}_{T_{ij},\text{base}} \circ F_{T_{ij}}\bigl(X_i(T_{ij})\bigr)\).
  \item[3.] Estimate copula parameter: Substitute these quantities into~\eqref{eq:lik} and maximize over
        \(\theta\) to obtain \(\hat{\theta}\).
  \item[4.] Generate new samples: Draw \(\tilde{Z}_0 \sim H_{\text{base},\hat{\theta}}\), and solve the
        forward ordinary differential equation~\eqref{est_backward_ode} to simulate new functional data.
\end{itemize}
We summarize the entire workflow in Algorithm~\ref{algo:total_algorithm_general}.

\begin{algorithm}[ht!]
\renewcommand{\thealgorithm}{S2}
\caption{Smooth Flow Matching for General Copula Processes}
\label{algo:total_algorithm_general}
\small
\begin{algorithmic}[1]
\State \textbf{Input:} Observed data $\{(X_i(T_{ij}), T_{ij});\, i=1,\ldots,n,\ j=1,\ldots,J_i\}$, number of samples $H$ for estimation, number of samples $M$ for generation, and base marginals $\{F_{t,\text{base}}:\ t\in \mathcal{S}\}$.
\State \textbf{Input:} A regularly spaced time grid $\mathcal{S}_{r}$ such that $\cup_{i=1}^n\big\{T_{ij};\, j\in [J_i]\big\}\subset \mathcal{S}_{r}$.
\State Generate \(H\) i.i.d.\ base processes $Z_0^{(h)}(t)\sim F_{t,\text{base}}$, for \(h = 1, \ldots, H\) and \(t\in \mathcal{S}_r\).
\State Obtain the vector field $\hat{V}$ from the generated data $\{Z_0^{(h)}(t)\}_{h=1}^H$ on $\mathcal{S}_r$:
\begin{align*}
    \hat{V}:=\operatorname{argmin}_{U\in {\mathcal{B}_{L_V,4}([0,1],\mathcal{S},\mathcal{X})}} \big\{\hat{\mathcal{L}}_n(U)+ {\mathcal{J}(U)}\big\}.
\end{align*}
\State Estimate the copula parameter $\hat{\theta}$ by
\begin{eqnarray*}
\hat{\theta}
=
\argmax_{\theta}\sum_{i=1}^{n}
  \log\!
  \biggl[
     \frac{\partial^{J_i}
       H_{\text{base},\theta}\!\bigl(
         z_1,\ldots,
         z_{J_i};\,
         T_{i1},\ldots,T_{iJ_i}
       \bigr)}
       {\partial z_1\cdots\partial z_{J_i}}
  \biggr]_{\,
        z_j = \hat{\psi}_{1,T_{ij}}\!\bigl(X_i(T_{ij})\bigr)
      },
\end{eqnarray*}
where $\hat{\psi}_{1,t}(z)$ solves
$\frac{\partial \hat{\psi}_{u,t}(z)}{\partial u} = -\hat{V}(1-u,t,\hat{\psi}_{u,t}(z))$
subject to $\hat{\psi}_{0,t}(z) = z$.
\State Generate $M$ i.i.d.\ base processes $\tilde{Z}_0^{(l)}$, $l=1,\ldots,M$, sampled on $\mathcal{S}_r$, from the joint distribution $H_{\text{base},\hat{\theta}}$.
\State For $l=1,\ldots,M$ and $t\in \mathcal{S}_r$, solve
\[
\frac{\partial \tilde{Z}_u^{(l)}(t)}{\partial u}=\hat{V}(u,t,\tilde{Z}^{(l)}_u(t)),
\]
with initial value $\tilde{Z}_0^{(l)}(t)$, obtaining $\tilde{Z}_1^{(l)}(t)$ for each $t\in \mathcal{S}_r$.
\State \textbf{Output:} $\tilde{Z}_1^{(l)}(t)$ for $t\in \mathcal{S}_r$ and $l=1,\ldots,M$.
\end{algorithmic}
\end{algorithm}

\subsection{Degrees of Freedom in Student \texorpdfstring{\( t \)}{t}-Copula Processes}\label{sec: t-copula_nu}

In this section, we focus on estimating the degrees of freedom \( \nu \) for a \( t \)-copula process, based on the observed data \( \{(X_i(T_{ij}), T_{ij}) : i = 1, \ldots, n,\; j = 1, \ldots, J_i\} \). Since the observed time points \( T_{ij} \) vary across subjects, we first transform the time points by rounding them to an equally spaced dense grid over the interval \([0,1]\) that includes all observed time points \( \{T_{ij} : i = 1, \ldots, n;\ j = 1, \ldots, J_i\} \). For each subject \( i \) and each time point \( t \) on this grid, we preprocess the observed data by: (1) assigning the value \( X_i(T_{ij}) \) if \( T_{ij} = t \); or, if no such value exists, (2) marking the data as missing. This results in \( n \) random vectors with missing entries, which are then used to estimate \( \nu \) based on the partially observed data.

In the following, we estimate the degrees of freedom \( \nu \) from multivariate \( t \)-copula data, following the method in \cite{demarta2005t}.
Let \( \{ \boldsymbol{x}_i \}_{i=1}^n \) denote \( n \) independent observations of a \( d \)-dimensional random vector, where each \( \boldsymbol{x}_i = (x_i^{(1)}, \ldots, x_i^{(d)}) \in \mathbb{R}^d \) may contain missing entries.  
We assume that \( \boldsymbol{x}_i \) follows a multivariate \( t \)-copula with base distribution \( \mathbf{T}_{\nu,\bm{\Sigma}} \), where \( \nu \) {is the degrees-of-freedom parameter} and \( \bm{\Sigma} = (\rho_{jk})_{j,k=1}^d \) is the correlation matrix.  
To handle the presence of missing data, we construct a composite pairwise likelihood to estimate the degrees of freedom \( \nu \).

For each pair of variables \( (j,k) \) with \( 1 \leq j < k \leq d \), define the index set
\[
\mathcal{I}_{jk} = \left\{ i \in \{1,\ldots,n\} : x_i^{(j)} \text{ and } x_i^{(k)} \text{ are both observed} \right\}.
\]
To construct the pairwise composite likelihood, we first transform the data to the copula scale. For each variable \( j \), we use the empirical cumulative distribution function:
\[
\hat{F}_j(x) = \frac{1}{n_j} \sum_{i=1}^n \mathbb{I}\{x_i^{(j)} \leq x,\ x_i^{(j)} \text{ observed} \}, 
\quad 
n_j = \left| \{ i : x_i^{(j)} \text{ observed} \} \right|.
\]
Then, for each observation \( i \in \mathcal{I}_{jk} \), we define the pseudo-observations:
\[
u_i^{(j)} = \hat{F}_j\!\big(x_i^{(j)}\big), 
\quad 
u_i^{(k)} = \hat{F}_k\!\big(x_i^{(k)}\big).
\]
The pairwise composite log-likelihood is then {constructed as}
\[
\ell_{\mathrm{PCL}}\!\big(\nu,(\rho_{jk})_{j,k=1}^d \big) 
= 
\sum_{1 \leq j < k \leq d} \ \sum_{i \in \mathcal{I}_{jk}} 
\log c_{\nu} \!\left( u_i^{(j)}, u_i^{(k)}; \rho_{jk}, \nu \right),
\]
where \( c_{\nu}(u, v; \rho, \nu) \) is the density of the bivariate $t$-copula with correlation \( \rho \) and degrees of freedom \( \nu \). This density is given by
\[
c_{\nu}(u,v;\rho,\nu) 
= 
\frac{\mathbf{t}_{\nu,\rho}\!\big(\text{T}_\nu^{-1}(u), \text{T}_\nu^{-1}(v)\big)}
{\text{t}_\nu\!\big(\text{T}_\nu^{-1}(u)\big)\cdot \text{t}_\nu\!\big(\text{T}_\nu^{-1}(v)\big)},
\]
where \( \text{t}_\nu(\cdot) \) is the density of the univariate $t$-distribution with \( \nu \) degrees of freedom, and \( \mathbf{t}_{\nu,\rho}(\cdot,\cdot) \) is the bivariate $t$-density with degrees of freedom \( \nu \) and correlation \( \rho \).
See Equation~(6) in \cite{demarta2005t} for details.

To alleviate computation, we follow \cite{demarta2005t} and estimate \( \rho_{jk} \) using Kendall's tau correlation. Define Kendall's tau \( \hat{\tau}_{jk} \) between variables \( j \) and \( k \) as
\[
\hat{\tau}_{jk} 
= 
\frac{2}{|\mathcal{I}_{jk}| (|\mathcal{I}_{jk}| - 1)} 
\sum_{1 \leq r < s \leq |\mathcal{I}_{jk}|} 
\operatorname{sign}\!\left( (x_r^{(j)} - x_s^{(j)})(x_r^{(k)} - x_s^{(k)}) \right),
\]
where the sum is taken over all distinct pairs of observations in \( \mathcal{I}_{jk} \), and \( \operatorname{sign}(\cdot) \) denotes the sign function.  
According to Equation~(9) in \cite{demarta2005t}, the correlation \( \rho_{jk} \) is related to Kendall's tau by
\[
\rho_{jk} = \sin\left( \frac{\pi}{2}\, \mathbb{E}[\hat{\tau}_{jk}] \right).
\]
Thus, an estimate of \( \rho_{jk} \) can be obtained as
\[
\hat{\rho}_{jk} = \sin\left( \frac{\pi}{2}\, \hat{\tau}_{jk} \right).
\]
By plugging in these estimates, the degrees of freedom \( \nu \) are obtained by maximizing the composite log-likelihood:
\[
\hat{\nu} = \argmax_{\nu > 2} \ \ell_{\mathrm{PCL}}\big(\nu, (\hat{\rho}_{jk})_{j,k=1}^d \big).
\]

\

\subsection{Sklar's Theorem}\label{sec: sklar}

Sklar's theorem is a fundamental result in copula theory, providing a link between multivariate distribution functions and their univariate margins.

\begin{TheoremS}[Sklar's Theorem]\label{thm:sklar}
Let $H$ be an $n$-dimensional joint cumulative distribution function (CDF) with marginal CDFs $F_1,\ldots,F_n$. Then there exists a copula $C:[0,1]^n\to[0,1]$ such that, for all $x_1,\ldots,x_n\in\mathbb{R}$,
\[
H(x_1,\ldots,x_n)=C\bigl(F_1(x_1),\ldots,F_n(x_n)\bigr).
\]
If each $F_i$ is continuous, then $C$ is unique. Conversely, if $C$ is a copula and $F_1,\ldots,F_n$ are univariate CDFs, then the function $H$ defined above is a joint CDF with marginals $F_1,\ldots,F_n$.
\end{TheoremS}

Assume that $H$ is absolutely continuous with joint density $h$, and that each marginal $F_i$ is absolutely continuous with density $f_i$. Then, under regularity conditions, $C$ is differentiable and Sklar's theorem can be written in terms of densities as
\[
h(x_1,\ldots,x_n)
=
c\bigl(F_1(x_1),\ldots,F_n(x_n)\bigr)\,\prod_{i=1}^n f_i(x_i),
\]
where the copula density $c$ is defined by
\[
c(u_1,\ldots,u_n)
=
\frac{\partial^n}{\partial u_1\cdots \partial u_n}\,C(u_1,\ldots,u_n).
\]

\

\subsection{Restricted Maximum Likelihood}\label{sec: REML}
Recall that the loss function in \eqref{est_vecfil} is defined as
\begin{eqnarray}\label{ori_loss}
  &&  \frac{1}{nHF}\sum_{i=1}^{n}\frac{1}{J_i}\sum_{j=1}^{J_i}
\sum_{h=1}^{H}\sum_{f=1}^{F}
\Bigl[X_i(T_{ij})-Z_0^{(h)}(T_{ij})-
U\bigl(u_f,T_{ij},Z^{(h)}_{u_f,i}(T_{ij})\bigr)\Bigr]^2\nonumber\\
&&\quad+
\int_{\mathcal{X}}\int_{\mathcal{S}} \int_0^1 
\bigg\{
\lambda_u\left( \frac{\partial^2 U}{\partial u^2}\right)^2
+\lambda_t\left( \frac{\partial^2 U}{\partial t^2}\right)^2
+\lambda_x\left( \frac{\partial^2 U}{\partial x^2}\right)^2
\bigg\}\ \mathrm{d}u\,\mathrm{d}t\,\mathrm{d}x .
\end{eqnarray}
Denote the tensor-product basis for $U$ as
\(\{\psi_k(u,t,x)\}_{k=1}^{K}\), and write
\[
U(u,t,x)=\sum_{k=1}^{K}\beta_k\,\psi_k(u,t,x),
\qquad
\boldsymbol\beta=(\beta_1,\dots,\beta_K)^{\!\top}.
\]
Stack every residual
\(
y_{ijhf}=X_i(T_{ij})-Z_0^{(h)}(T_{ij})
\)
into
\(
\mathbf y\in\mathbb R^{m},
\)
with $m=\sum_{i=1}^{n}J_iHF$.
Define the design matrix
\[
\mathbf W_{(ijhf),k}=\psi_k\!\bigl(u_f,T_{ij},
              Z^{(h)}_{u_f,i}(T_{ij})\bigr),
\]
and introduce the diagonal weight matrix
\[
\mathbf D=\operatorname{diag}(d_{ijhf}),\;
d_{ijhf}=1/(nHFJ_i).
\]
Utilizing the above notation, we formulate \eqref{ori_loss} as
\begin{eqnarray}\label{mody_loss}
\Bigl\|
\mathbf D^{1/2}\!\bigl(\mathbf y-\mathbf W\boldsymbol\beta\bigr)
\Bigr\|^{2}
\;+\;
\boldsymbol\beta^{\!\top}
\bigl(\lambda_uR_u+\lambda_tR_t+\lambda_xR_x\bigr)\boldsymbol\beta,
\end{eqnarray}
where the penalty matrices \(R_u,R_t,R_x\) are given by
\[
\begin{aligned}
(R_u)_{kl}
&=\int_{\mathcal{X}}\!\int_{\mathcal{S}}\!\int_{0}^{1}
\frac{\partial^{2}\psi_k}{\partial u^{2}}(u,t,x)\,
\frac{\partial^{2}\psi_l}{\partial u^{2}}(u,t,x)\,
\mathrm{d}u\,\mathrm{d}t\,\mathrm{d}x, \\[6pt]
(R_t)_{kl}
&=\int_{\mathcal{X}}\!\int_{\mathcal{S}}\!\int_{0}^{1}
\frac{\partial^{2}\psi_k}{\partial t^{2}}(u,t,x)\,
\frac{\partial^{2}\psi_l}{\partial t^{2}}(u,t,x)\,
\mathrm{d}u\,\mathrm{d}t\,\mathrm{d}x, \\[6pt]
(R_x)_{kl}
&=\int_{\mathcal{X}}\!\int_{\mathcal{S}}\!\int_{0}^{1}
\frac{\partial^{2}\psi_k}{\partial x^{2}}(u,t,x)\,
\frac{\partial^{2}\psi_l}{\partial x^{2}}(u,t,x)\,
\mathrm{d}u\,\mathrm{d}t\,\mathrm{d}x.
\end{aligned}
\]

Consider the model
\[
\mathbf y\mid\boldsymbol\beta,\sigma^{2}
\;\sim\;
\mathcal N\!\bigl(\mathbf W\boldsymbol\beta,\;
\sigma^{2}\mathbf D^{-1}\bigr)
\]
with the prior
\[
\boldsymbol\beta\mid\sigma^{2},\lambda_u,\lambda_t,\lambda_x
\;\sim\;
\mathcal N\!\Bigl(\mathbf 0,\;
\sigma^{2}
\bigl[\lambda_uR_u+\lambda_tR_t+\lambda_xR_x\bigr]^{-1}\Bigr).
\]
The negative log-posterior of \( \boldsymbol{\beta} \), up to an additive constant, is the same as \eqref{mody_loss}; hence, the original minimizer coincides with the maximum a posteriori (MAP) estimator in the Bayesian model.

To obtain the tuning parameters $\lambda:=\{\lambda_u,\lambda_t,\lambda_x\}$, we integrate out \(\boldsymbol\beta\) and obtain
\[
\mathbf y \mid \sigma^{2},\lambda
\;\sim\;
\mathcal N\!\bigl(\mathbf 0,\;
\sigma^{2}\mathbf V(\lambda)\bigr),
\]
where
\[
\mathbf V(\lambda)=
\mathbf D^{-1}+
\mathbf W\,\mathbf G(\lambda)\,\mathbf W^{\!\top},
\qquad
\mathbf G(\lambda)=
\bigl[\lambda_uR_u+\lambda_tR_t+\lambda_xR_x\bigr]^{-1}.
\]
The likelihood for $\lambda$ and $\sigma^{2}$ is then
\[
\ell_{\mathrm R}(\lambda_u,\lambda_t,\lambda_x,\sigma^{2})
=-\tfrac12\Bigl[
m\log\sigma^{2}
+\log|\mathbf V(\lambda)|
+\tfrac{1}{\sigma^{2}}\mathbf y^{\!\top}\mathbf V^{-1}(\lambda)\mathbf y
\Bigr].
\]
Profiling out \(\sigma^{2}\) yields
\[
\hat{\sigma}^{2}(\lambda)=
\frac{\mathbf y^{\!\top}\mathbf V(\lambda)^{-1}\mathbf y}{m},
\]
which leads to the profile likelihood of $\lambda$:
\[
\tilde\ell_{\mathrm R}(\lambda_u,\lambda_t,\lambda_x)=
-\tfrac12\Bigl[
m\log\bigl\{\mathbf y^{\!\top}\mathbf V^{-1}(\lambda)\mathbf y\bigr\}
+\log|\mathbf V(\lambda)|
\Bigr].
\]
Maximizing the above likelihood gives the REML estimates
\(\hat\lambda_u,\hat\lambda_t,\hat\lambda_x\).

\

\subsection{Gaussian-based Sampling for Functional Data}

As Gaussian-based benchmarks, we consider two approaches for generating functional data from estimated second-order structure.

\paragraph*{GP (Gaussian process sampling)}
Let \(X(t)\) denote the functional data on \(t\in\mathcal{S}\). We first estimate the mean function \(\mu(t)=\mathbb{E}\{X(t)\}\) and covariance function
\[
G(s,t)=\mathrm{Cov}\{X(s),X(t)\}
\]
using functional data approaches \citep{yao2005functional, hsing2015theoretical}. Denote the resulting estimators by \(\hat\mu(t)\) and \(\hat G(s,t)\). We then generate a new random trajectory \(\tilde X(t)\) from the fitted Gaussian process (GP)
\[
\tilde X(\cdot)\sim GP\bigl(\hat\mu(\cdot), \hat G(\cdot,\cdot)\bigr).
\]
In practice, on a finite grid \(t_1,\dots,t_m\), this amounts to sampling
\[
\bigl(\tilde X(t_1),\dots,\tilde X(t_m)\bigr)^\top
\sim
\mathcal{N}\bigl(\hat{\bm\mu}, \hat{\Sigma}\bigr),
\]
where
\[
\hat{\bm\mu}
=
\bigl(\hat\mu(t_1),\dots,\hat\mu(t_m)\bigr)^\top,
\qquad
\hat{\Sigma}
=
\bigl(\hat G(t_j,t_k)\bigr)_{1\le j,k\le m}.
\]
The sampled values are then treated as a generated function over the grid.

\paragraph*{KL (Karhunen--Lo\`eve expansion with FPCA)}
As another Gaussian-based benchmark, we apply functional principal component analysis (FPCA) to estimate the eigenvalues and eigenfunctions of the covariance operator \citep{yao2005functional, hsing2015theoretical}. Specifically, let
$$
\hat G(s,t)=\sum_{k\ge 1}\hat\lambda_k \hat\phi_k(s)\hat\phi_k(t)
$$
be the spectral decomposition of the estimated covariance function, where $\hat\lambda_k$ and $\hat\phi_k$ are the estimated eigenvalues and eigenfunctions, respectively. We then generate a new trajectory through the truncated Karhunen--Lo\`eve expansion
$$
\tilde X(t)=\hat\mu(t)+\sum_{k=1}^{K}\sqrt{\hat\lambda_k}\,Z_k\,\hat\phi_k(t),
\qquad
Z_k\stackrel{\mathrm{i.i.d.}}{\sim}\mathcal{N}(0,1),
$$
where $K$ is selected as in \citep{yao2005functional}.
That is, KL sampling reconstructs each generated function by combining the estimated mean with the leading empirical eigenfunctions and randomly sampled Gaussian principal component scores.

\

\subsection{Regression Procedure for Data Prediction}\label{sec: RPDP}
In real data analysis, we estimate \( f \) by minimizing the penalized squared error loss:
\[
\hat{f}
=
\argmin_{f\in \mathcal{B}_{L,4}(\mathcal{X},\mathcal{S})}
\sum_{i\in \mathcal{E}} \sum_{j=1}^{29}
\big\{ X_i(T_{j+1}) - f(X_i(T_j), T_j) \big\}^2\cdot \mathcal{I}_{T_j,T_{j+1}}(X_i)
+\mathcal{J}(f),
\]
where \( \mathcal{E} \) indexes the training set,
\( \mathcal{I}_{T_j,T_{j+1}}(X_i)=1 \) if \( X_i \) is observed at both \( T_j \) and \( T_{j+1} \), and \( 0 \) otherwise.
We set \( L=5 \) for the spline basis, and \(\mathcal{J}(f)\) denotes a roughness penalty {analogous to} \eqref{est_rho}.
We estimate \(\hat{f}\) using the \texttt{R} package \texttt{mgcv} \citep{wood2015package}.

\

\section{Supporting Results}\label{sec: ID}

\subsection{Sensitivity Analysis}

\subsubsection{\texorpdfstring{Tuning Parameters \( \lambda_u \), \( \lambda_t \), and \( \lambda_x \)}{Tuning Parameters λᵤ, λₜ, and λₓ}}\label{sen_tuning} 

In this part, we quantify how the estimator $\hat{V}$ behaves across different choices of tuning parameters \( \lambda_u \), \( \lambda_t \), and \( \lambda_x \). 
We separate this assessment into two aspects:

\noindent\textbf{Bias vs.\ Variance:} To quantify the bias and variance of the estimated vector field $\hat{V}$ using Algorithm~\ref{algo:total_algorithm}, we compute the flow map $\hat{\phi}_{1,t}(x)$ by solving the differential equation
\[
\frac{\partial \hat{\phi}_{u,t}(x)}{\partial u}
= \hat{V}\big(u,t,\hat{\phi}_{u,t}(x)\big),
\qquad 
\hat{\phi}_{0,t}(x)=x.
\]
We then evaluate the bias and variance using the median values of
\[
\bigg|\mathbb{E}\hat{\phi}_{1,t}(x) - \phi_{1,t}(x)\bigg|
\ \text{and}\
\sqrt{\operatorname{Var}\!\big(\hat{\phi}_{1,t}(x)\big)},
\]
computed over dense grids of $t \in [0,1]$ and $x \in [-3,3]$, where $\phi_{1,t}(x)$ is the true forward map and the interval $[-3,3]$ contains the majority of samples generated from a standard normal distribution (the base used in SFM). Similarly, the MSE of the vector field estimates is defined by $\mathbb{E}\big|\hat{\phi}_{1,t}(x) - \phi_{1,t}(x)\big|^2$.
{These expectations and variances are approximated using $100$ simulation replications.}
We report the median values of
\(
\big|\mathbb{E}\{\hat{\phi}_{1,t}(x)\}-\phi_{1,t}(x)\big|,\
\sqrt{\operatorname{Var}\!\big(\hat{\phi}_{1,t}(x)\big)},
\ \text{and}\
\mathbb{E}\big|\hat{\phi}_{1,t}(x)-\phi_{1,t}(x)\big|^2
\)
over dense grids of $t\in[0,1]$ and $x\in[-3,3]$ as the BIAS, VAR, and MSE, respectively.

\begin{figure}[ht!]
    \centering
    \includegraphics[scale = 0.65]{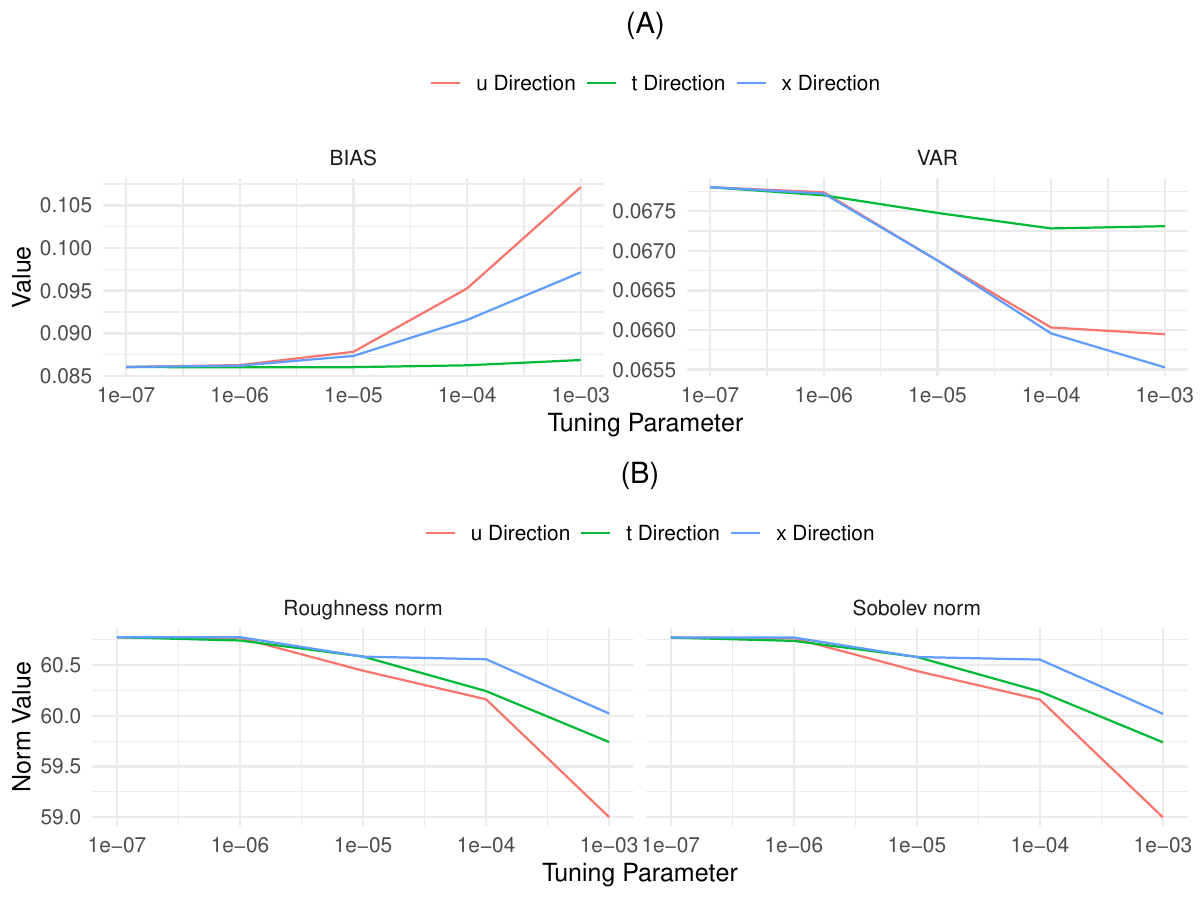}
    \caption{
    Median values of the bias and variance of the estimated flow map $\hat{\phi}_{1,t}(x)$ over dense grids of $t \in [0,1]$ and $x \in [-3,3]$ (A), and {median values of} the averaged empirical roughness norms and Sobolev norms across 100 simulation replications (B). 
    In each replication, one of 
    \(\lambda_u\), \(\lambda_t\), or \(\lambda_x\) is varied over a range of values, while the other two are fixed at \(10^{-7}\) when obtaining \(\hat{V}\). 
    The averaged norm is computed as the mean of the norms of all generated functions within that simulation. 
    The simulations follow the Gamma--process setting, with the sample size fixed at 100 and the numbers of observed time points set to $\{2,\ldots,6\}$.
    }
    \label{fig:sen_tuning_(A)(B)}
\end{figure}

Following the above setting, we rotate the choice of tuning parameters by setting one of 
\(\lambda_u\), \(\lambda_t\), or \(\lambda_x\) to a range of values while fixing the remaining two at \(10^{-7}\) to obtain $\hat{V}$. 
We then examine how the bias and variance change as a single tuning parameter varies while the others remain fixed.
The results, shown in Figure~S\ref{fig:sen_tuning_(A)(B)}(A), are obtained under the Gamma--process setting in the simulation study (see Section~\ref{sec:simu}). 
We find that the bias of $\hat{V}$ generally increases, whereas the variance decreases, as the tuning parameter becomes larger. 
This aligns with the intuition that a smoother estimated vector field yields lower variance but higher bias.
The estimates with $\lambda_u=\lambda_t=\lambda_x$ and the corresponding bias, variance, and MSE are also provided in Figure~\ref{fig:mse_vector} in the main text.

\noindent\textbf{Smoothness of the generated function:} We also evaluate how the Sobolev norm and the smoothness of the generated functions change as the tuning parameters vary.
Specifically, we focus on another Sobolev norm
\[
\|f\|_{\mathcal W^{2}_{2}(\mathcal{S})}
    = \sqrt{\|f\|^{2}_{\mathcal{L}^{2}} + \|f^{(2)}\|^{2}_{\mathcal{L}^{2}}}, 
    \qquad f \in \mathcal W^{2}_{2}(\mathcal{S}),
\]
and approximate the derivative using discrete differences based on discrete evaluations of a function \( f \).
In this experiment, we rotate the choice of tuning parameters by setting one of 
\(\lambda_u\), \(\lambda_t\), or \(\lambda_x\) to a range of values, while fixing the remaining two at \(10^{-7}\). 
We then examine how the empirical Sobolev norms, as well as the roughness norm 
(defined as \(\|(\cdot)^{(2)}\|_{\mathcal{L}^{2}}\)) of the generated functions {change} as one tuning parameter varies.
The results, shown in Figure~S\ref{fig:sen_tuning_(A)(B)}(B), are obtained under the Gamma--process setting in our simulation study. 
We find that the Sobolev norm of the generated functions is dominated by the roughness norm. 
When the estimated vector field becomes smoother (i.e., when the corresponding tuning parameter is large), the generated functions also become smoother, as indicated by a smaller roughness norm in Figure~S\ref{fig:sen_tuning_(A)(B)}(B). 
This coincides with our intuition that a smoother estimated vector field produces smoother generated functions.

\subsubsection{Effect of NPD on Accuracy and Smoothness:} 
\begin{figure}[ht!]
    \centering
    \includegraphics[width=0.8\linewidth]{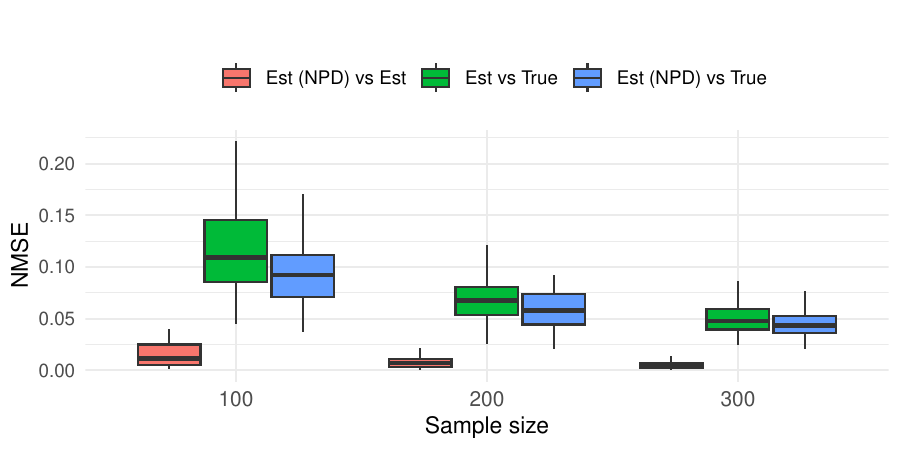}
    \caption{Normalized mean squared error (NMSE) between correlation functions. 
    We compare the raw estimated correlation functions without NPD (denoted by Est), 
    the estimated correlation functions after applying NPD (denoted by Est (NPD)), 
    and the true correlation functions (denoted by True). 
    For calculating NMSE, the denominator for normalization is the $\mathcal{L}^2$ norm of the latter correlation function; 
    that is, ``$A$ vs.\ $B$'' represents 
    $\|A - B\|_{\mathcal{L}^2}^2 / \|B\|_{\mathcal{L}^2}^2$, 
    where $A$ and $B$ are kernels defined on $\mathcal{S}\times\mathcal{S}$ with norm $\|\cdot\|_{\mathcal{L}^2}$. 
    The simulation follows the Gamma--process setting, with the numbers of observed time points set to $\{2,\ldots,6\}$.}
    \label{fig:sen_rho}
\end{figure}

In this part, we examine how NPD affects the accuracy of estimating the true latent correlation functions $\rho$.
{Specifically,} we estimate $\rho$ with and without the NPD procedure and compare each estimator with the true $\rho$, measured by normalized mean squared errors (NMSEs). The results are shown in Figure~S\ref{fig:sen_rho}. 

We observe that applying NPD generally alters the raw estimator $\hat{\rho}$ by less than 5\% in terms of NMSE.
Meanwhile, NPD can further improve the accuracy of {the estimator} compared with the estimator obtained without NPD. This demonstrates a clear benefit of incorporating NPD in our estimation procedure. 

\begin{figure}[ht!]
    \centering
    \includegraphics[width=0.8\linewidth]{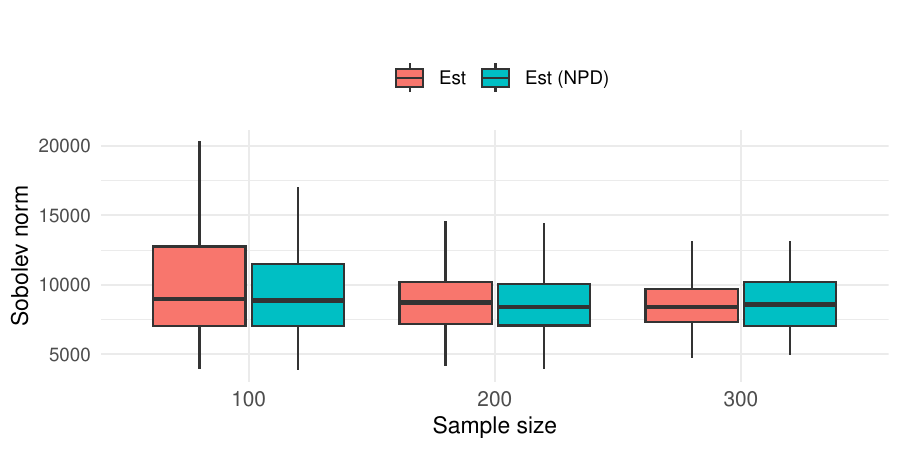}
    \caption{Empirical Sobolev norms of the estimated correlation functions across 100 simulation replications. 
    We compare the raw estimator without NPD (denoted by Est) and the estimator after applying NPD (denoted by Est (NPD)). 
    The simulation follows the Gamma--process setting, with the numbers of observed time points set to $\{2,\ldots,6\}$.}
    \label{fig:sen_rho_smoothness}
\end{figure}

We also compare the empirical Sobolev norms of the correlation functions with and without the NPD procedure, as shown in Figure~S\ref{fig:sen_rho_smoothness}. 

Here, the Sobolev norm of a correlation kernel $f$ is defined by
\(
\|f\|_{\mathcal W^{2}_{2}(\mathcal{S}\times\mathcal{S})}.
\)
We observe that the {median norm} across simulation replications is similar for the two estimators, implying that the NPD procedure does not alter the estimator significantly.

\subsubsection{Discretization Error on Smoothness}
In this part, we examine how the smoothness of the generated functions is affected by {the discretization induced by the time grid} $\mathcal{S}_r$ and the step size $\Delta u$ used in the Runge--Kutta method in Proposition~\ref{theo_smooth_numerical}.
Here, smoothness is measured by the roughness norm defined previously.
\begin{figure}[ht!]
    \centering
    \includegraphics[width=0.9\linewidth]{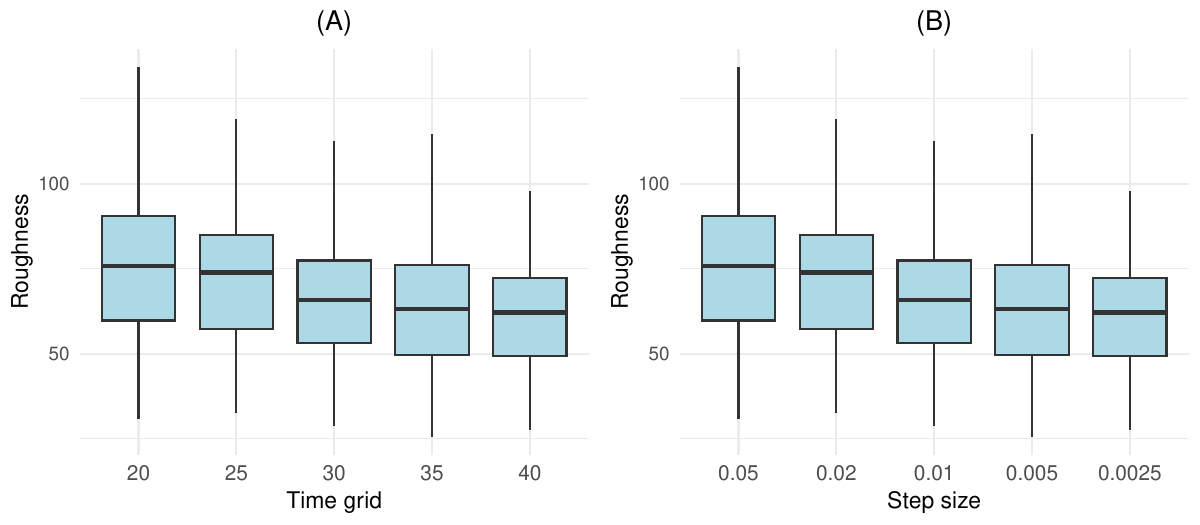}
    \caption{
        Averaged empirical roughness norms from 100 simulation replications. 
        For each replication, the averaged norm is computed as the mean of the norms of all generated functions within that simulation.  
        The simulation follows the Gamma--process setting, with the sample size fixed at 100 and the numbers of observed time points set to $\{2,\ldots,6\}$.
    }
    \label{fig:smoothness_sensitive}
\end{figure}

The results, shown in Figure~S\ref{fig:smoothness_sensitive}, indicate that the smoothness of the generated functions stabilizes once the {number of grid points in} $\mathcal{S}_r$ exceeds $30$ and the Runge--Kutta step size is smaller than $0.01$. 
In our implementation, we therefore set $|\mathcal{S}_r|>30$ and use a step size $\Delta u=0.01$, ensuring stable smoothness in the generated data.

\subsubsection{Variance of Monte Carlo Approximation} 
In this part, we experiment with different values of $F$ and $H$, the number of grid points over \(u\) and the number of Monte Carlo base samples, respectively, used to estimate the vector field $\hat{V}$ in \eqref{Empirical_FM_appro}. 
To quantify the variance of $\hat{V}$ conditional on the observed data, we repeatedly generate random samples $Z_0^{(h)}$ and compute the corresponding estimator $\hat{V}$ for each fixed choice of $F$ and $H$.

In detail, for each resulting $\hat{V}$, we obtain the flow map $\hat{\phi}_{1,t}(x)$ by solving the differential equation
\[
\frac{\partial \hat{\phi}_{u,t}(x)}{\partial u}
= \hat{V}\big(u,t,\hat{\phi}_{u,t}(x)\big),
\qquad 
\hat{\phi}_{0,t}(x)=x.
\]
We then apply Monte Carlo sampling using replicated draws of \(Z_0^{(h)}\) to obtain samples of \(\hat{\phi}_{1,t}\), facilitating the computation of the conditional variance
\[
\int_{0}^{1}\!\!\int_{-3}^{3}
\sqrt{\operatorname{Var}\!\big(\hat{\phi}_{1,t}(x)\mid \text{data}\big)}\,\mathrm{d}x\,\mathrm{d}t.
\]
Here, the interval \([-3,3]\) used in the integration covers the majority of samples generated from a standard normal distribution (the base used in SFM).

\begin{figure}[ht!]
    \centering
    \includegraphics[scale=0.55,trim=0cm 0cm 10.9cm 0cm, clip]{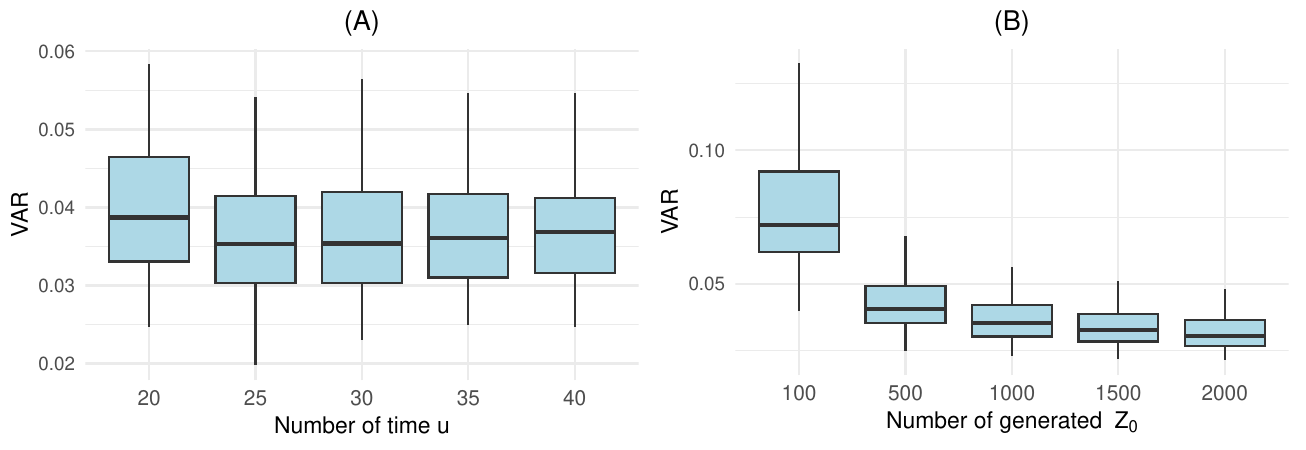}
    \caption{
    Boxplots of the conditional variance from 100 simulation replications under each setting. 
    (A) Varying the number of discretization points in $u$ ($F$) with $H$ fixed at $1000$; 
    (B) varying the number of generated base samples ($H$) with $F$ fixed at $30$. 
    The simulations follow the Gamma--process setting, with sample size fixed at $100$ and the numbers of observed time points set to $\{2,\ldots,6\}$.
    }
    \label{fig:sen_vecfield}
\end{figure}

{The conditional variances are shown in Figure~S\ref{fig:sen_vecfield}.}
We observe that the conditional variance tends to decrease as $F$ and $H$ increase, and becomes stable when $F \ge 30$ and $H \ge 1000$. 
In our implementation, we therefore set $F = 30$ and $H = 1000$.

\subsubsection{Sensitivity Analysis to the Base}

In this part, we conduct a sensitivity analysis regarding the choice of the base $Z_0^{(h)}$ used in Algorithm~\ref{algo:total_algorithm}.
To this end, we consider Gaussian bases with different covariance structures to estimate $\hat{V}$. 
For each case, we compute the flow map $\hat{\phi}_{1,t}(x)$ by solving the differential equation
\[
\frac{\partial \hat{\phi}_{u,t}(x)}{\partial u}
= \hat{V}\big(u,t,\hat{\phi}_{u,t}(x)\big),
\qquad 
\hat{\phi}_{0,t}(x)=x.
\]
We then evaluate the mean squared error (MSE)
\[
\frac{1}{6}\int_{0}^{1}\!\!\int_{-3}^{3}
\bigl|\hat{\phi}_{1,t}(x) - \phi_{1,t}(x)\bigr|^{2}
\,\mathrm{d}x\,\mathrm{d}t,
\]
where $\phi_{1,t}$ denotes the true flow map and the interval $[-3,3]$ contains the majority of samples generated from a standard normal distribution (the base used in SFM). 
The MSE results from 100 simulation replications are shown in Figure~S \ref{fig:sen_vecfield_3}. 
We observe that the resulting forward map based on $\hat{V}$ is generally not sensitive to the choice of the base covariance.

\begin{figure}[ht!]
    \centering
    \includegraphics[scale=0.9,trim=22cm 0cm 0cm 0.7cm, clip]{Figure/sen_vecfield.pdf}
    \caption{
    Mean squared error
    $\frac{1}{6}\int_{0}^{1}\!\!\int_{-3}^{3}
        \bigl|\hat{\phi}_{1,t}(x) - \phi_{1,t}(x)\bigr|^{2}
        \,\mathrm{d}x\,\mathrm{d}t$
    from 100 simulation replications, using a Gaussian base with covariance function
    \(K_{v}(t,s)=\exp(-|t-s|/v)\). Here, $v$ is varied over a range of values.
    The simulation follows the Gamma--process setting, with the numbers of observed time points set to $\{2,\ldots,6\}$.
    }
    \label{fig:sen_vecfield_3}
\end{figure}

\

\subsection{MSE for Mean, Eigen, and Median Function Estimation}

To further evaluate generation quality, we compute the mean squared error (MSE) between the ground-truth targets---namely the mean function (MF), the first two eigenfunctions (EF1, EF2), and the median function (MDF)---and their corresponding estimates from the generated data. The median function is defined pointwise as the median of the random variable \(X(t)\), for \(t\in\mathcal{S}\). In the Gaussian case, we report MSEs for MF, EF1, and EF2; in the Gamma case, we additionally report MDF. As a baseline, we also include an oracle method that applies FPCA \citep{hsing2015theoretical} directly to the observed training data to estimate MF, EF1, and EF2.

We present the MSEs between the true MF, EF1, EF2, and (when applicable) MDF and their corresponding estimates obtained from the raw data and from the generated data in the simulation study, as summarized in Table~S\ref{tab: MSE}. The MDF values are not reported for the Gaussian process case (A), since the mean and median coincide in this setting, implying that MDF \(=\) MF. The MDF values are also not reported for the Oracle method, since the oracle FPCA procedure \citep{hsing2015theoretical} is not designed for median-function estimation.

The results in Table~S\ref{tab: MSE} show that SFM, GP, and KL achieve comparable performance in recovering the mean function and the leading eigenfunctions, and they consistently outperform DSM and FM. Compared to GP and KL, SFM provides more accurate estimation of the median trend in the Gamma setting, highlighting its advantage in capturing distributional features beyond second-order moments---features that GP and KL are not designed to target. Notably, in both settings, SFM performs near or on par with the Oracle method in estimating both the mean function and the eigenfunctions. These findings collectively demonstrate the robustness and flexibility of SFM in handling complex and heavy-tailed functional data.

\begin{table}[htbp]

\caption{Mean squared errors (MSE) between the true mean function (MF), eigenfunctions (EF), median function (MDF), and their estimates from raw/generated data. "Oracle" refers to MF/EF estimated directly from observed data using the smoothing method of \cite{hsing2015theoretical}. } \label{tab: MSE}
\centering
\renewcommand{\arraystretch}{1.1}
\setlength\tabcolsep{6pt}
\footnotesize
\begin{tabular}{c|c|c|cccc|cccc}
  \hline
\multicolumn{3}{c|}{\multirow{2}{*}{MSE}}
   & \multicolumn{4}{c|}{$\{2,\ldots,6\}$} & \multicolumn{4}{c}{$\{6,\ldots,10\}$} \\
\multicolumn{3}{c|}{} & MF & EF1 & EF2 & MDF & MF & EF1 & EF2 & MDF  \\
  \hline
  \multirow{18}{*}{(A)} 
  & \multirow{6}{*}{$n=100$}  & 
DSM & 0.45 & 1.07 & 1.10 &  & 0.20 & 1.12 & 1.18 &  \\ 
&&  FM & 0.73 & 1.02 & 0.99 &  & 0.43 & 1.06 & 1.00 &  \\ 
&&  GP & 0.12 & 0.25 & 0.39 &  & 0.11 & 0.19 & 0.31 &  \\ 
&&  KL & 0.13 & 0.27 & 0.41 &  & 0.10 & 0.19 & 0.30 &  \\ 
&&  SFM & 0.12 & 0.29 & 0.39 &  & 0.12 & 0.20 & 0.30 &  \\ 
&&  Oracle & 0.10 & 0.21 & 0.36 &  & 0.08 & 0.16 & 0.27 &  \\ 
  \cline{2-11}
  & \multirow{6}{*}{$n=200$}  &    
DSM & 0.38 & 1.07 & 1.07 &  & 0.18 & 1.17 & 1.17 &  \\ 
&&  FM & 0.55 & 1.00 & 1.01 &  & 0.28 & 1.08 & 1.06 &  \\ 
&&  GP & 0.11 & 0.20 & 0.32 &  & 0.09 & 0.16 & 0.24 &  \\ 
&&  KL & 0.10 & 0.20 & 0.29 &  & 0.09 & 0.15 & 0.22 &  \\ 
&&  SFM & 0.10 & 0.22 & 0.31 &  & 0.09 & 0.15 & 0.21 &  \\ 
&&  Oracle & 0.08 & 0.17 & 0.26 &  & 0.06 & 0.12 & 0.19 &  \\ 
  \cline{2-11}
  & \multirow{6}{*}{$n=300$}  & 
DSM & 0.41 & 1.11 & 1.11 &  & 0.22 & 1.24 & 1.24 &  \\ 
&&  FM & 0.52 & 1.05 & 1.02 &  & 0.33 & 1.13 & 1.04 &  \\ 
&&  GP & 0.10 & 0.17 & 0.26 &  & 0.09 & 0.15 & 0.22 &  \\ 
&&  KL & 0.10 & 0.17 & 0.25 &  & 0.09 & 0.14 & 0.21 &  \\ 
&&  SFM & 0.09 & 0.19 & 0.26 &  & 0.09 & 0.14 & 0.21 &  \\ 
&&  Oracle & 0.07 & 0.15 & 0.22 &  & 0.05 & 0.10 & 0.15 &  \\ 
  \hline
\multirow{18}{*}{(B)} 
  & \multirow{6}{*}{$n=100$}  & 
DSM & 0.59 & 1.15 & 1.15 & 0.37 & 0.28 & 1.26 & 1.21 & 0.27 \\ 
&&  FM & 0.78 & 1.08 & 1.02 & 0.20 & 0.37 & 1.17 & 0.95 & 0.18 \\ 
&&  GP & 0.11 & 0.66 & 0.76 & 0.29 & 0.09 & 0.58 & 0.65 & 0.29 \\ 
&&  KL & 0.11 & 0.71 & 0.80 & 0.29 & 0.10 & 0.58 & 0.70 & 0.29 \\ 
&&  SFM & 0.12 & 0.57 & 0.65 & 0.12 & 0.10 & 0.54 & 0.58 & 0.12 \\ 
&&  Oracle & 0.09 & 0.65 & 0.75 &  & 0.07 & 0.53 & 0.63 &  \\ 
  \cline{2-11}
  & \multirow{6}{*}{$n=200$}  &    
DSM & 0.52 & 1.18 & 1.17 & 0.44 & 0.24 & 1.25 & 1.19 & 0.24 \\ 
&&  FM & 0.81 & 1.06 & 1.04 & 0.20 & 0.51 & 1.18 & 1.03 & 0.13 \\ 
&&  GP & 0.08 & 0.59 & 0.72 & 0.28 & 0.08 & 0.50 & 0.59 & 0.28 \\ 
&&  KL & 0.09 & 0.59 & 0.72 & 0.29 & 0.08 & 0.51 & 0.60 & 0.28 \\ 
&&  SFM & 0.10 & 0.48 & 0.54 & 0.11 & 0.09 & 0.52 & 0.57 & 0.10 \\ 
&&  Oracle & 0.06 & 0.55 & 0.69 &  & 0.05 & 0.42 & 0.52 &  \\ 
  \cline{2-11}
  & \multirow{6}{*}{$n=300$}  & 
DSM & 0.43 & 1.21 & 1.20 & 0.33 & 0.28 & 1.27 & 1.24 & 0.26 \\ 
&&  FM & 0.69 & 1.11 & 1.05 & 0.21 & 0.39 & 1.16 & 0.96 & 0.12 \\ 
&&  GP & 0.09 & 0.56 & 0.65 & 0.29 & 0.08 & 0.41 & 0.49 & 0.29 \\ 
&&  KL & 0.08 & 0.59 & 0.68 & 0.29 & 0.08 & 0.48 & 0.57 & 0.29 \\ 
&&  SFM & 0.10 & 0.52 & 0.57 & 0.10 & 0.08 & 0.45 & 0.47 & 0.10 \\ 
&&  Oracle & 0.05 & 0.50 & 0.59 &  & 0.04 & 0.37 & 0.43 &  \\
  \hline
\end{tabular}
\end{table}

\subsection{Simulation {Results} for Densely Observed Functional Data}

\begin{figure}[ht!]
    \centering
    \includegraphics[width=0.7\linewidth]{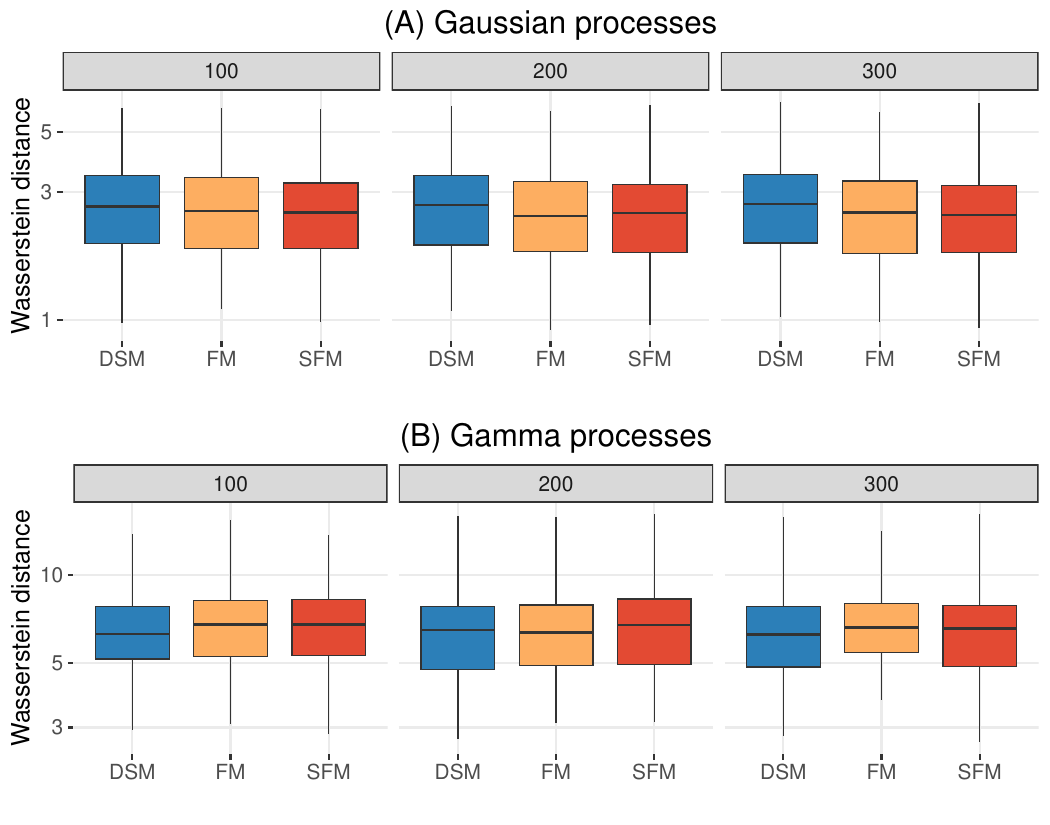}
    \caption{
    Boxplots of Wasserstein distances from 100 simulation replications under varying sample sizes \( n \) and numbers of observed time points \( J_i \in \{27, \ldots, 31\} \).
    }
    \label{fig:curve_error_dense}
\end{figure}

In this part, we consider the dense case for data generation in Section~\ref{sec:simu}, where the number of observed time points satisfies 
\( J_i \in \{27, \ldots, 31\} \). 
In this setting, the interpolation step does not introduce substantial error (unlike in the sparse case), and thus the {interpolation} has limited impact on subsequent data generation. 
The results are provided in Figure~S\ref{fig:curve_error_dense}, where we observe that the deep generative methods (DSM and FM) and SFM behave similarly across different scenarios. 
This finding suggests that the poorer performance of the deep generative methods under sparse designs in Section~\ref{sec:simu} is attributable to the interpolation step required prior to training.

\

\subsection{Smoothness Diagnostics}

\begin{figure}[ht!]
    \centering
    \includegraphics[width=0.8\linewidth]{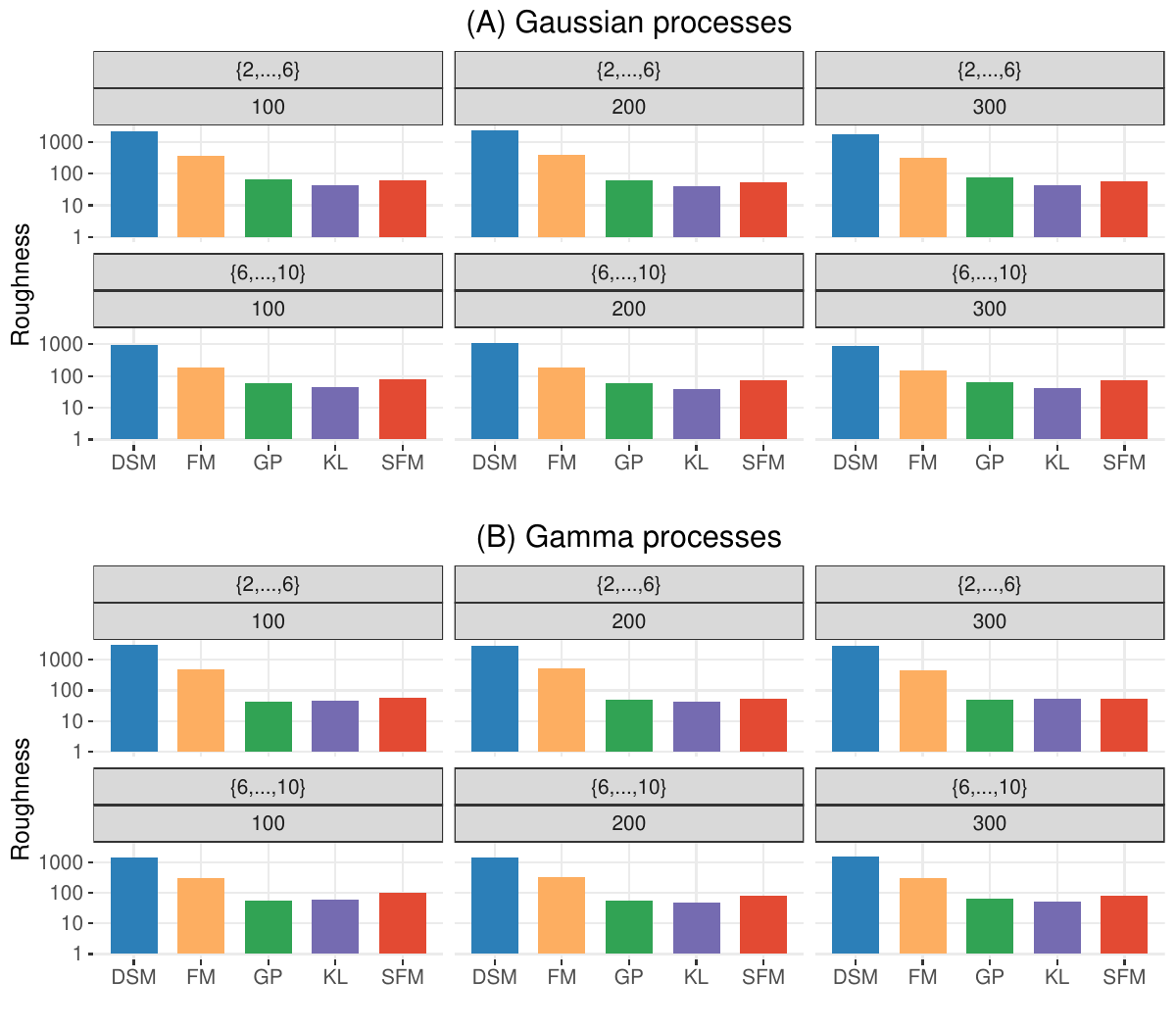}
    \caption{
    Averaged empirical roughness norms $\|f^{(2)}\|_{\mathcal{L}^{2}}$ for a function $f$ from 100 simulation replications. 
    For each replication, the averaged norm is computed as the mean of the norms of all generated functions within that simulation.
    }
    \label{fig:smoothness_diagnostics}
\end{figure}

In this part, we conduct smoothness diagnostics on the generated curves in Section~\ref{sec:simu} (see Figure~S\ref{fig:smoothness_diagnostics}), where smoothness is quantified using the roughness norm $\|f^{(2)}\|_{\mathcal{L}^{2}}$ for a function $f$. 
In both the Gaussian and Gamma settings, we find that the roughness of the curves generated by GP, KL, and SFM {is} quite similar, whereas the roughness of the curves generated by the deep generative methods (DSM and FM) is substantially larger.  
This demonstrates the superior ability of SFM to produce smooth, realistic functional trajectories, consistent with Theorem~\ref{theo: smooth}.

\

\subsection{Noise Impact in Smooth Flow Matching}\label{sec:sen_noise}

Following the Gamma process model in Section~\ref{sec:simu}, we generate noisy observations as
\(Y_{ij}=X_i(T_{ij})+\tau_{ij}\), for \(i=1,\ldots,n\) and \(j=1,\ldots,J_i\),
where \(\tau_{ij}\) are independent mean-zero Gaussian {noise variables} with variance
\[
\operatorname{Var}(\tau_{ij})
=
\Big(\frac{1}{J_i}\sum_{k=1}^{J_i}X_i^2(T_{ik})\Big)\times \text{noise level},
\qquad i=1,\ldots,n.
\]
Based on these noisy and irregular observations, we apply SFM to learn the vector field for data generation.
In addition, we apply SFM to the denoised data $\hat X_i(T_{ij})$ obtained from Remark~\ref{Re: noise}, where the denoised values are used as inputs to SFM.

\begin{figure}[htbp]
    \centering
    \includegraphics[width=1\linewidth]{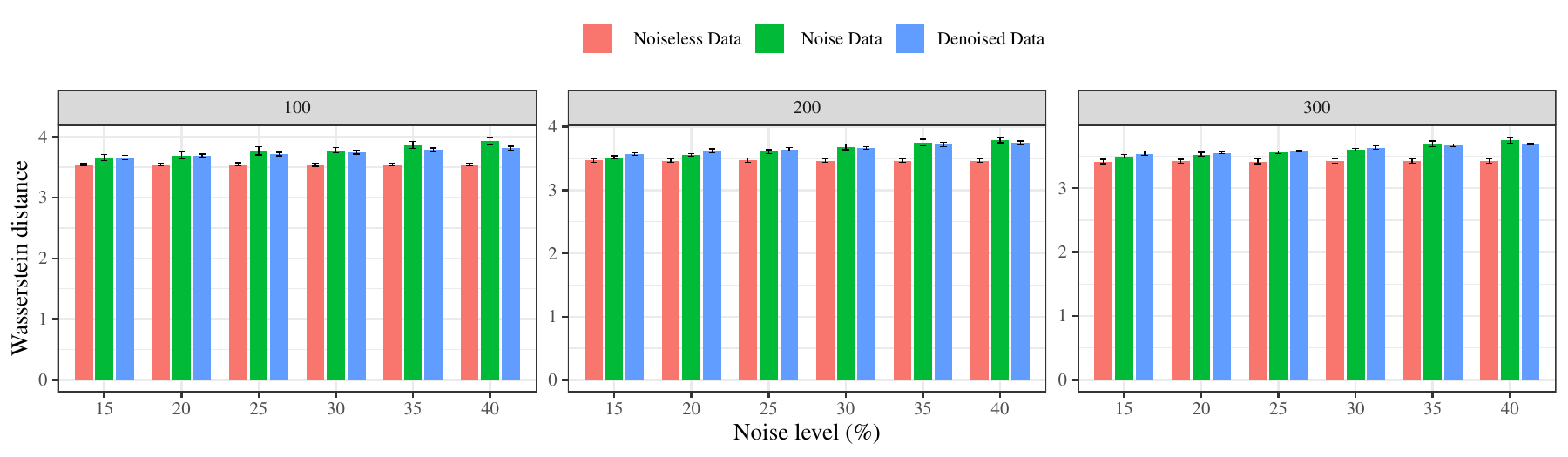}
    \caption{Wasserstein distances from 100 simulation replications under varying sample sizes \(n\).
    The simulation follows the Gamma--process setting, with the numbers of observed time points set to \(\{6,\ldots,10\}\).}
    \label{sensitivity}
\end{figure}

\

\subsection{Estimated Mean and Eigenfunctions from Real Data}
\begin{figure}[htbp]
    \centering
    \includegraphics[width=0.9\linewidth]{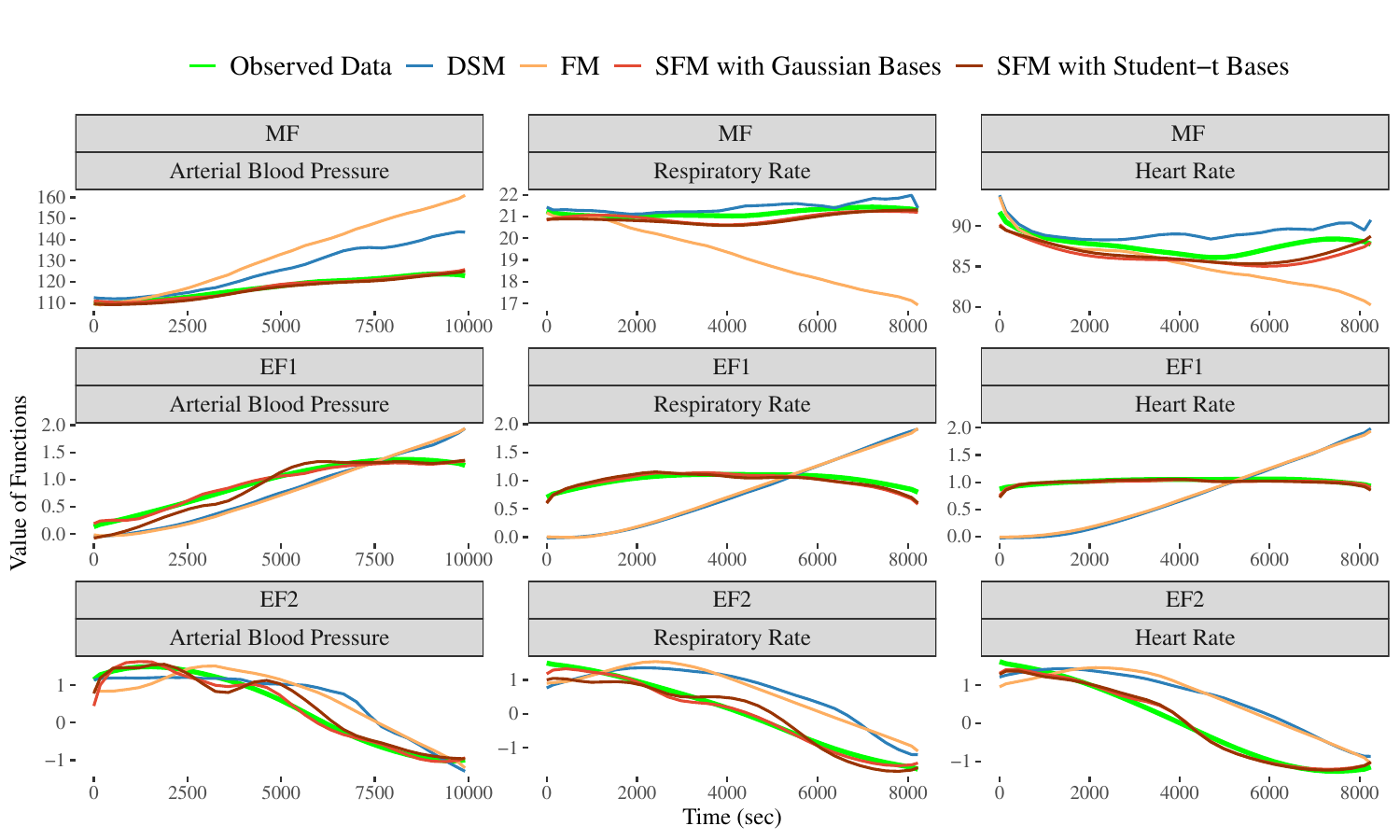}
    \caption{Estimated mean function (MF) and the first two eigenfunctions (EF1 and EF2) for the observed and synthesized functional data.}
    \label{fig:functional_pattern_2}
\end{figure}

The estimated mean function and eigenfunctions from the real data analysis in Section~\ref{sec:real} are shown in Figure~S\ref{fig:functional_pattern_2}.

\end{document}